\RequirePackage{fix-cm}

\let\accentvec\vec

\documentclass[twocolumn]{svjour3} 

\let\vec\accentvec
\smartqed  % flush right qed marks, e.g. at end of proof
\usepackage{graphicx}
\usepackage{natbib}
\usepackage{marvosym}
\usepackage{amssymb}
\usepackage{booktabs} % Allows the use of \toprule, \midrule and \bottomrule in tables
\usepackage{multirow}
\usepackage{hyperref}
\usepackage{color}

\usepackage[linesnumbered,ruled]{algorithm2e}
\SetKwProg{Fn}{Function}{}{end}\SetKwFunction{FRecurs}{FnRecursive}%

\usepackage{amsmath}

\usepackage{etoolbox}
\newcommand*{\affaddr}[1]{#1} % No op here. Customize it for different styles.
\newcommand*{\affmark}[1][*]{\textsuperscript{#1}}

\begin{document}

\title{An Improved Epsilon Constraint-handling Method in MOEA/D for CMOPs with Large Infeasible Regions}

%An Improved $\varepsilon$-constrained Method in MOEA/D for CMOPs with Large Infeasible Regions Near Their Pareto Fronts

% \subtitle{Do you have a subtitle?\\ If so, write it here}

\titlerunning{An Improved $\varepsilon$-constrained Method in MOEA/D for CMOPs with Large Infeasible Regions}        % if too long for running head

\author{Zhun Fan  \affmark[1]         \and
        Wenji Li  \affmark[1]         \and
        Xinye Cai \affmark[2]         \and
        Han Huang \affmark[3]         \and
        Yi Fang   \affmark[1]         \and
        Yugen You \affmark[1]         \and
        Jiajie Mo \affmark[1]         \and
        Caimin Wei \affmark[4]        \and   
        Erik Goodman  \affmark[5]
}


% \institute{Department of Electronic Engineering, Shantou University, Guangdong, 515063, China \and
%            College of Computer Science and Technology, Nanjing University of Aeronautics and Astronautics, Jiangsu, 210016, China \and
%            Department of Mathematics, Shantou University, Guangdong, 515063, China \and
%            School of Software Engineering, South China University of Technology, Guangdong, 515063, China \and
%            BEACON Center for the Study of Evolution in Action, Michigan State University. East Lansing, Michigan, USA.
%            } 

\institute{
\affaddr{\affmark[1]Department of Electronic Engineering, Shantou University, Guangdong, 515063, China}\\
\affaddr{\affmark[2]College of Computer Science and Technology, Nanjing University of Aeronautics and Astronautics, Jiangsu, 210016, China}\\
\affaddr{\affmark[3]School of Software Engineering, South China University of Technology, Guangdong, 515063, China}\\
\affaddr{\affmark[4]Department of Mathematics, Shantou University, Guangdong, 515063, China}\\
\affaddr{\affmark[5]BEACON Center for the Study of Evolution in Action, Michigan State University. East Lansing, Michigan, USA.}\\
}

\date{Received: date / Accepted: date}

\maketitle

\begin{abstract}
This paper proposes an improved epsilon constraint-handling mechanism, and combines it with a decomposition-based multi-objective evolutionary algorithm (MOEA/D) to solve constrained multi-objective optimization problems (CMOPs). The proposed constrained multi-objective evolutionary algorithm (CMOEA) is named MOEA/D-IEpsilon. It adjusts the epsilon level dynamically according to the ratio of feasible to total solutions (RFS) in the current population. In order to evaluate the performance of MOEA/D-IEpsilon, a new set of CMOPs with two and three objectives is designed, having large infeasible regions (relative to the feasible regions), and they are called LIR-CMOPs. Then the fourteen benchmarks, including LIR-CMOP1-14, are used to test MOEA/D-IEpsilon and four other decomposition-based CMOEAs, including MOEA/D-Epsilon, MOEA/D-SR, MOEA/D-CDP and C-MOEA/D. The experimental results indicate that MOEA/D-IEpsilon is significantly better than the other four CMOEAs on all of the test instances, which shows that MOEA/D-IEpsilon is more suitable for solving CMOPs with large infeasible regions. Furthermore, a real-world problem, namely the robot gripper optimization problem, is used to test the five CMOEAs. The experimental results demonstrate that MOEA/D-IEpsilon also outperforms the other four CMOEAs on this problem.

\end{abstract}

\section{Introduction}
\label{intro}
Real-world optimization problems usually involve the simultaneous optimization of multiple conflicting objectives with a number of constraints. Without loss of generality, a CMOP considered in this paper is defined as follows (\cite{deb2001multi}): 

\begin{equation}
\label{equ:cmop_definition}
\begin{cases}
\mbox{minimize} &\mathbf{F}(\mathbf{x}) = {(f_{1}(\mathbf{x}),\ldots,f_{m}(\mathbf{x}))} ^ {T} \\
\mbox{subject to} & g_i(\mathbf{x}) \ge 0, i = 1,\ldots,q \\
& h_j(\mathbf{x}) = 0, j= 1,\ldots,p \\
&\mathbf{x} \in{\mathbb{R}^n}
\end{cases}
\end{equation}
where $F(\mathbf{x}) = ({f_1}(\mathbf{x}),{f_2}(\mathbf{x}), \ldots ,{f_m}(\mathbf{x})) ^ T \in \mathbb{R} ^m$ is an $m$-dimensional objective vector, ${g_i}(\mathbf{x}) \ge 0$ is an inequality constraint, and ${h_j}(\mathbf{x})=0$ is an equality constraint. $\mathbf{x} \in \mathbb{R}^n$ is an $n$-dimensional decision vector. The feasible region $S$ is defined as the set $\{\mathbf{x} | g_i(\mathbf{x}) \ge 0, i = 1,\ldots,q$  and $h_j(\mathbf{x}) = 0, j= 1,\ldots,p\}$.

In CMOPs, there are usually more than one constraint. The overall constraint violation is a widely used approach to deal with constraint violations, as it summarizes them into a single scalar, as follows: 
\begin{eqnarray}
\label{equ:constraint}
\phi(\mathbf{x}) = \sum_{i=1}^{q} |\min(g_i(\mathbf{x}),0)| + \sum_{j = 1}^{p} |h_j(\mathbf{x})|
\end{eqnarray}
If $\phi(\mathbf{x}) = 0$, $\mathbf{x}$ is feasible; otherwise, it is infeasible. Any solution in set $S$ is feasible, and for any two solutions $\mathbf{x}^1 \in S$ and $\mathbf{x}^2 \in S$, $\mathbf{x}^1$ is said to dominate $\mathbf{x}^2$ if $f_i(\mathbf{x}^1) \le f_i(\mathbf{x}^2)$ for each $i \in \{1,...,m\}$ and $f_j(\mathbf{x}^1) < f_j(\mathbf{x}^2)$ for at least one $j \in \{1,...,m\}$, denoted as $\mathbf{x}^1 \preceq \mathbf{x}^2$. For a solution $\mathbf{x}^* \in S$, if there is no other solution in $S$ dominating $\mathbf{x}^*$, then $\mathbf{x}^*$ is called a Pareto optimal solution. A set including all of the Pareto optimal solutions is called a Pareto optimal set ($PS$). Mapping the $PS$ into the objective space obtains a set of objective vectors, which is called a Pareto optimal front ($PF$), and $PF = \{F(\mathbf{x})| \mathbf{x} \in PS\}$. 

CMOEAs aim to find a representative set of Pareto optimal solutions. They have to tackle the multiple conflicting objectives with a number of constraints simultaneously, and to maintain a good balance between convergence and diversity of the achieved solutions. In CMOEAs, there are two basic components: one is the constraint-handling mechanism, and the other is the multi-objective evolutionary algorithm (MOEA).

In terms of constraint-handling, many methods have been proposed in evolutionary optimization (\cite{Cai:2013iz,Hu:2013kc}). They can be roughly divided into penalty function methods, special representations and operators, repair methods, separation of objectives and constraints and hybrid methods (\cite{CoelloCoello20021245}). The penalty function method is widely used due to its simplicity in the constraint handling (\cite{Runarsson:2005jd}). However, the ideal penalty factors cannot be known in advance for an arbitrary CMOP, and tuning the penalty factors can be a very tedious task. 

In recent years, a number of other constraint-handling techniques have had a relatively high impact in evolutionary optimization, including feasibility rules, stochastic ranking, $\varepsilon$-constrained method, novel penalty functions, novel special operators, multi-objective concepts and ensemble of constraint-handling techniques (\cite{MezuraMontes:2011cj}). However, most of them aim to solve constrained scalar optimization problems when they are first proposed.

MOEAs can be classified into three different types according to their selection approaches. The first type is non-dominated-based methods, and representative examples include NSGA-II (\cite{996017}), PAES-II (\cite{corne2001pesa}), SPEA-II (\cite{zitzler2001spea2}), NSGA-III (\cite{Deb:2014du}) and so on. The second type is decomposition-based approaches, and typical 
examples include MOEA/D (\cite{Zhang:2007va}), MOEA/D-DE (\cite{Li:2009vo}), EAG-MOEA/D (\cite{Cai:2015gi}), MOEA/D-M2M (\cite{Liu:2014jb}), MOEA/D-SAS (\cite{Cai:2016ii}) and so on. Currently, MOEA/D is a popular algorithm to solve unconstrained multi-objective optimization problems (MOPs). MOEA/D decomposes a MOP into many scalar optimization subproblems, and optimizes them simultaneously in a collaborative way. The last type is indicator-based methods. This type of MOEAs selects solutions based on the improvement of a performance metric. Representative methods include IBEA (\cite{Zitzler:2004tm}), SMS-EMOA (\cite{Beume:2007to}), HypE (\cite{Bader:2011tg}), FV-MOEA (\cite{Jiang:2015hj}) and so on. 

There are two commonly used test suites of CMOPs, including CTP (\cite{ deb2001multi}) and CF test instances (\cite{zhang2008multiobjective}). For CTP1-CTP5 and CF1-CF10, the feasible regions are relatively large, and a CMOEA can approximate their PFs without encountering any infeasible obstacles during the entire evolutionary process. Thus, CTP1-5 and CF1-10 are not good test problems to evaluate the performance of constraint-handling mechanisms. For the remaining test problems CTP6-8, the feasible regions are relatively large, and the population of a CMOEA can reach these regions with high probability. Thus, CTP and CF test suites can not effectively measure the performance of constraint-handling techniques. When solving CTP (\cite{deb2001multi}) and CF (\cite{zhang2008multiobjective}) test instances, the constraint dominance principle (CDP) (\cite{996017}) is good enough to handle the constraints.

To overcome the shortcomings of the CTP and CF test suites discussed above, we propose a set of new CMOPs (named LIR-CMOP1-14). Each of them has a number of large infeasible regions, and the feasible regions are relatively small. The population of a CMOEA cannot easily discover these small feasible regions, which brings new challenges to the existing CMOEAs. In fact, many real-world optimization problems also have this characteristic. For example, the robot gripper optimization problem considered in this paper has large infeasible regions as illustrated in Section \ref{sec:6}. Thus, it has important significance in practice to design specific mechanisms for solving CMOPs with large infeasible regions.

In this paper, we propose an improved $\varepsilon$-constrained version of MOEA/D to deal with CMOPs. Compared with the original $\varepsilon$-constrained method (\cite{takahama2006constrained}), the proposed method can keep a good balance in the search between the feasible and infeasible regions. It uses the information of the feasible ratio of the population to dynamically balance the exploration between the feasible regions and infeasible regions.

The remainder of the paper is organized as follows. Section \ref{sec:2} introduces related work on MOEA/D and the existing CMOEAs based on MOEA/D. Section \ref{sec:3} illustrates the improved epsilon constraint-handling method as here embedded in MOEA/D. Section \ref{sec:4} designs a set of new CMOPs (LIR-CMOPs) with large infeasible regions. Section \ref{sec:5} describes a comprehensive set of experiments to compare the proposed CMOEA (MOEA/D-IEpsilon) with four other CMOEAs, including MOEA/D-Epsilon, MOEA/D-SR, MOEA/D-CDP and C-MOEA/D. In Section \ref{sec:6}, a robot gripper optimization problem is used to test MOEA/D-IEpsilon and the other four CMOEAs. Finally, Section \ref{sec:7} presents the conclusions.

\section{Related work}
\label{sec:2}
\subsection{MOEA/D}
\label{sec:2.1}

MOEA/D (\cite{Zhang:2007va}) decomposes a MOP into a number of scalar optimization subproblems and optimizes them simultaneously in a collaborative way. Each subproblem is defined by a decomposition function with a weight vector $\lambda^i$. In MOEA/D, a set of $N$ uniformly spread weight vectors $\lambda^1,\ldots,\lambda^N$ are adopted to formulate $N$ subproblems. The weight vectors $\lambda^i$ satisfy $\sum_{k = 1} ^ {m}\lambda_{k}^{i} = 1$ and $\lambda_{k}^i \ge 0$ for each $k \in \{1,\ldots,m\}$. In terms of decomposition methods, there are three commonly used approaches, including weighted sum (\cite{miettinen1999nonlinear}), Tchebycheff (\cite{miettinen1999nonlinear}) and boundary intersection approaches (\cite{Zhang:2007va}).

In the weighted sum approach, each subproblem is defined by summing each objective weighted by a different weight. The $j$-th subproblem with the weighted sum decomposition method is defined as follows:
\begin{eqnarray}
&\nonumber \mbox{minimize} &g^{te}(\mathbf{x}|\lambda) = \sum_{i = 1} ^{m} \lambda_i^jf_i(\mathbf{x})\\
&\mbox{subject to} & \mathbf{x} \in{S}
\label{equ:wsmethod}
\end{eqnarray}
For a minimizing MOP, in the case of a convex PF, the weighted sum approach can work well. However, if the PF is non-convex, only a part of PF can be found by this approach.

In the Tchebycheff decomposition method, the $j$-th subproblem is defined as follows: 
\begin{eqnarray}
&\nonumber \mbox{minimize} &g^{te}(\mathbf{x}|\lambda,z^{*}) = \max_{1 \le i \le m} \{\lambda_i^j |f_i(\mathbf{x}) - z_{i}^{*}| \}\\
&\mbox{subject to} & \mathbf{x} \in{S}
\label{equ:tchmethod-a}
\end{eqnarray}
where $z^* = (z_1^*,\ldots,z_m^*)$ is the ideal point, and $z_i^* = \min \{f_i(\mathbf{x} | \mathbf{x} \in S\}$. The Tchebycheff method is a widely used decomposition approach. It can approximate both concave and convex parts of PFs.

In the boundary intersection approach, two distances $d_1$ and $d_2$ are defined to evaluate the convergence and diversity respectively. The $j$-th subproblem is defined as follows: 
\begin{eqnarray}
&\nonumber \mbox{minimize} &g^{pbi}(\mathbf{x}|\lambda^j,z^{*}) = d_1 + \theta d_2\\
&\mbox{subject to} & \mathbf{x} \in{S}\\
&\nonumber \mbox{where} & d_1 = \frac{\lVert (F(\mathbf{x}) - z^*)^{T} \lambda^j \rVert}{ \lVert \lambda^j \rVert}\\
&&\nonumber d_2 = \lVert (F(\mathbf{x}) - z^* ) - d_1 \frac{\lambda^j} {\lVert \lambda^j \rVert}) \rVert
\label{equ:pbimethod}
\end{eqnarray}
The boundary intersection method is able to solve MOPs with any shape of PFs. However, the penalty factor $\theta$ must be set in advance.

\subsection{Decomposition-based CMOEAs}
\label{sec:2.2}

In decomposition-based CMOEAs, a CMOP is decomposed into a set of constrained scalar optimization subproblems, and these subproblems are solved in a collaborative way simultaneously. Representative methods include C-MOEA/D (\cite{asafuddoula2012adaptive}), MOEA/D-Epsilon (\cite{Yang:2014vt}), MOEA/D-CDP (\cite{jan2013study}) and MOEA/D-SR (\cite{jan2013study}).

C-MOEA/D (\cite{asafuddoula2012adaptive}) embeds an epsilon constraint-handling approach into MOEA/D, and the epsilon value is set adaptively. To be more specific, the epsilon level is set to $CV_{mean} * FR$. $CV_{mean}$ denotes the mean value of the overall constraint violation in the current population, and $FR$ ($\frac{\text{Number of feasible solutions}}{\text{Population size}}$) denotes the feasible ratio of solutions in the current population. For two solutions, if their overall constraint violations are both less than $CV_{mean} * FR$ or their overall constraint violations are equal, the one with the better aggregation value is selected. Otherwise, the one with the smaller overall constraint violation is selected.

MOEA/D-Epsilon (\cite{Yang:2014vt}) also adopts the epsilon method to handle constraints. Unlike C-MOEA/D, the epsilon value in MOEA/D-Epsilon is set dynamically with the increase of generation counter $K$. The detailed setting of the epsilon value can be found in (\cite{takahama2006constrained}).

MOEA/D-CDP (\cite{jan2013study}) adopts CDP (\cite{996017}) to deal with constraints in the framework of MOEA/D. There are three basic rules to select solutions. For two feasible solutions, the one with the better aggregation value is selected. For two infeasible solutions, the one with the smaller overall constraint violation is selected. For a feasible and an infeasible solution, the feasible one is selected.

MOEA/D-SR (\cite{jan2013study}) embeds the stochastic ranking method (SR) (\cite{runarsson2000stochastic}) in MOEA/D to deal with constraints. A parameter $p_f \in [0,1]$ is set to balance the selection between the objectives and the constraints in MOEA/D-SR. For two solutions, if a random number is less than $p_f$, the one with the better aggregation value is selected into the next generation. If the random number is greater than $p_f$, the solutions selection is similar to that of MOEA/D-CDP. In the case of $p_f = 0$, MOEA/D-SR is equivalent to MOEA/D-CDP. 

In summary, C-MOEA/D and MOEA/D-Epsilon both adopt the epsilon constraint-handling approach to solve CMOPs. To get across large infeasible regions, $\varepsilon$ should be increased at sometimes, and be greater than the maximum overall constraint violation in the current population. However, in C-MOEA/D, $\varepsilon$ is always less or equal than $CV_{mean}$, and in MOEA/D-Epsilon, $\varepsilon$ is always decreasing during the evolutionary process. In MOEA/D-CDP, feasible solutions are always better than infeasible solutions. Thus, the infeasible solutions which can help to get across large infeasible regions are difficult to survive. MOEA/D-SR applies a parameter $p_f$ to balance the searching between the feasible and infeasible regions. In order to get across large infeasible regions, $p_f$ should be set dynamically. However, $p_f$ is a static parameter in MOEA/D-SR. To overcome the shortcomings of the four decomposition-based CMOEAs discussed above, an improved epsilon constraint-handling method embedded in MOEA/D is proposed.

\section{The Proposed method}
\label{sec:3}

In this section, the concept of epsilon level comparison, the original epsilon level setting method and the improved epsilon level setting approach are described.

\subsection{Epsilon Level Comparison}
In the epsilon constraint handling approach (\cite{takahama2006constrained}), the relaxation of constraints is controlled by the epsilon level $\varepsilon$. For two solutions $\mathbf{x}^1$ and $\mathbf{x}^2$, their overall constraint violations are $\phi^1$ and $\phi^2$. Then, for any $\varepsilon$ satisfying $\varepsilon \ge 0$, the epsilon level comparison $\preceq_{\varepsilon}$ is defined as follows:

\begin{eqnarray}
\label{equ:epsilon_comparisons}
& (\mathbf{x}^1,\phi^1) \preceq_{\varepsilon}  (\mathbf{x}^2,\phi^2)\Leftrightarrow 
& \begin{cases}
\mathbf{x}^1 \preceq \mathbf{x}^2, \text{if } \phi^1, \phi^2 \le \varepsilon \\
\mathbf{x}^1 \preceq \mathbf{x}^2, \text{if } \phi^1 = \phi^2 \\
\phi^1 < \phi^2, \text{otherwise}\\
\end{cases}
\end{eqnarray}
In Eq. (\ref{equ:epsilon_comparisons}), the epsilon comparison approach is equivalent to CDP (\cite{996017}) when $\varepsilon = 0$. In the case of $\varepsilon = \infty$, it does not consider any constraints. In other words, the comparison between any two solutions is based on their non-dominated ranks on objectives when $\varepsilon = \infty$.

\subsection{Epsilon Level Setting}
In the epsilon constraint-handling method, the setting of $\varepsilon$ is quite critical. In (\cite{takahama2006constrained}), an epsilon level setting method is suggested as follows:

\begin{eqnarray}
&\varepsilon(k) = \begin{cases}
\varepsilon(0)(1 - \frac{k}{T_c})^{cp}, 0 < k < T_c, \varepsilon(0) = \phi(\mathbf{x}^{\theta})  \\
0, k \ge T_c \\
\end{cases}
\label{equ:epsilon_setting}
\end{eqnarray}
where $\mathbf{x}^\theta$ is the top $\theta$-th individual of the initial population sorted by overall constraint violations in a descending order. $cp$ is to control the speed of reducing relaxation of constraints. $\varepsilon(k)$ is updated until the generation counter $k$ reaches the control generation $T_c$. When $k \ge T_c$, $\varepsilon(k) = 0$. The recommended parameter ranges in (\cite{takahama2006constrained}) are listed as follows: $\theta = (0.05 * N)$, $cp \in [2,10]$ and $T_c \in [0.1T_{max}, 0.8T_{max}]$. $N$ denotes the population size, and $T_{max}$ represents the maximum evolutionary generation.

\subsection{Improved Epsilon Level Setting}
The setting of $\varepsilon(k)$ in Eq.\eqref{equ:epsilon_setting} is always decreasing during the evolutionary process, which may not be suitable to solve CMOPs with large infeasible regions. To overcome this problem, an improved epsilon setting approach is suggested as follows:

\begin{eqnarray}
&\varepsilon(k) = \begin{cases}
rule 1:\phi(\mathbf{x}^{\theta}), \text{if } k = 0\\
rule 2:(1 - \tau)\varepsilon(k-1), \text{if } r_k < \alpha \text{ and } k < T_c\\
rule 3:(1 + \tau)\phi_{max}, \text{if } r_k \ge \alpha \text{ and } k < T_c \\
rule 4:0, \text{if } k \ge T_c\\
\end{cases}
\label{equ:improved_dynamic_epsilon}
\end{eqnarray}
where $\phi_k(\mathbf{x}^{\theta})$ is the overall constraint violation of the top $\theta$-th individual in the initial population, $r_k$ is the ratio of feasible solutions in the $k$-th generation. $\tau$ ranges between $0$ and $1$, and has two functions. One is to control the speed of reducing the relaxation of constraints, and the other is to control the scale factor multiplied by the maximum overall constraint violation. $\alpha$ is to control the searching preference between the feasible and infeasible regions, and $\alpha \in[0,1]$. $\phi_{max}$ is the maximum overall constraint violation found so far.

The $\varepsilon(0)$ setting method in Eq. \ref{equ:improved_dynamic_epsilon} is sometimes the same as that in Eq. \ref{equ:epsilon_setting}. If $\varepsilon(0) = 0$, $\varepsilon(k)$ in Eq. \ref{equ:epsilon_setting} is identically equal to zero, which tends to hinder a CMOEA's exploration of the infeasible regions. However, $\varepsilon(k)$ in Eq. \ref{equ:improved_dynamic_epsilon} is not identically equal to zero when $\varepsilon(0) = 0$ according to the third rule of the proposed epsilon setting approach.

In the case $k > 0$, three rules are adopted to control the value of $\varepsilon$ in Eq. \ref{equ:improved_dynamic_epsilon}. $Rule 2$ is adopted to strengthen the searching in the feasible regions. $Rule 3$ is used to strengthen the exploration in the infeasible regions. The last $rule 4$ is same as in the CDP (\cite{996017}) constraint-handling method.

Two parameters $k$ and $r_k$ are applied to choose the right control rule for $\varepsilon(k)$. If $k < T_c$ and $r_k < \alpha$, $rule 2$ for setting $\varepsilon(k)$ is adopted. In this circumstance, $\varepsilon(k)$ is set to $(1 - \tau)\varepsilon(k-1)$, which has an exponential decreasing rate. It has a faster descent rate than the epsilon setting in Eq. \eqref{equ:epsilon_setting}, which can help to enhance the searching in the feasible regions more effectively. If $k < T_c$ and $r_k \ge \alpha$, $rule 3$ for setting $\varepsilon(k)$ is applied. In this situation, most solutions are feasible. Thus, strengthening the exploration in the infeasible regions may help a CMOEA to get across a number of large infeasible regions. In $rule 3$, $\varepsilon(k) = (1 + \tau)\phi_{max}$, which strengthens the exploration in the infeasible regions. Thus, the improved epsilon method has the balanced ability to explore the feasible and infeasible regions simultaneously. 

$\alpha$ is a critical parameter to balance the searching between the feasible and infeasible regions. If the RFS $r_k$ is less than $\alpha$, $rule 2$ is adopted to enhance the exploration in the feasible regions. Otherwise, $rule 3$ is applied to enhance the exploration in the infeasible regions. Thus, the proposed epsilon constraint method can keep a good balance of exploration between the feasible and infeasible regions. It utilizes the RFS to dynamically balance the exploration between the feasible regions and infeasible regions. 

Compared with the $\varepsilon$ setting in Eq. \eqref{equ:epsilon_setting}, the proposed method in Eq. \eqref{equ:improved_dynamic_epsilon} has the ability to increase $\varepsilon(k)$ during the evolutionary process, which can help to solve CMOPs with large infeasible regions.   

In the case of $k \ge T_c$, $rule 4$ is applied. In this situation, $\varepsilon(k) = 0$, and the epsilon constraint-handling method exerts the highest selection pressure toward the feasible regions.

\subsection{Embedding the improved epsilon method in MOEA/D}

The proposed MOEA/D-IEpsilon integrates the improved epsilon constraint-handling method in Eq. \ref{equ:improved_dynamic_epsilon} into the framework of MOEA/D. In MOEA/D-IEpsilon, a CMOP is decomposed into a number of constrained scalar subproblems, and these subproblems are optimized simultaneously in a collaborative way. In our experimental studies, the Tchebycheff approach is adopted, and its detailed definition is listed in Eq. \eqref{equ:tchmethod-a}.

For a given weight vector $\lambda$, there exists an optimal solution of Eq. \eqref{equ:tchmethod-a}, and this optimal solution is also a Pareto optimal solution of Eq. \eqref{equ:cmop_definition}. Therefore, we can achieve different Pareto optimal solutions of Eq. \eqref{equ:cmop_definition} by setting different weight vectors.

\begin{algorithm}
    \KwIn{\\
    $N$: the number of subproblems.\\
    $T_{max}$: the maximum generation.\\
    $N$ weight vectors: $\mathbf{\lambda}^1,\ldots,\mathbf{\lambda}^N$.\\
    $T$: the size of the neighborhood.\\
    $\delta$: the selecting probability from neighbors.\\
    $n_r$: the maximal number of solutions replaced by a child.\\
    }
    \KwOut{$NS:$ a set of feasible non-dominated solutions}

    Decompose a CMOP into $N$ subproblems associated with $\mathbf{\lambda}^1,\ldots,\mathbf{\lambda}^N$.\\
    Generate an initial population $P=\{\mathbf{x}^1, \ldots, \mathbf{x}^N \}$.\\
    Initialize $\varepsilon(0)$ according to Eq. \eqref{equ:improved_dynamic_epsilon}. \\
    Initialize the ideal point $z^*=(z_1,\ldots,z_m)$.\\
    For each $i = 1, \dots, N$, set $B(i) = \{i_1,\dots,i_T\}$, where $\mathbf{\lambda}^{i_1},\dots,\mathbf{\lambda}^{i_T}$ are the $T$ closest weight vectors to $\mathbf{\lambda}^i$.\\
    $k = 1$.\\
    \While{$k \le T_{max}$}{
    Set $\varepsilon(k)$ according to Eq. \eqref{equ:improved_dynamic_epsilon}.\\
    Generate a random permutation $rp$ from $\{1,\ldots,N\}$.\\
    \For{$i \leftarrow 1$ \KwTo $N$}{
    Generate a random number $r\in[0,1]$.\\
    $j = rp(i)$.\\
    \eIf{$r < \delta$}{
    $S = B(j)$
    }{
    $ S = \{1,\ldots,N\}$
    }
    Generate $\mathbf{y}^j$ through the DE operator.\\
    Perform polynomial mutation on $\mathbf{y}^j$.\\

    \For{$t \leftarrow 1$ \KwTo $m$}{
    \lIf{$z^*_t > f_t(\mathbf{y}^j)$}{
        $z^*_t  = f_t(\mathbf{y}^j)$
    }
    }
    Set $c = 0$.\\
    \While{$c \neq n_r$ or $S \neq \varnothing$}{
     select an index $j$ from $S$ randomly.\\
     $result$ = UpdateSubproblems($\mathbf{x}^j$, $\mathbf{y}^j$, $\varepsilon(k)$)\\
     \lIf{$result == true$}{
     $c = c+1$}
     $S = S \backslash\{j\}$
    }

    }
    $k = k + 1$\\
    Update $\varepsilon(0)$ according to Eq. \eqref{equ:improved_dynamic_epsilon}\\
    $NS$ = NondominatedSelect($NS  \bigcup P$)
    }
\caption{MOEA/D-IEpsilon}
\label{alg:moead-iepsilon}
\end{algorithm}

The psuecode of MOEA/D-IEpsilon is listed in Algorithm 1. It is almost the same as that of MOEA/D, except for the method of subproblem updating. Lines 1-6 initialize a number of parameters in MOEA/D-IEpsilon. First, a CMOP is decomposed into $N$ subproblems which are associated with $\lambda^1,\ldots, \lambda^N$. Then the population $P$, the initial epsilon value $\varepsilon(0)$, the ideal point $z^{*}$ and the neighbor indexes $B(i)$ are initialized.

Lines 11-22 generate a set of new solutions and update the ideal point $z^{*}$. To be more specific, a set of solutions which may be updated by a newly generated solution $\mathbf{y}^j$ is selected (lines 11-17). In line 18, the differential evolution (DE) crossover is adopted to generate a new solution $\mathbf{y}^j$. The polynomial mutation operator is executed to mutate $\mathbf{y}^j$ in line 19. The ideal point $z^{*}$ is updated (lines 20-22).

Lines 23-30 implement the updating process of subproblems. In line 26, the subproblems are updated based on the improved epsilon constraint-handling approach, and the detailed procedures are listed in Algorithm 2. Finally, a set of non-dominated solutions ($NS$) is selected based on the non-dominated sort in line 33.

\begin{algorithm}
\Fn{result = UpdateSubproblems($\mathbf{x}^j$,$\mathbf{y}^j$,$\varepsilon(k)$)}{
    $result = false$

    \uIf{$\phi(\mathbf{y}^j) \le \varepsilon(k)$ and $\phi(\mathbf{x}^j) \le \varepsilon(k)$}{
        \If{$g^{te}(\mathbf{y}^i|\lambda^j,z^{*}) \leq g^{te}(\mathbf{x}^j|\lambda^j,z^{*})$}{
            $\mathbf{x}^j$ = $\mathbf{y}^j$\\
            $result = ture$
        }
    }
   \uElseIf{$\phi(\mathbf{y}^j) == \phi(\mathbf{x}^j)$}{
        \If{$g^{te}(\mathbf{y}^j|\lambda^j,z^{*}) \leq g^{te}(\mathbf{x}^j|\lambda^j,z^{*})$}{
            $\mathbf{x}^j$ = $\mathbf{y}^j$\\
            $result = ture$
        }
   }
   \ElseIf{$\phi(\mathbf{y}^j) < \phi(\mathbf{x}^j)$}{
        $\mathbf{x}^j$ = $\mathbf{y}^j$\\
        $result = ture$
    }

    \Return $result$
}
\caption{Subproblem Update}
\end{algorithm}

In Algorithm 2, there are three basic rules to update a subproblem. For two solutions $\mathbf{x}^j$ and $\mathbf{y}^j$, if their overall constraint violations are less than or equal to $\varepsilon(k)$, and $\mathbf{y}^j$ has a smaller aggregation value (the value of the decomposition function) than that of $\mathbf{x}^j$, then $\mathbf{x}^j$ is replaced by $\mathbf{y}^j$ (lines 3-7). If $\mathbf{x}^j$ and $\mathbf{y}^j$ have the same overall constraint violation, and $\mathbf{y}^j$ has a smaller aggregation value than that of $\mathbf{x}^j$, then $\mathbf{x}^j$ is replaced by $\mathbf{y}^j$ (lines 8-12). Otherwise, if $\mathbf{y}^j$ has a smaller overall constraint violation than that of $\mathbf{x}^j$, then $\mathbf{x}^j$ is replaced by $\mathbf{y}^j$ (lines 13-14). When the subproblem is updated, the function $UpdateSubproblems(\mathbf{x}^j,\mathbf{y}^j,\varepsilon(k))$ returns $true$, otherwise, it returns $false$. 

\section{Test instances}
\label{sec:4}

To evaluate the performance of the proposed MOEA/D-IEpsilon, a set of new CMOPs with large infeasible regions (named LIR-CMOPs) is designed according to our previous work (\cite{fan2016difficulty}). In terms of constraint functions, all of them have large infeasible regions. In term of objective functions, there are two components:  shape functions and distance functions (\cite{Huband:2006hi}).

The shape functions are applied to set the shape of the PFs. In the LIR-CMOP test suite, two types of shape functions, including both convex and concave shapes, are designed. Distance functions are adopted that test the convergence performance of a CMOEA. In LIR-CMOP5-14, the distance functions are multiplied by a scale factor, which increases difficulty of convergence. The detailed definitions of LIR-CMOPs are listed in the Appendix.  

\begin{figure*}
\begin{tabular}{cc}
\begin{minipage}[t]{0.33\linewidth}
\includegraphics[width = 5.5cm]{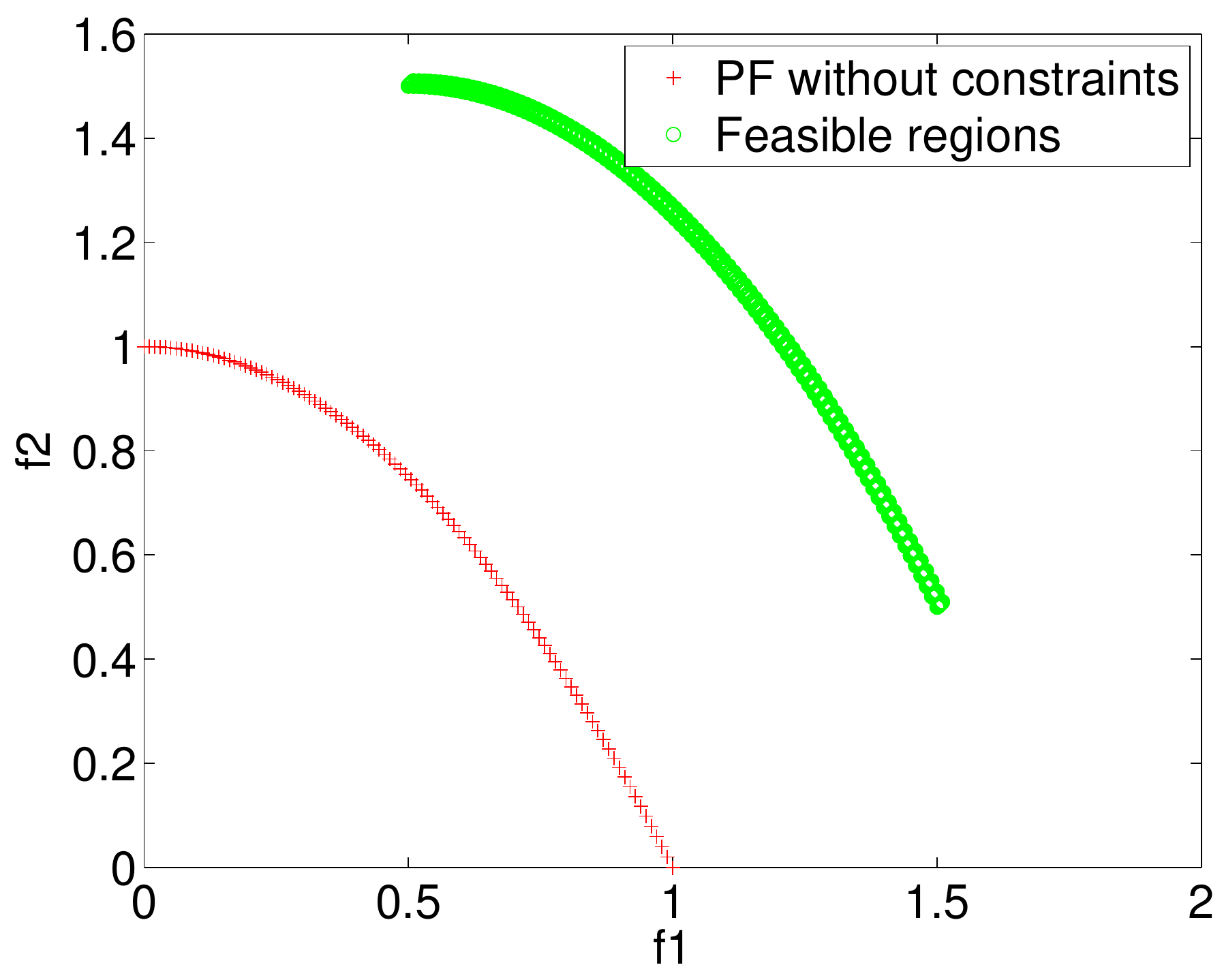}\\
\centering{\scriptsize{(a) LIR-CMOP1}}
\end{minipage}
\begin{minipage}[t]{0.33\linewidth}
\includegraphics[width = 5.5cm]{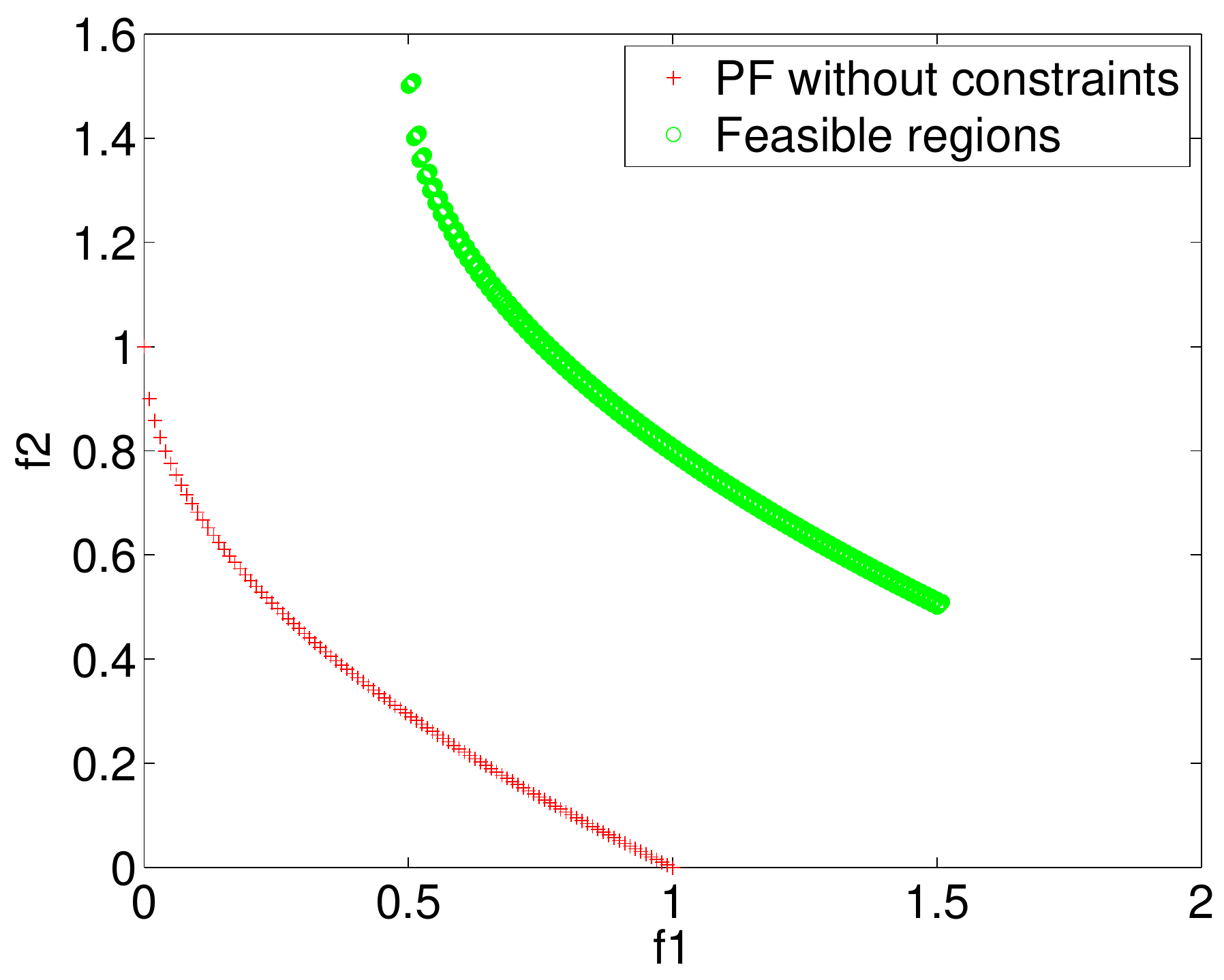}\\
\centering{\scriptsize{(b) LIR-CMOP2}}
\end{minipage}
\begin{minipage}[t]{0.33\linewidth}
\includegraphics[width = 5.5cm]{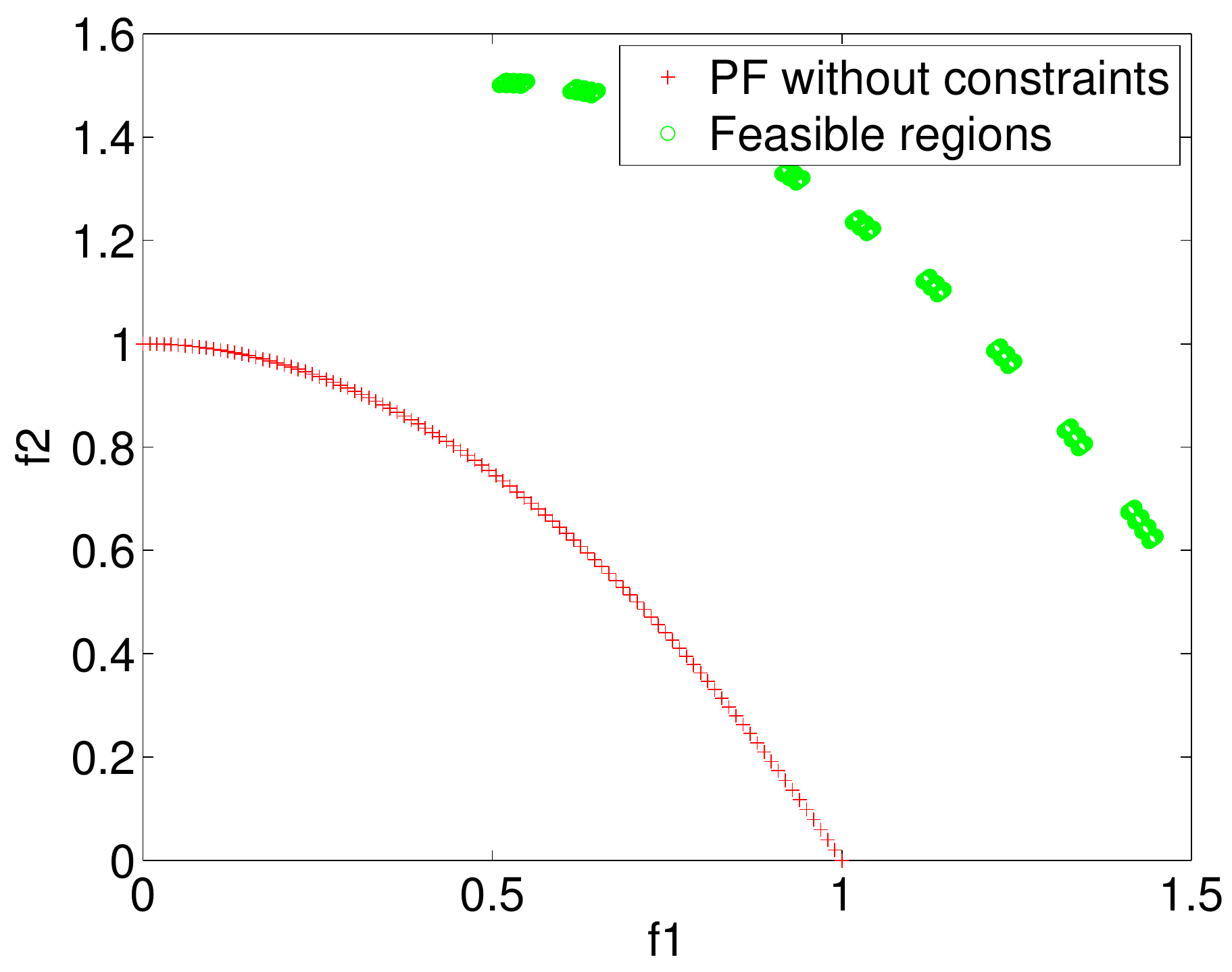}\\
\centering{\scriptsize{(c) LIR-CMOP3}}
\end{minipage}
\end{tabular}

\begin{tabular}{cc}
\begin{minipage}[t]{0.33\linewidth}
\includegraphics[width = 5.5cm]{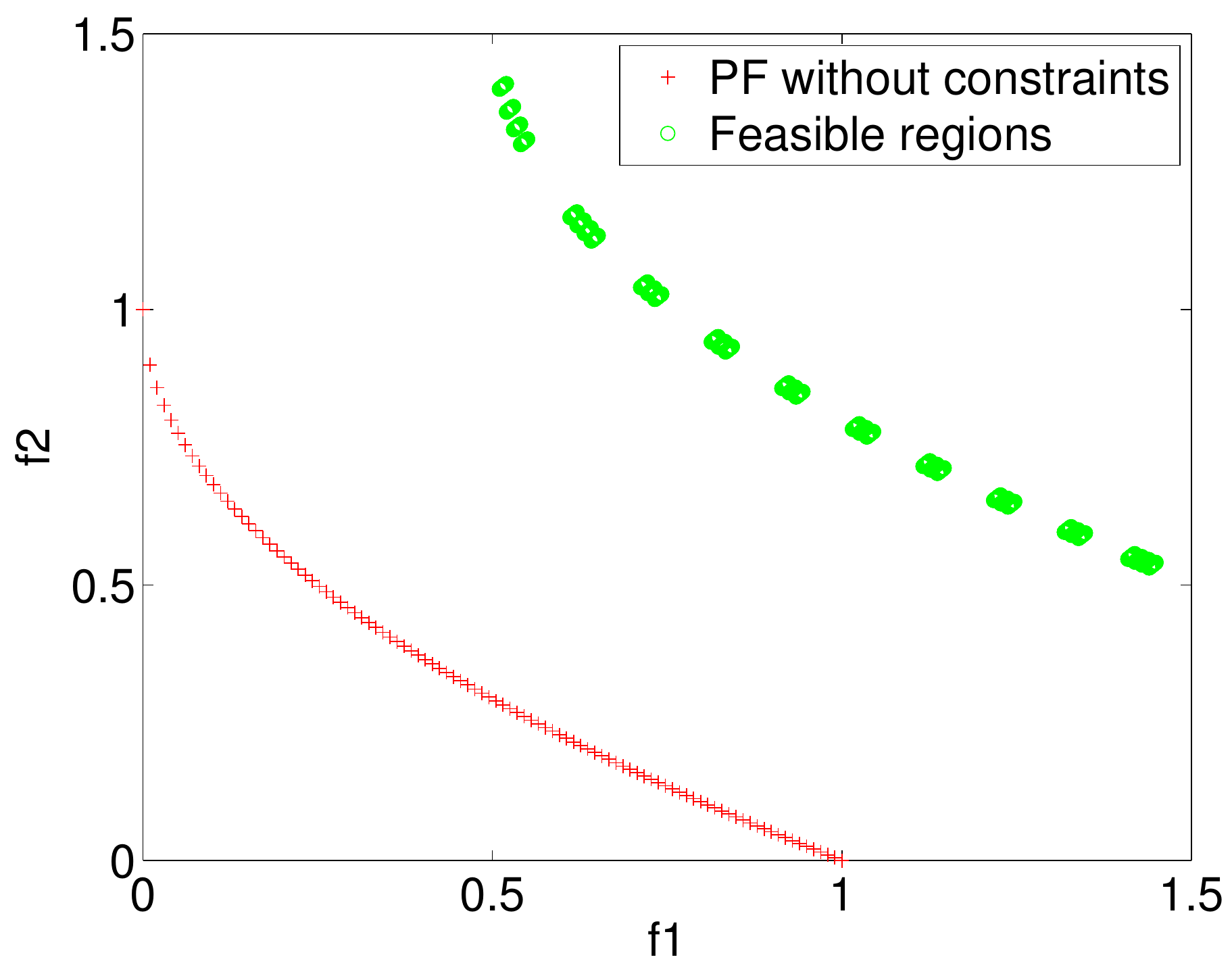}\\
\centering{\scriptsize{(d) LIR-CMOP4}}
\end{minipage}
\begin{minipage}[t]{0.33\linewidth}
\includegraphics[width = 5.5cm]{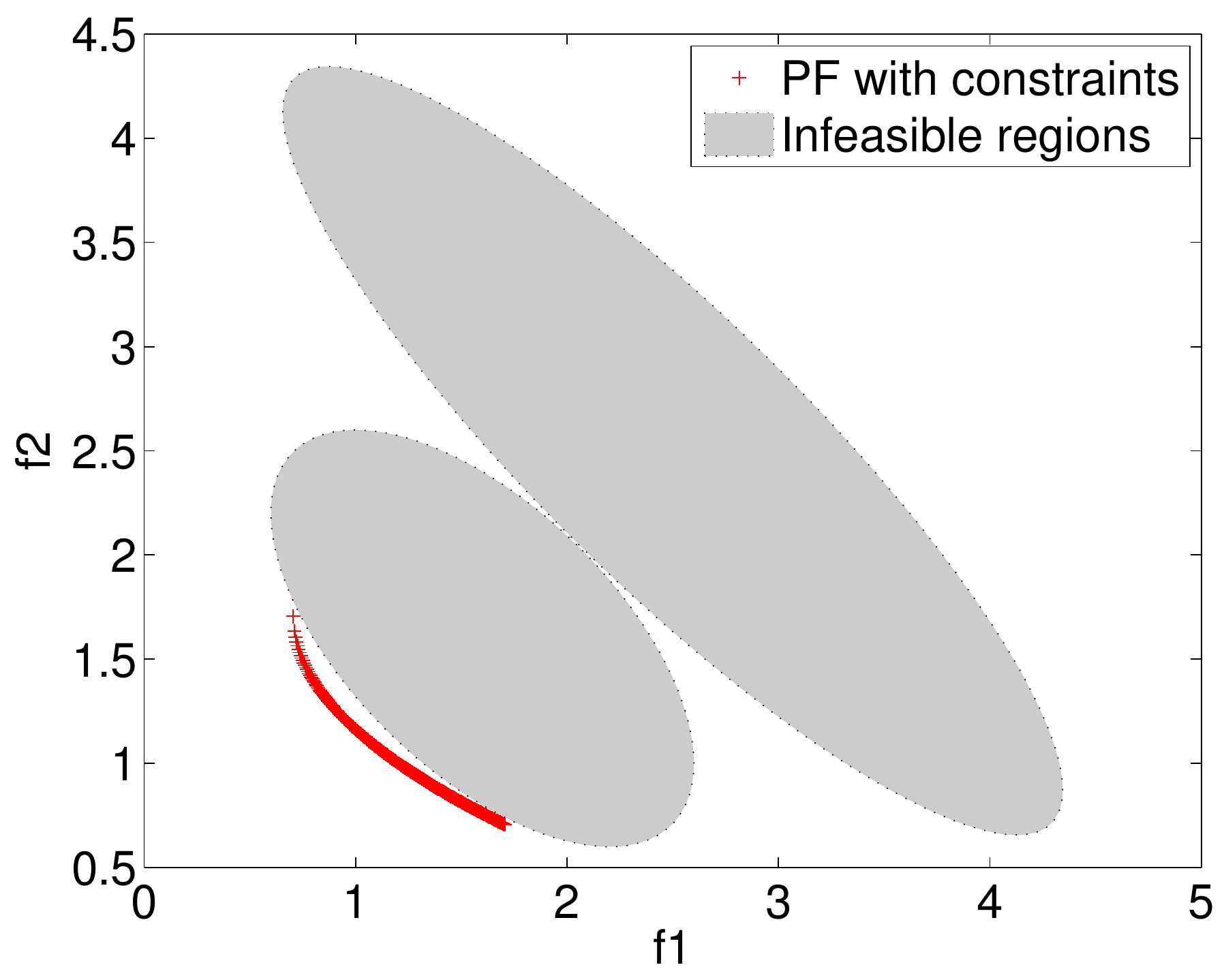}\\
\centering{\scriptsize{(e) LIR-CMOP5}}
\end{minipage}
\begin{minipage}[t]{0.33\linewidth}
\includegraphics[width = 5.5cm]{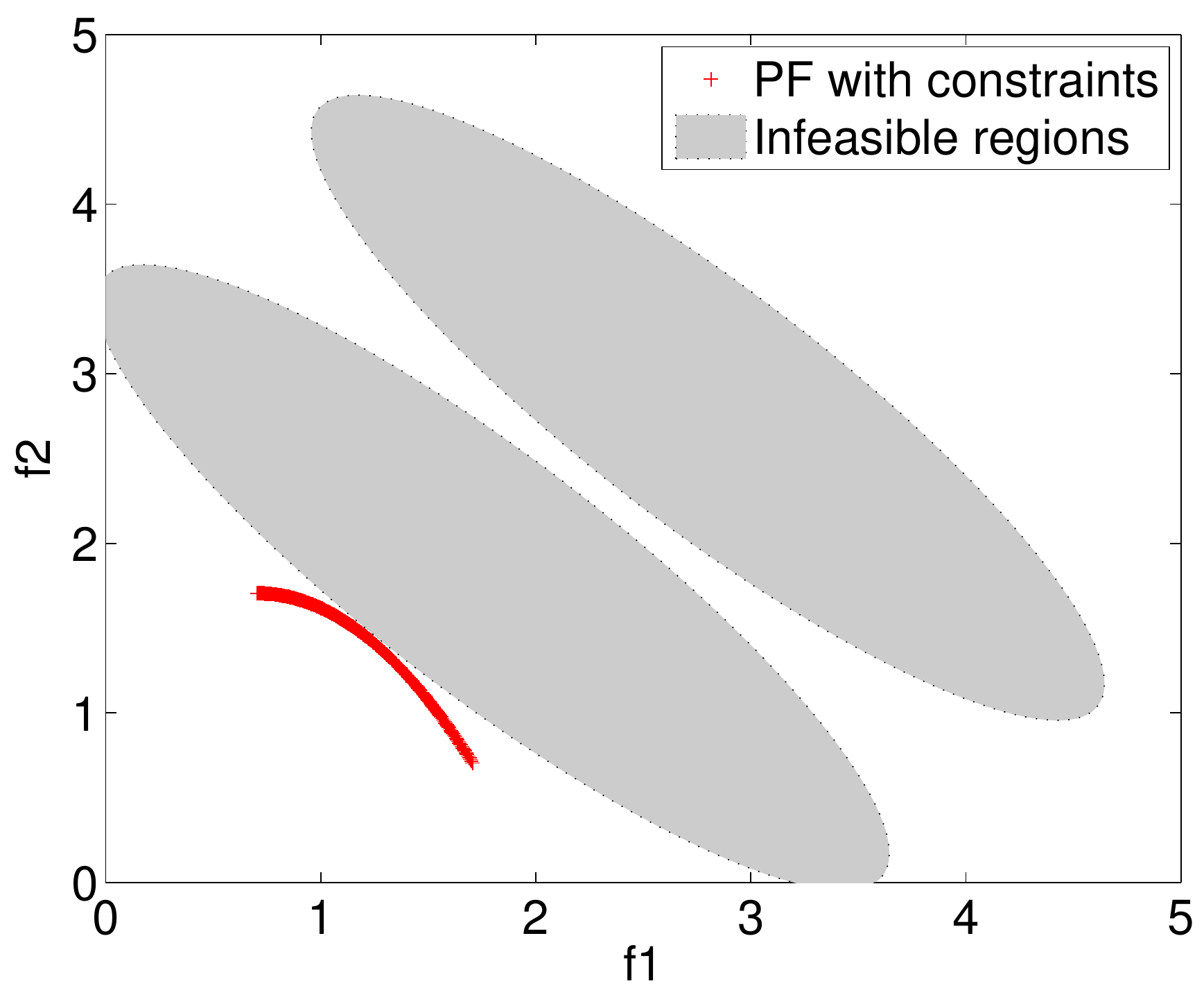}\\
\centering{\scriptsize{(f) LIR-CMOP6}}
\end{minipage}
\end{tabular}

\begin{tabular}{cc}
\begin{minipage}[t]{0.33\linewidth}
\includegraphics[width = 5.5cm]{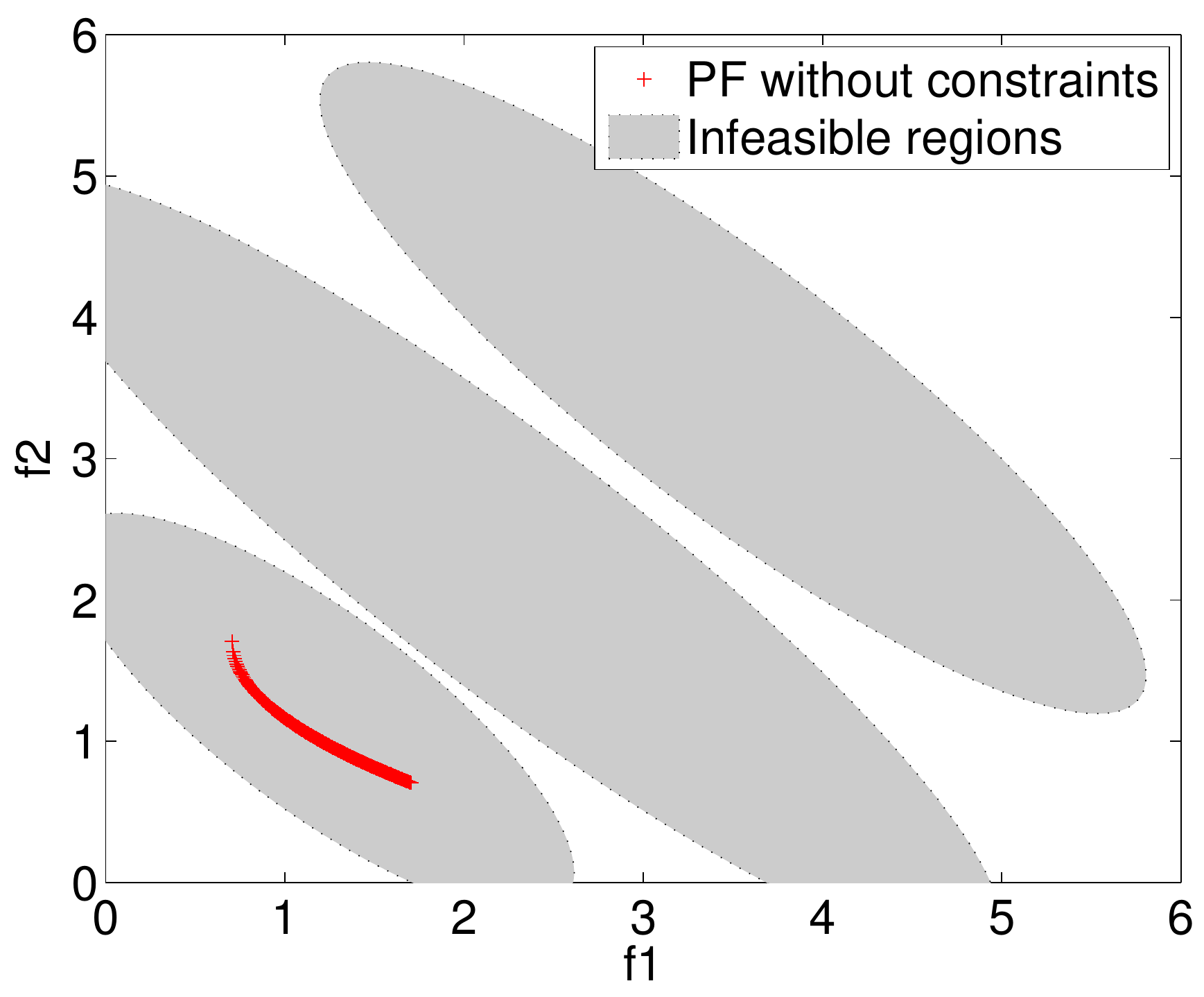}\\
\centering{\scriptsize{(g) LIR-CMOP7}}
\end{minipage}
\begin{minipage}[t]{0.33\linewidth}
\includegraphics[width = 5.5cm]{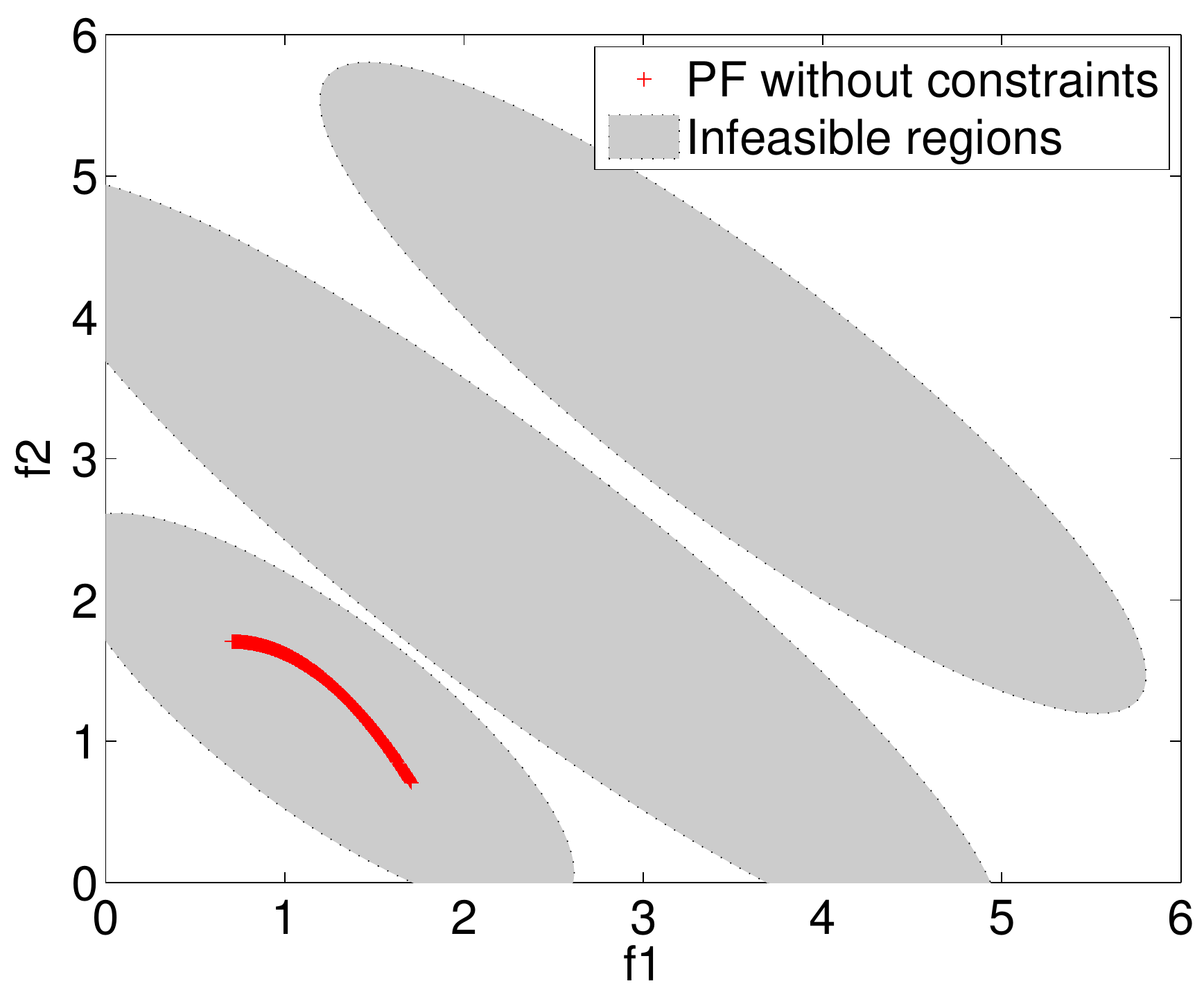}\\
\centering{\scriptsize{(h) LIR-CMOP8}}
\end{minipage}
\begin{minipage}[t]{0.33\linewidth}
\includegraphics[width = 5.5cm]{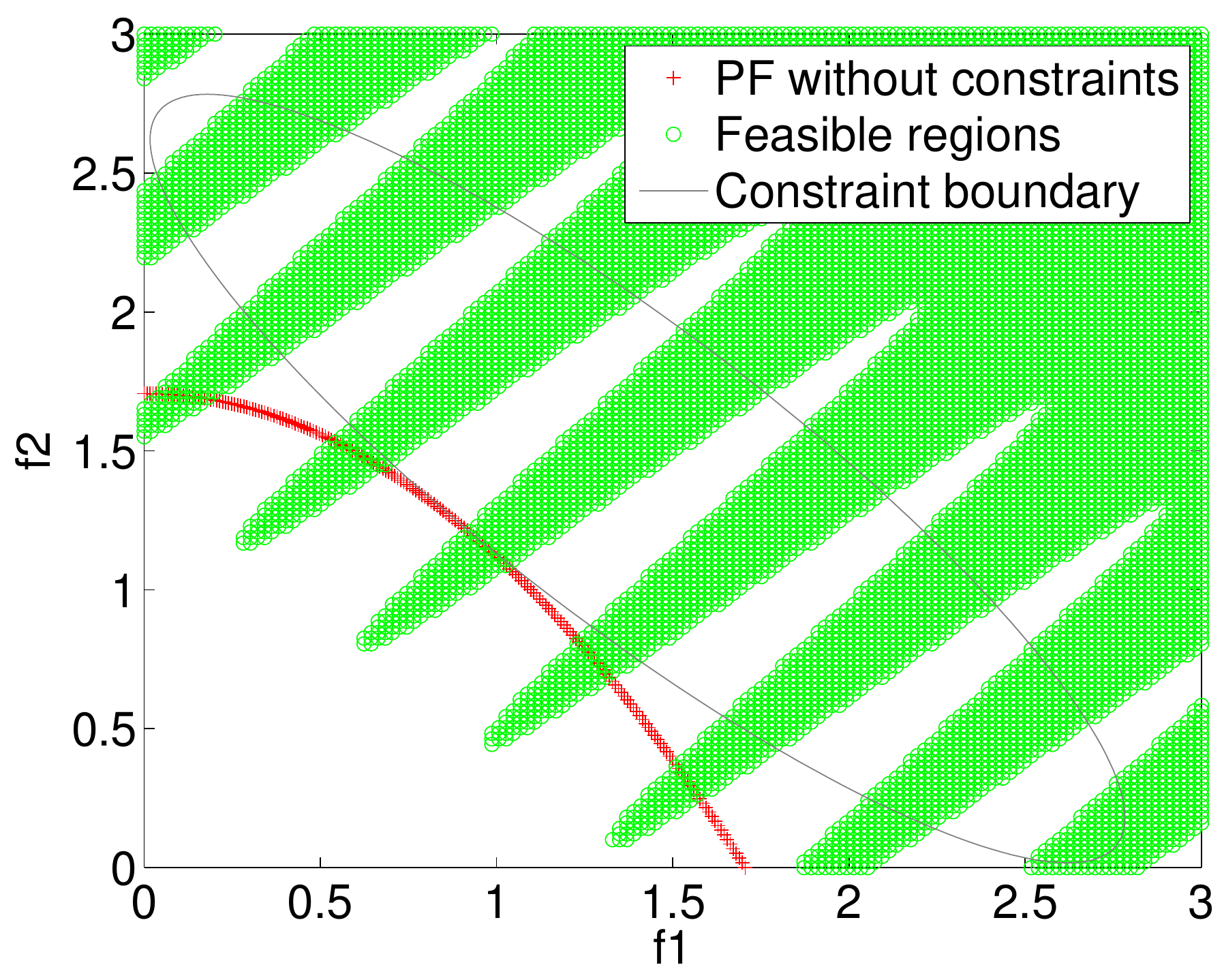}\\
\centering{\scriptsize{(i) LIR-CMOP9}}
\end{minipage}
\end{tabular}

\begin{tabular}{cc}
\begin{minipage}[t]{0.33\linewidth}
\includegraphics[width = 5.5cm]{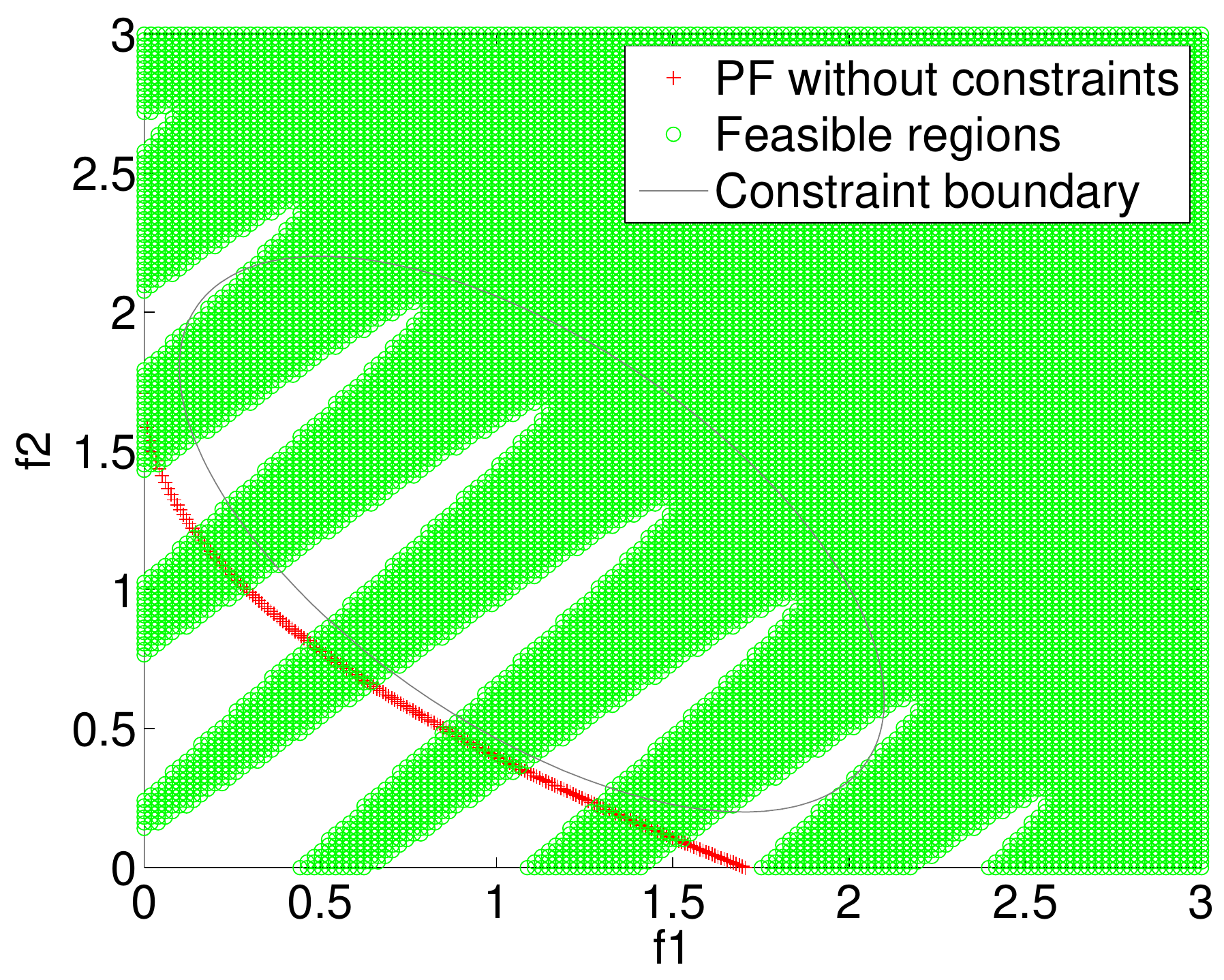}\\
\centering{\scriptsize{(j) LIR-CMOP10}}
\end{minipage}
\begin{minipage}[t]{0.33\linewidth}
\includegraphics[width = 5.5cm]{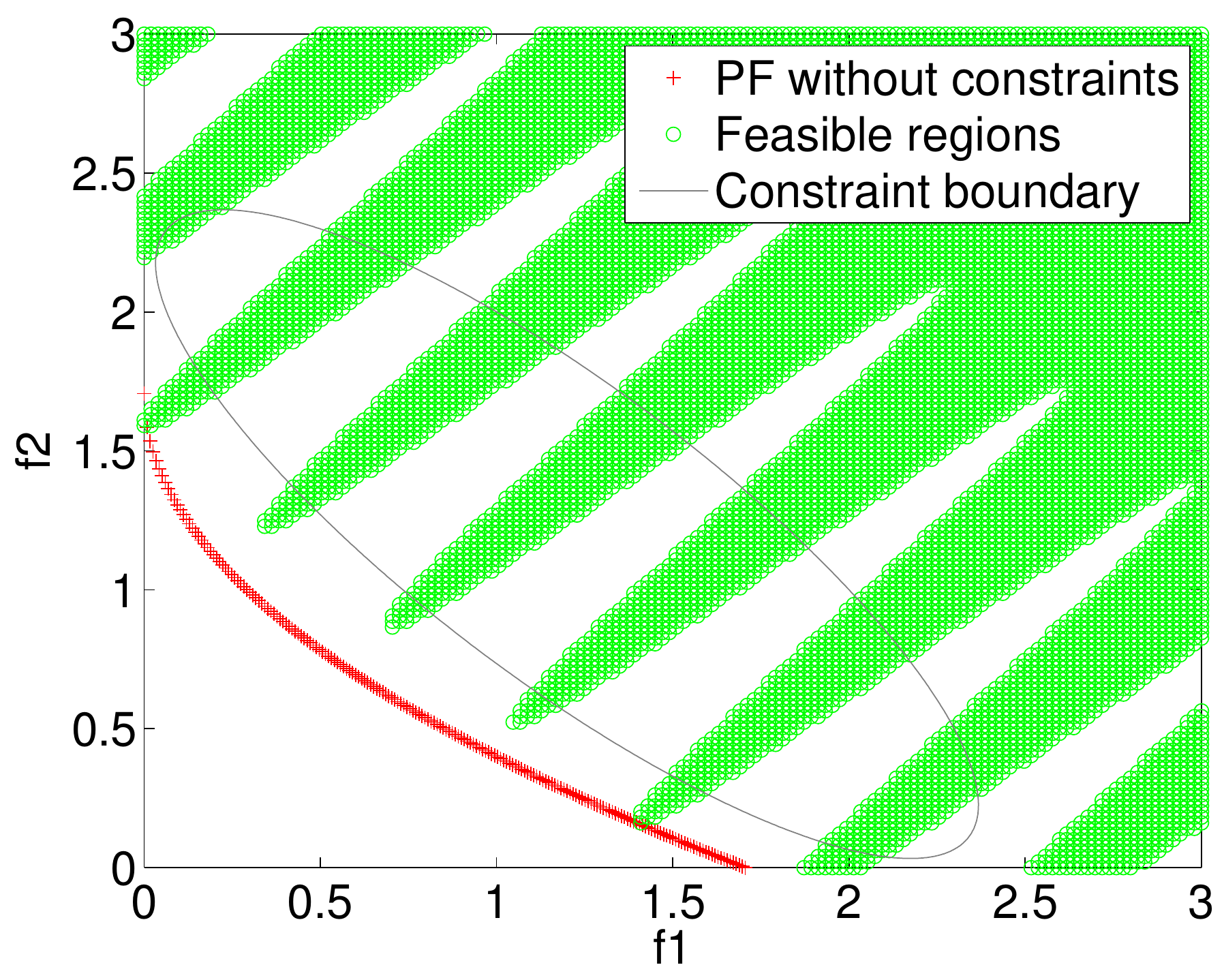}\\
\centering{\scriptsize{(k) LIR-CMOP11}}
\end{minipage}
\begin{minipage}[t]{0.33\linewidth}
\includegraphics[width = 5.5cm]{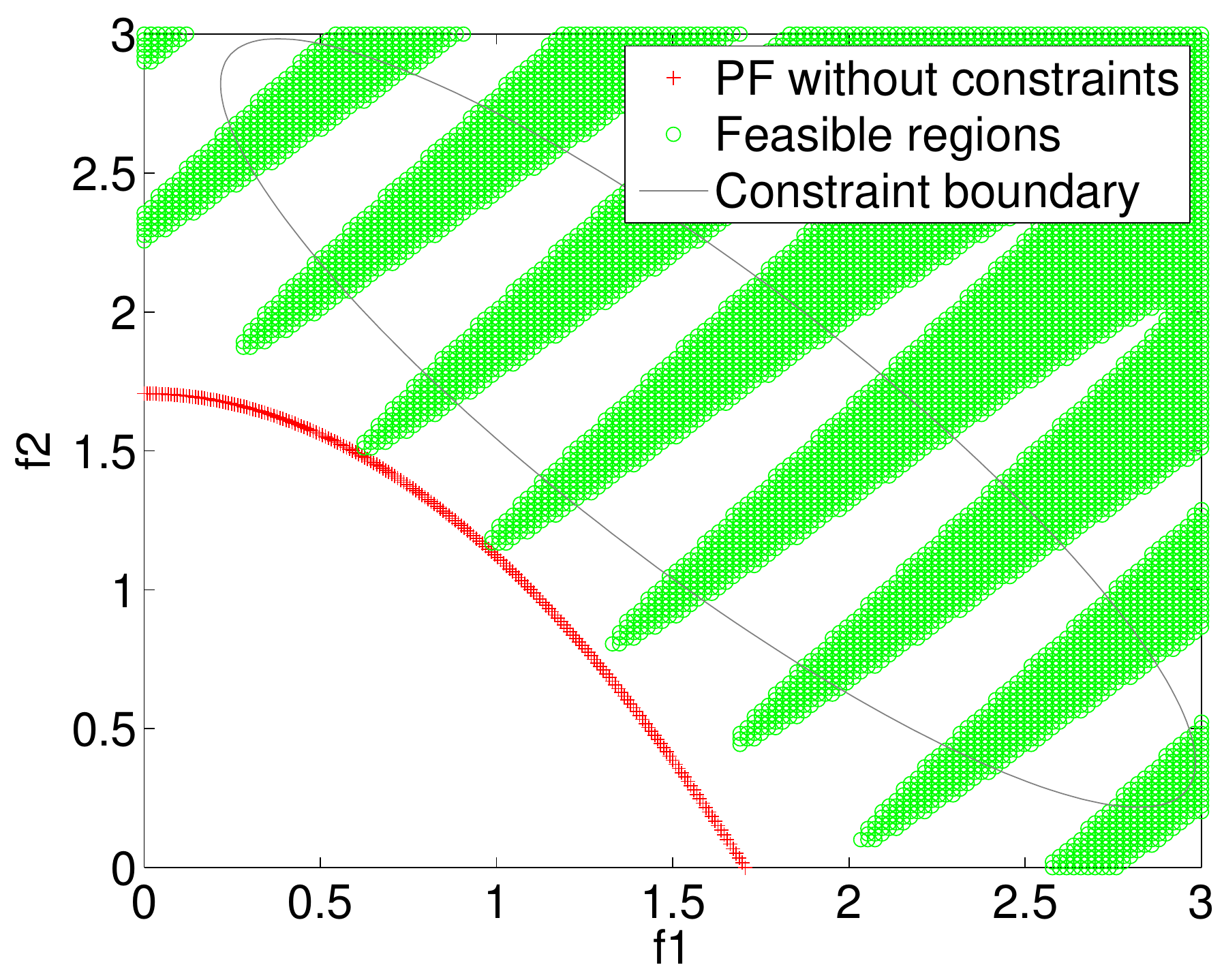}\\
\centering{\scriptsize{(l) LIR-CMOP12}}
\end{minipage}
\end{tabular}
\caption{Illustrations of the feasible and infeasible regions of LIR-CMOP1-12.} \label{Fig:lir-cmop-a}
\end{figure*}

In this test suite, four test problems, including LIR-CMOP1-4, have large infeasible regions. Fig. \ref{Fig:lir-cmop-a}(a)-(d) plot the feasible regions of LIR-CMOP1-4, respectively. It can be seen that the feasible regions of these test instances are very small. In other words, there are a number of large infeasible regions.  

LIR-CMOP5 and LIR-CMOP6 have convex and concave PFs, respectively, as shown in Fig. \ref{Fig:lir-cmop-a}(e)-(f) , and their PFs are the same as those of their unconstrained counterparts. The PFs of LIR-CMOP5 and LIR-CMOP6 can be achieved by a MOEA without any constraint-handling mechanisms.

In order to expand the test scope, LIR-CMOP7 and LIR-CMOP8 are designed. For these two test instance, their unconstrained PFs are located in the infeasible regions, and their PFs are situated on their constraint boundaries. Thus, a MOEA without constraint-handling methods cannot find the real PFs for LIR-CMOP7 and LIR-CMOP8, which are shown in Fig. \ref{Fig:lir-cmop-a}(g)-(h).

LIR-CMOP9-12 have two different types of constraints. The first type creates large infeasible regions as shown in the black ellipses in Fig. \ref{Fig:lir-cmop-a}(i)-(l). The second type creates difficulty in the entire objective space, as it divides the PFs of LIR-CMOP9-12 into a number of disconnected segments. For LIR-CMOP9-10, their PFs are a part of their unconstrained PFs, and for LIR-CMOP11-12, their PFs are situated on their constraint boundaries.

\begin{figure*}
\begin{tabular}{cc}
\begin{minipage}[t]{0.5\linewidth}
\includegraphics[width = 8cm]{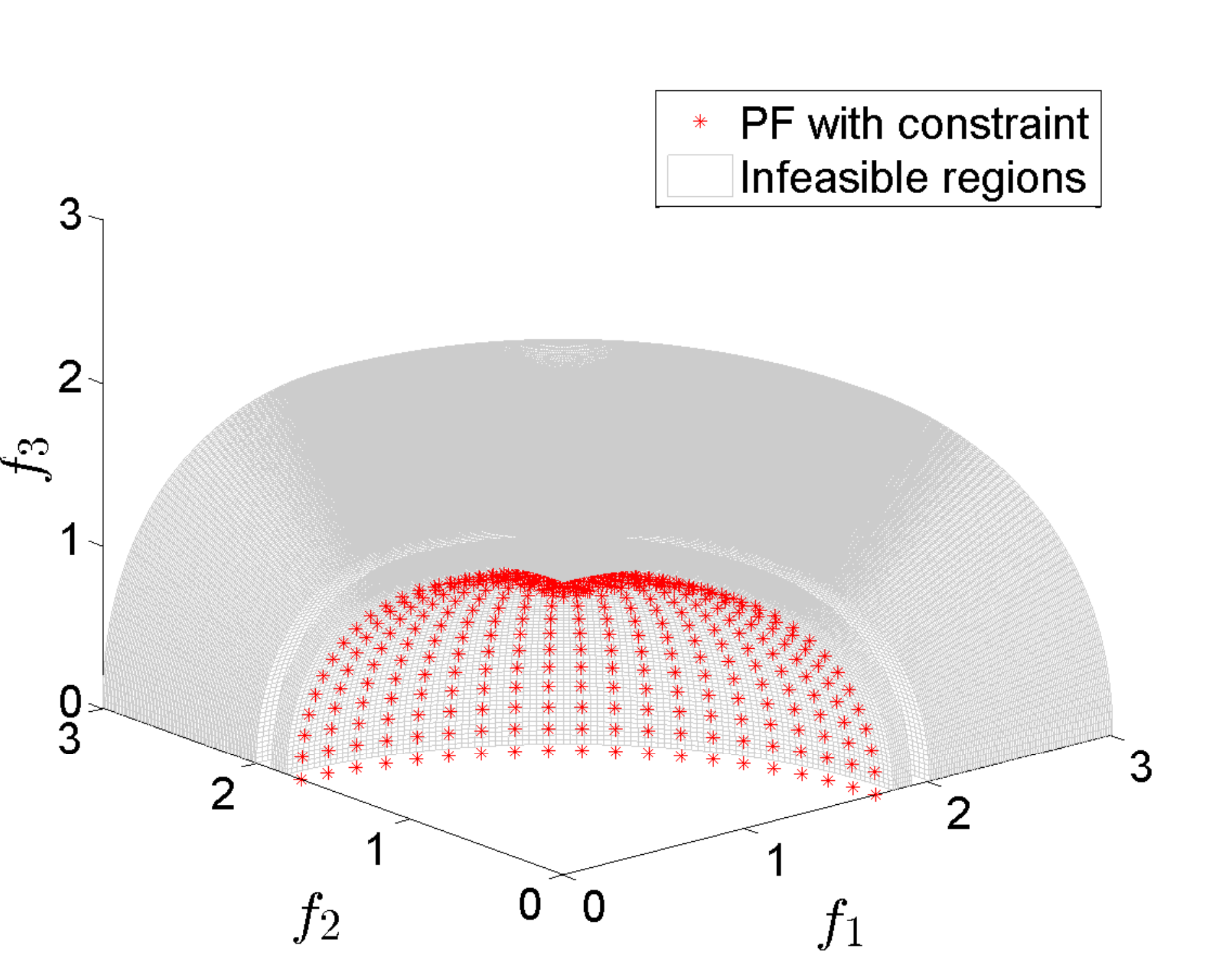}\\
\centering{\scriptsize{(a) LIR-CMOP13}}
\end{minipage}
\begin{minipage}[t]{0.5\linewidth}
\includegraphics[width = 8cm]{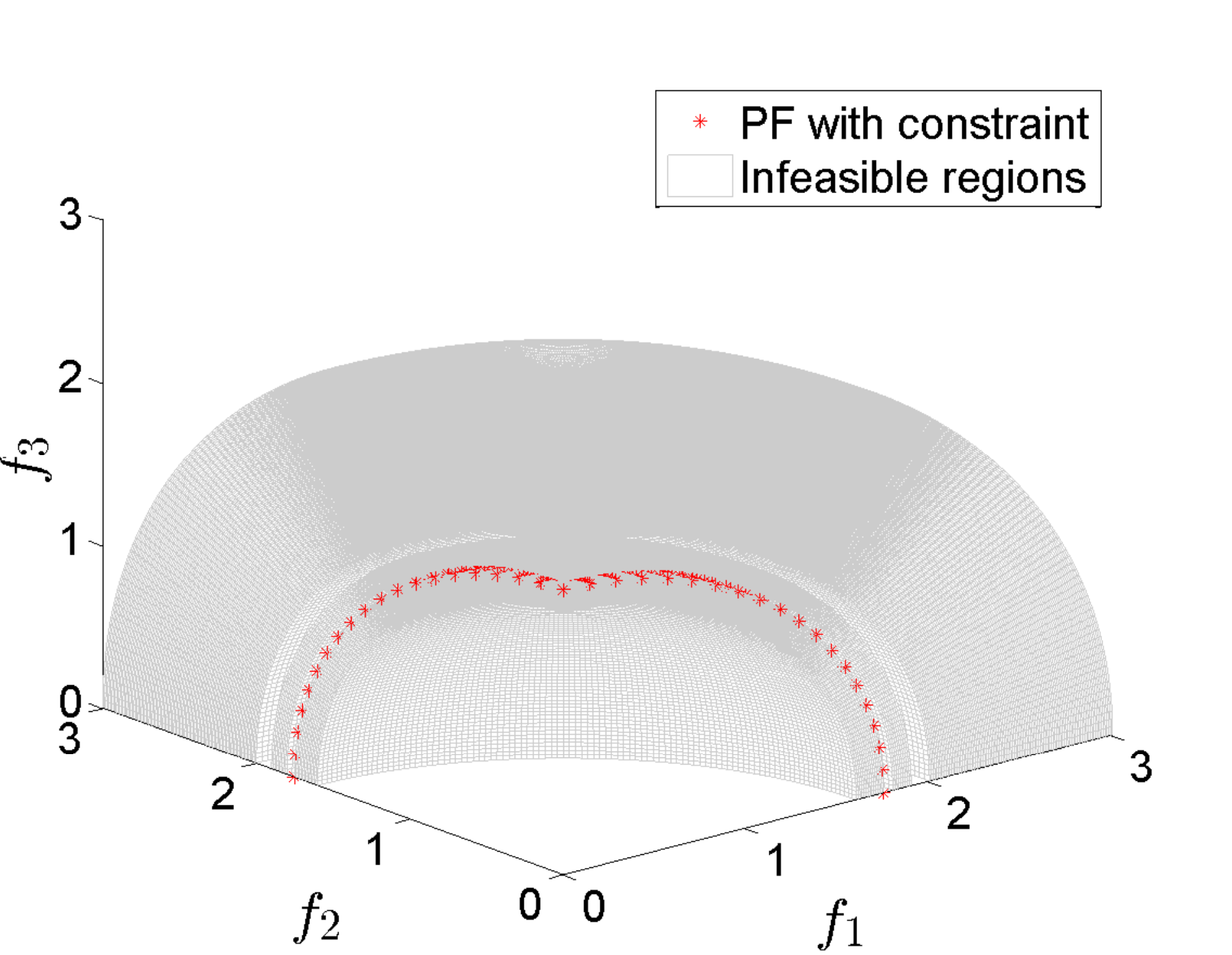}\\
\centering{\scriptsize{(b) LIR-CMOP14}}
\end{minipage}
\end{tabular}
\caption{Illustrations of the infeasible regions of LIR-CMOP13-14.} \label{Fig:lir-cmop-b}
\end{figure*}

In the LIR-CMOP test suite, CMOPs with three objectives are also designed. Two CMOPs, including LIR-CMOP13 and LIR-CMOP14, have three objectives as shown in Fig. \ref{Fig:lir-cmop-b} (a)-(b) . The PF of LIR-CMOP13 is the same as that of its unconstrained counterpart. The PF of LIR-CMOP14 is located on the boundaries of its constraints.

\section{Experimental study}
\label{sec:5}

\subsection{Experimental Settings}
\label{sec:5.1}
To evaluate the performance of the proposed MOEA/D-IEpsilon, four other CMOEAs (MOEA/D-Epsilon, MOEA/D-SR, MOEA/D-CDP and C-MOEA/D), with differential evolution (DE) crossover, are tested on LIR-CMOP1-14. The detailed parameters of these five CMOEAs are listed as follows:

\begin{enumerate}
\item Mutation probability $Pm = 1/n$ ($n$ is the number of decision variables) and its distribution index is set to 20. $CR = 1.0$, $f = 0.5$.
\item Population size: $N = 300$. Neighborhood size: $T = 30$.
\item Stopping condition: each algorithm runs for 30 times independently, and stops when 300,000 function evaluations are reached.
\item Probability of selecting individuals in the neighborhood: $\delta = 0.9$.
\item The maximal number of solutions replaced by a child: $nr = 2$.
\item Parameter setting in MOEA/D-IEpsilon: $T_c = 800$, $\alpha = 0.95$, $\tau = 0.1$ and $\theta = 0.05 N$.
\item Parameter setting in MOEA/D-Epsilon: $T_c = 800$, $cp = 2$, and $\theta = 0.05 N$. 
\item Parameter setting in MOEA/D-SR: $S_r = 0.05$.
\end{enumerate}

\subsection{Performance Metric}
To measure the performance of MOEA/D-IEpsilon, C-MOEA/D, MOEA/D-CDP, MOEA/D-SR and MOEA/D-Epsilon, two commonly used metrics--the inverted generation distance (IGD) (\cite{bosman2003balance}) and the hypervolume (\cite {Zitzler1999Multiobjective}) are adopted. The definition of IGD is shown next.

\begin{itemize}
\item \textbf{Inverted Generational Distance} (IGD):
\end{itemize}
The IGD metric reflects the performance regarding convergence and diversity simultaneously. The detailed definition is as follows:
\begin{equation} \label{IGD metric}
\begin{cases}
IGD(P^*,A) = \frac{\sum \limits_{y^* \in P^*}d(y^*,A)}{| P^* |}\\
\\
d(y^*,A) = \min \limits_{y \in A} \{ \sqrt {\sum_{i = 1} ^m (y^{*}_{i} - y_i)^2} \}
\end{cases}
\end{equation}
where $P^*$ is a set of representative solutions in the ideal PF, $A$ is an approximate PF achieved by a CMOEA. The value of IGD denotes the distance between $P^*$ and $A$. For CMOPs with two objectives, 1000 points are sampled uniformly from the true PF to construct $P^*$. (Note that this measure cannot be used if the true Pareto front is unknown, so it is used primarily for benchmarking purposes.) For CMOPs with three objectives, 10000 points are sampled uniformly from the PF to constitute $P^*$. It is worth noting that a smaller value of IGD represents better performance with regards to both diversity and convergence.

\begin{itemize}
\item \textbf{Hypervolume} ($HV$):
\end{itemize}
$HV$ reflects the closeness of the non-dominated set achieved by a CMOEA to the real PF. The larger $HV$ means that the corresponding non-dominated set is closer to the true PF. 

\begin{equation}
HV(S)=VOL(\bigcup \limits_{x\in S} [f_1(x),z_1^r]\times ...[f_m(x),z_m^r] ) \\
\end{equation}
where $VOL(\cdot)$ is the Lebesgue measure, $\mathbf{z}^r=(z_1^r,...,z_m^r)^T$ is a reference point in the objective space. For a LIR-CMOP, the reference point is placed at 1.2 times the distance to the nadir point of the true PF. It is worth noting that a larger value of $HV$ represents better performance regarding both diversity and convergence.

 % In the robot gripper optimization case, $z^r = [5,800]^T$.

\subsection{Discussion of Experiments}
\subsubsection{Performance comparison on LIR-CMOP test suite}
\begin{figure*}
\begin{tabular}{cc}
\begin{minipage}[t]{0.25\linewidth}
\includegraphics[width = 4.5cm]{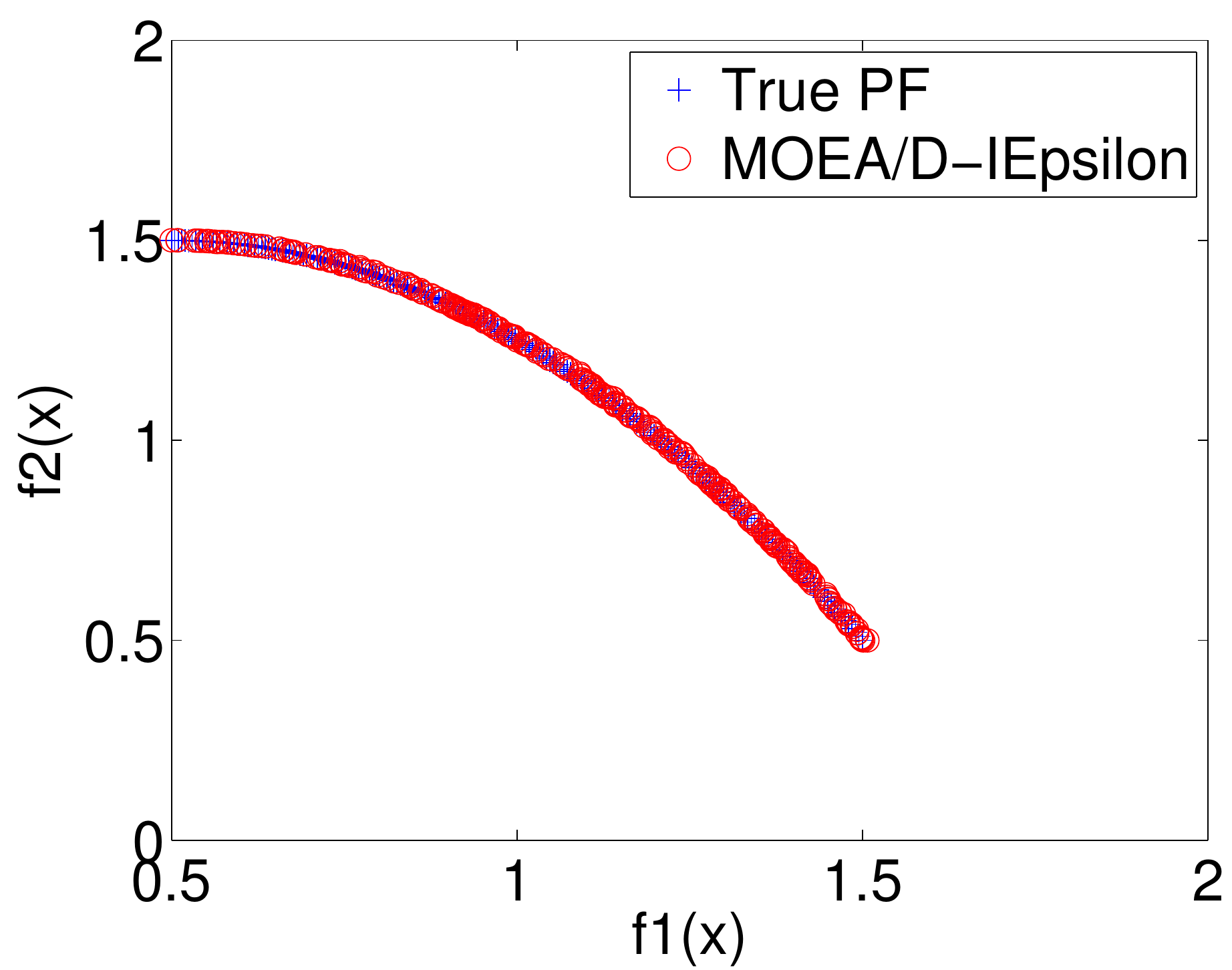}\\
\end{minipage}
\begin{minipage}[t]{0.25\linewidth}
\includegraphics[width = 4.5cm]{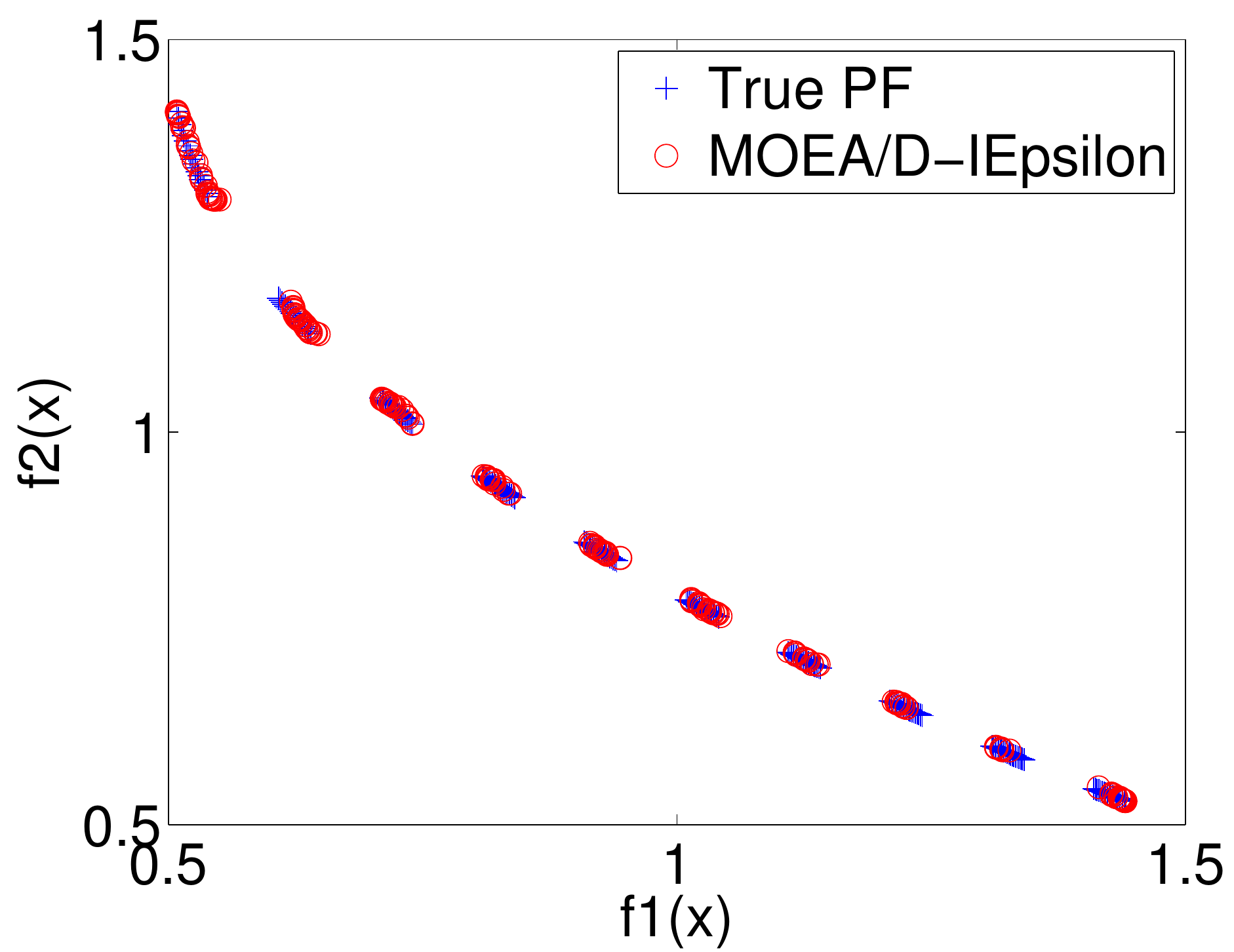}\\
\end{minipage}
\begin{minipage}[t]{0.25\linewidth}
\includegraphics[width = 4.5cm]{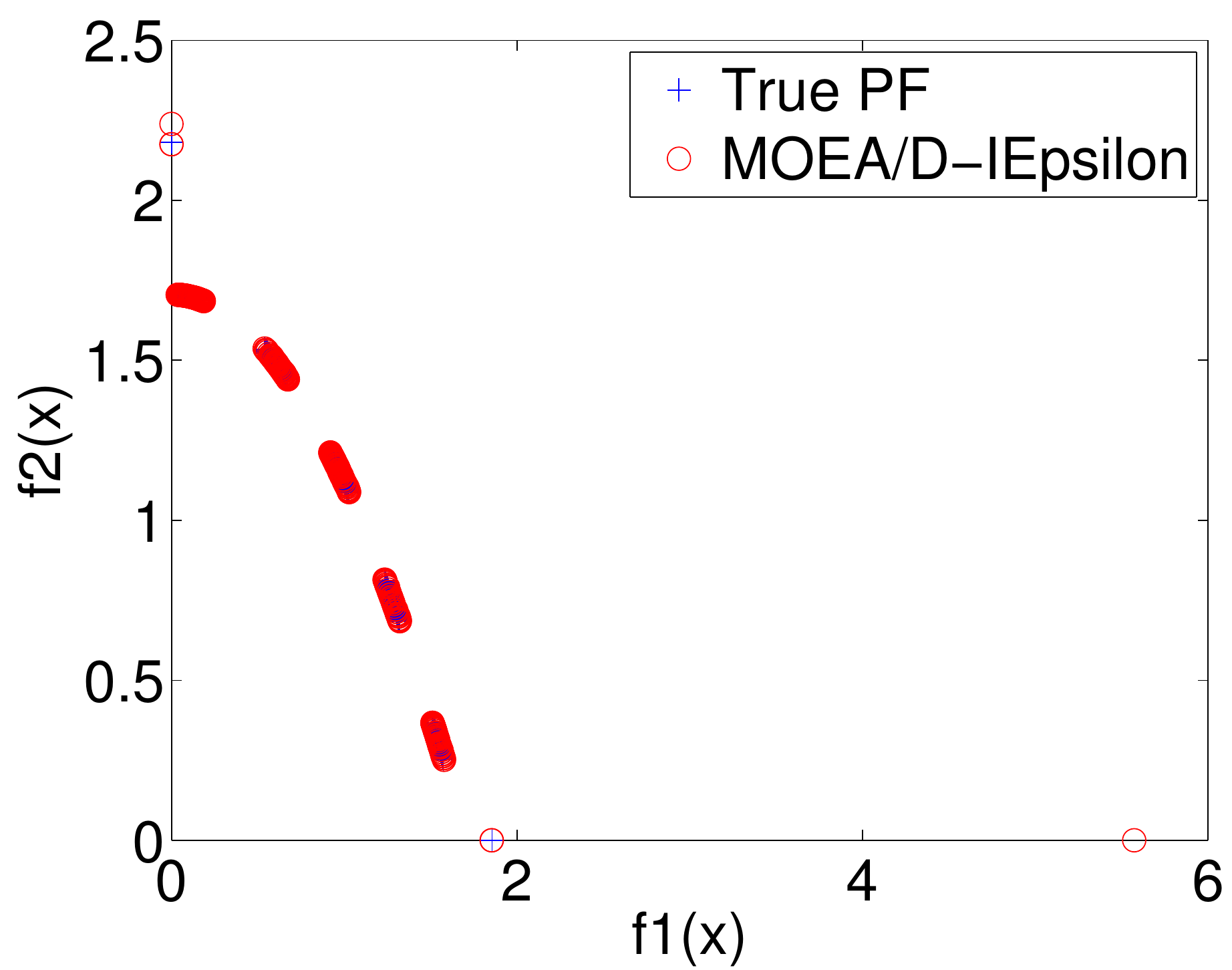}\\
\end{minipage}
\begin{minipage}[t]{0.25\linewidth}
\includegraphics[width = 4.5cm]{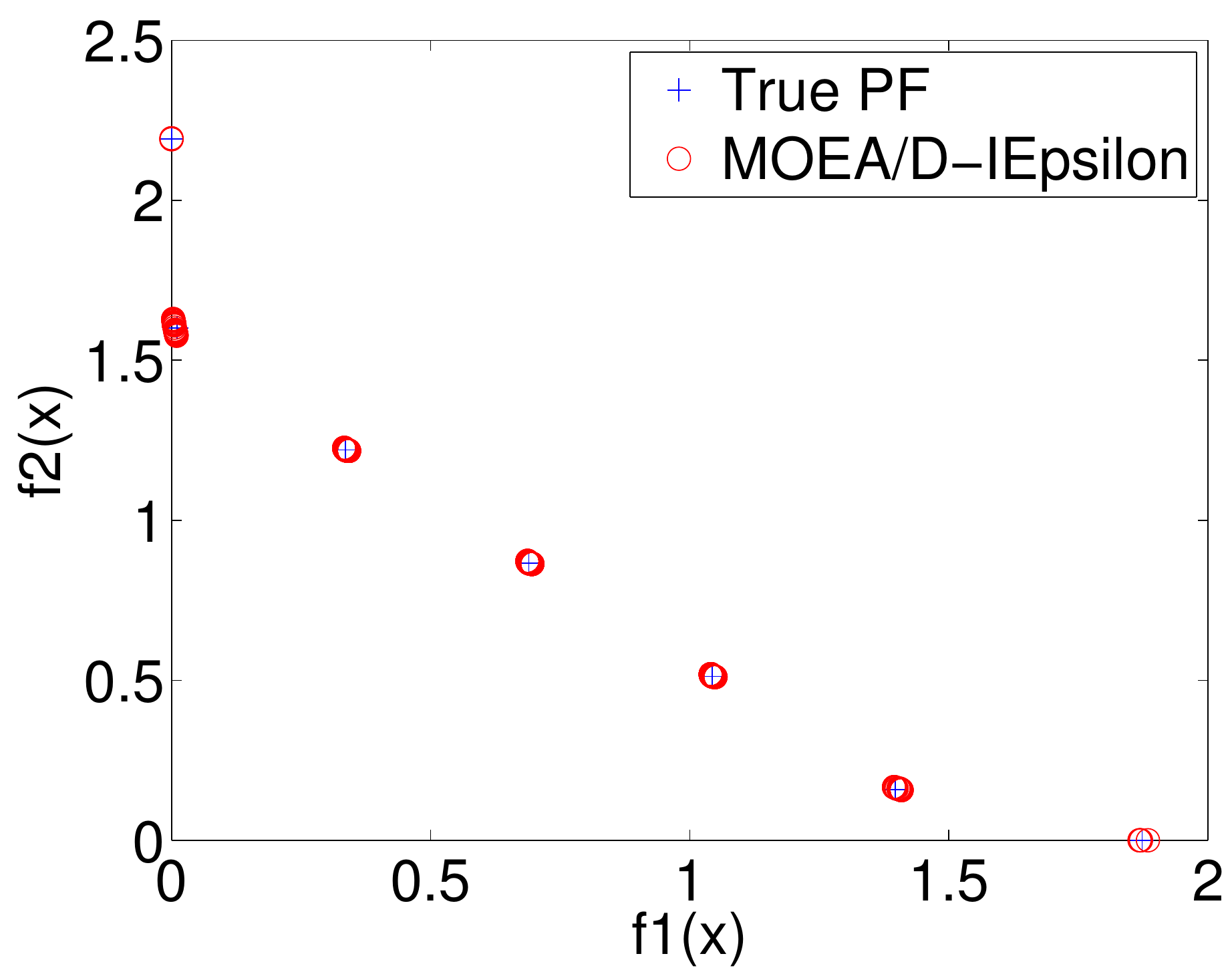}\\
\end{minipage}
\end{tabular}

\begin{tabular}{cc}
\begin{minipage}[t]{0.25\linewidth}
\includegraphics[width = 4.5cm]{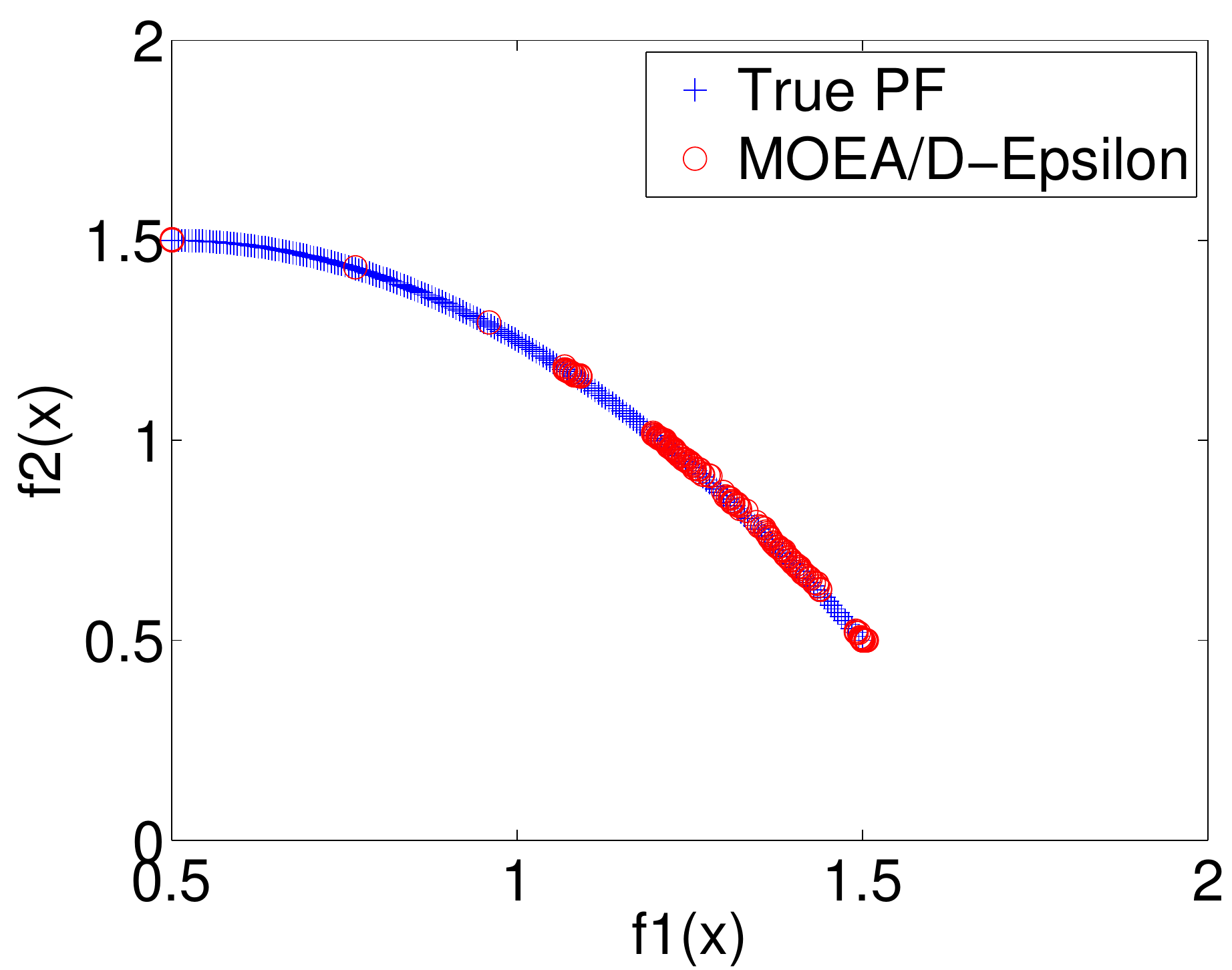}\\
\end{minipage}
\begin{minipage}[t]{0.25\linewidth}
\includegraphics[width = 4.5cm]{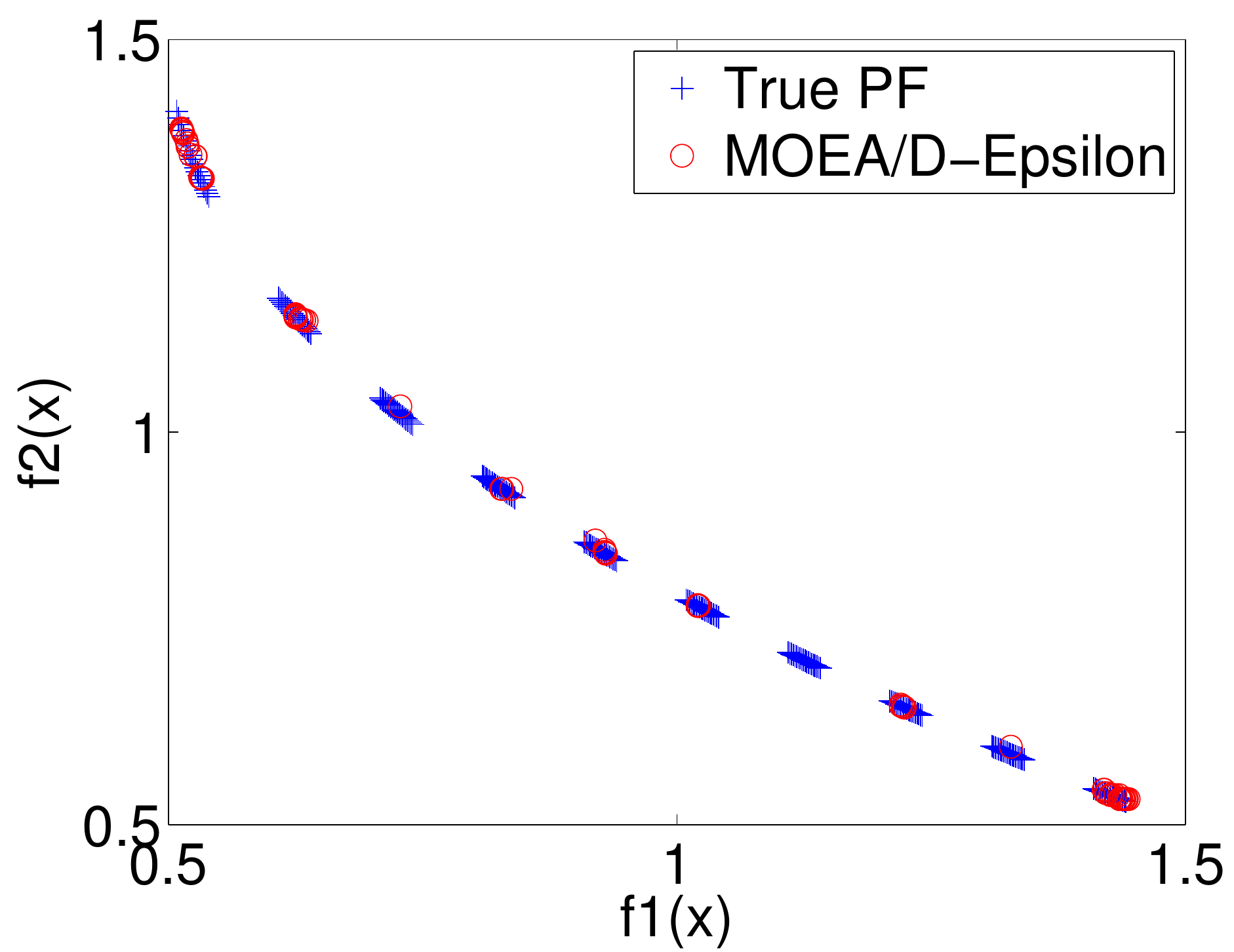}\\
\end{minipage}
\begin{minipage}[t]{0.25\linewidth}
\includegraphics[width = 4.5cm]{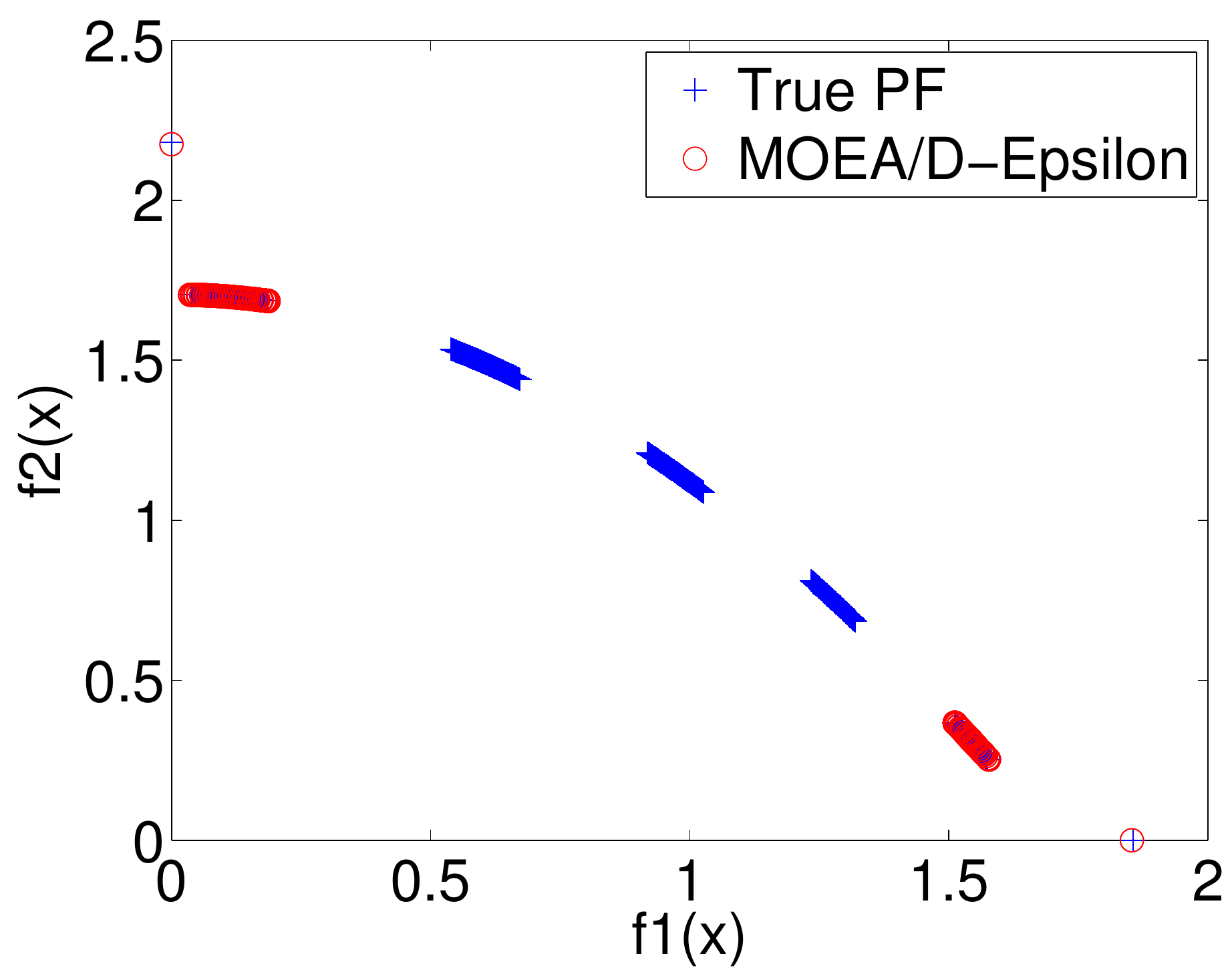}\\
\end{minipage}
\begin{minipage}[t]{0.25\linewidth}
\includegraphics[width = 4.5cm]{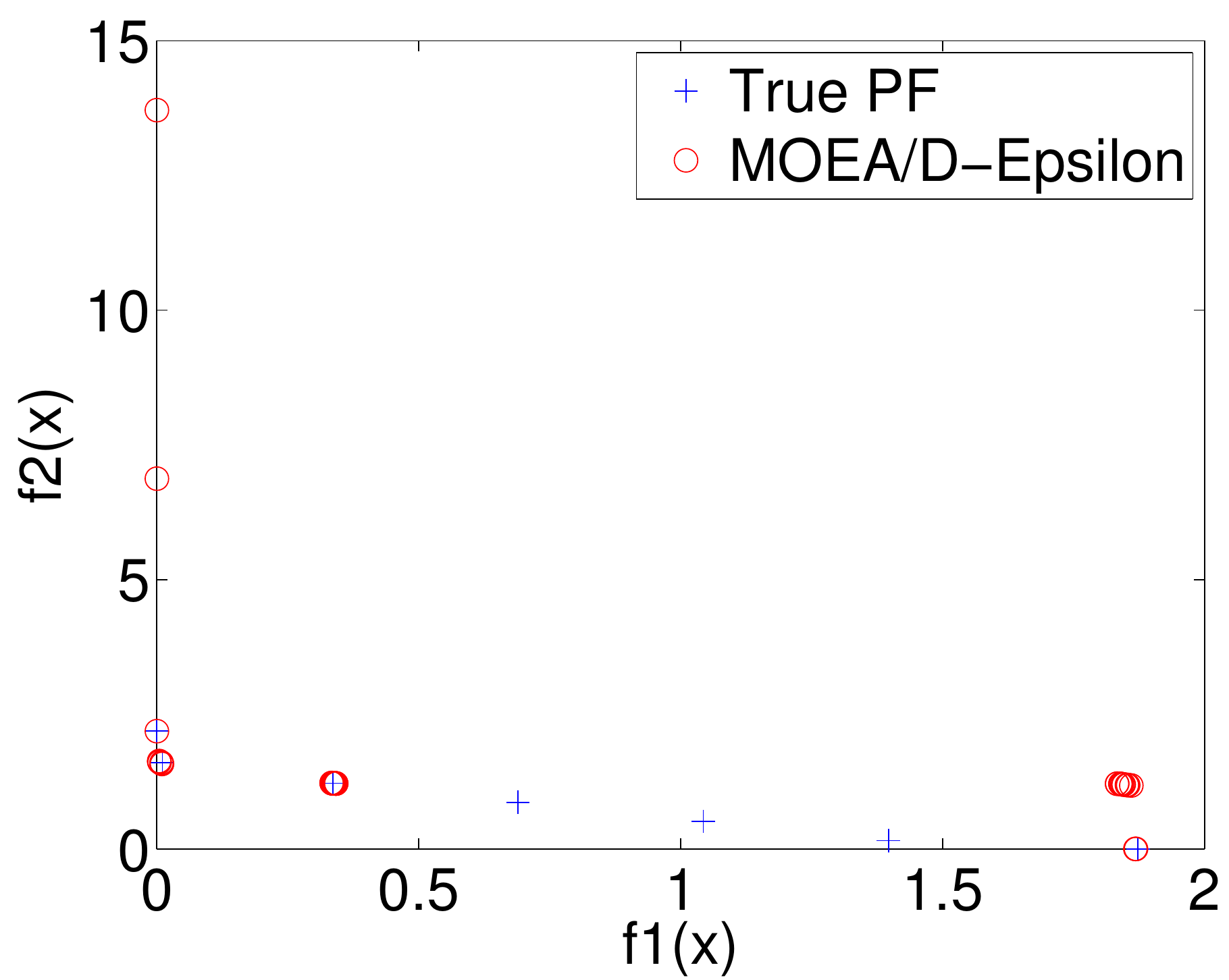}\\
\end{minipage}
\end{tabular}

\begin{tabular}{cc}
\begin{minipage}[t]{0.25\linewidth}
\includegraphics[width = 4.5cm]{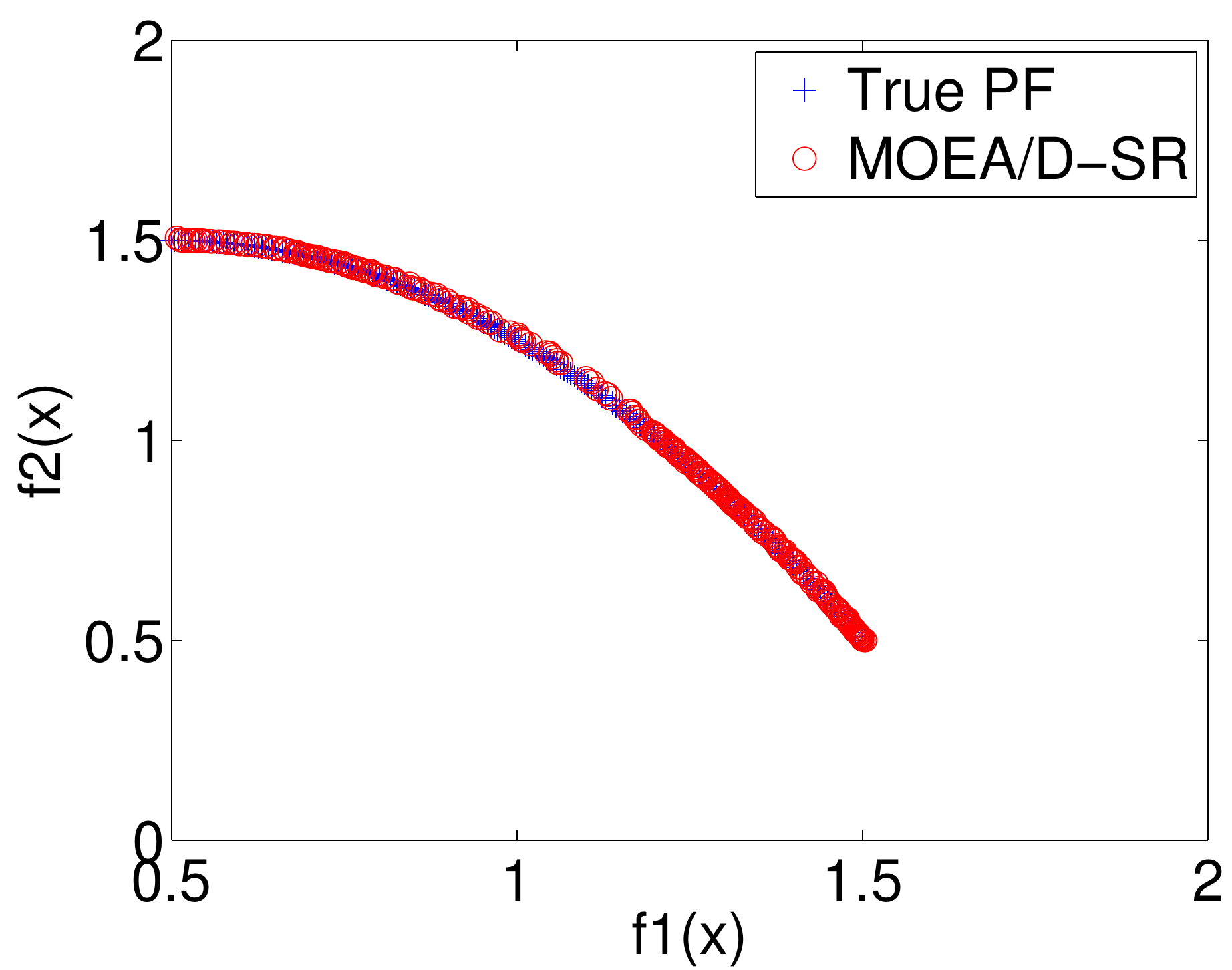}\\
\end{minipage}
\begin{minipage}[t]{0.25\linewidth}
\includegraphics[width = 4.5cm]{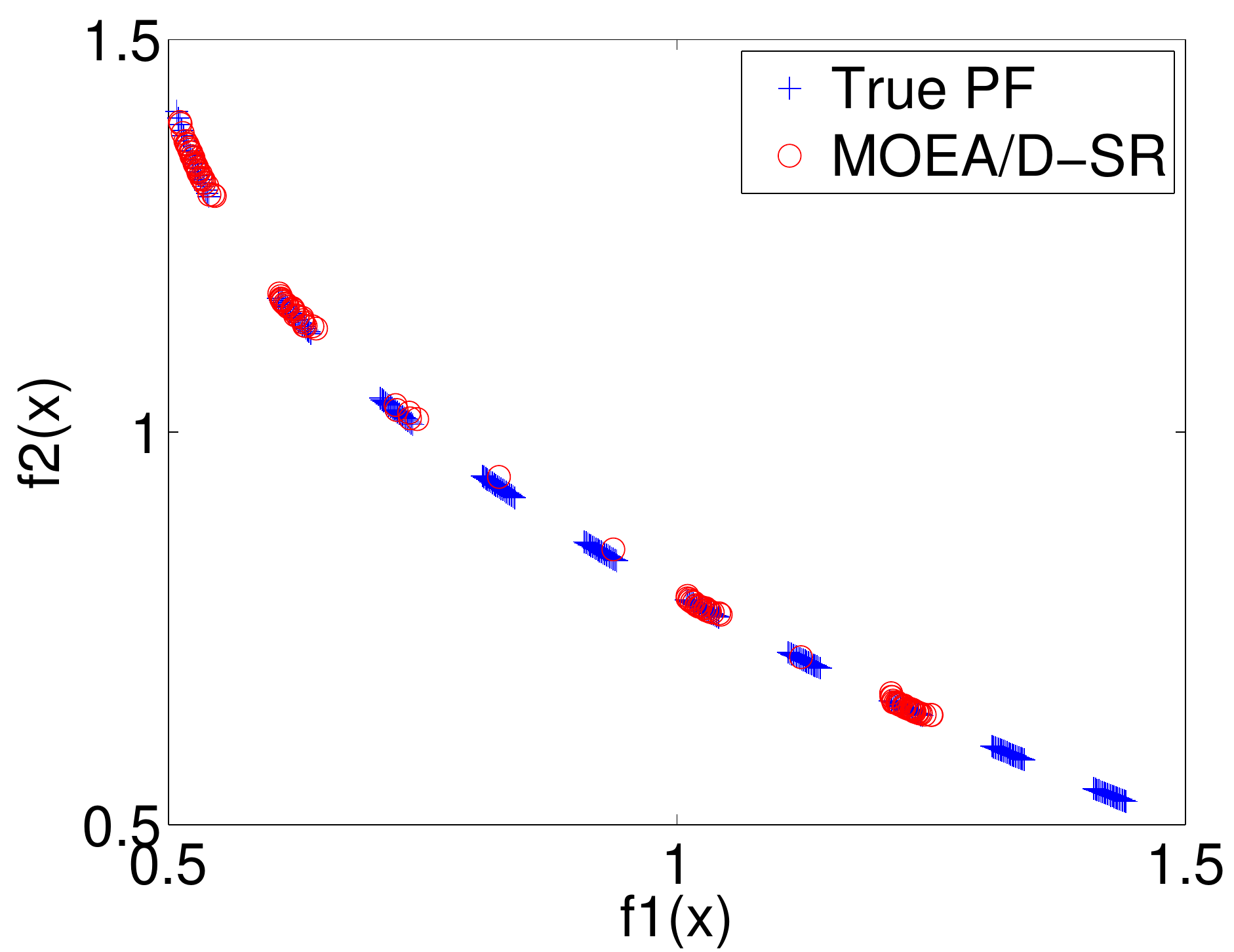}\\
\end{minipage}
\begin{minipage}[t]{0.25\linewidth}
\includegraphics[width = 4.5cm]{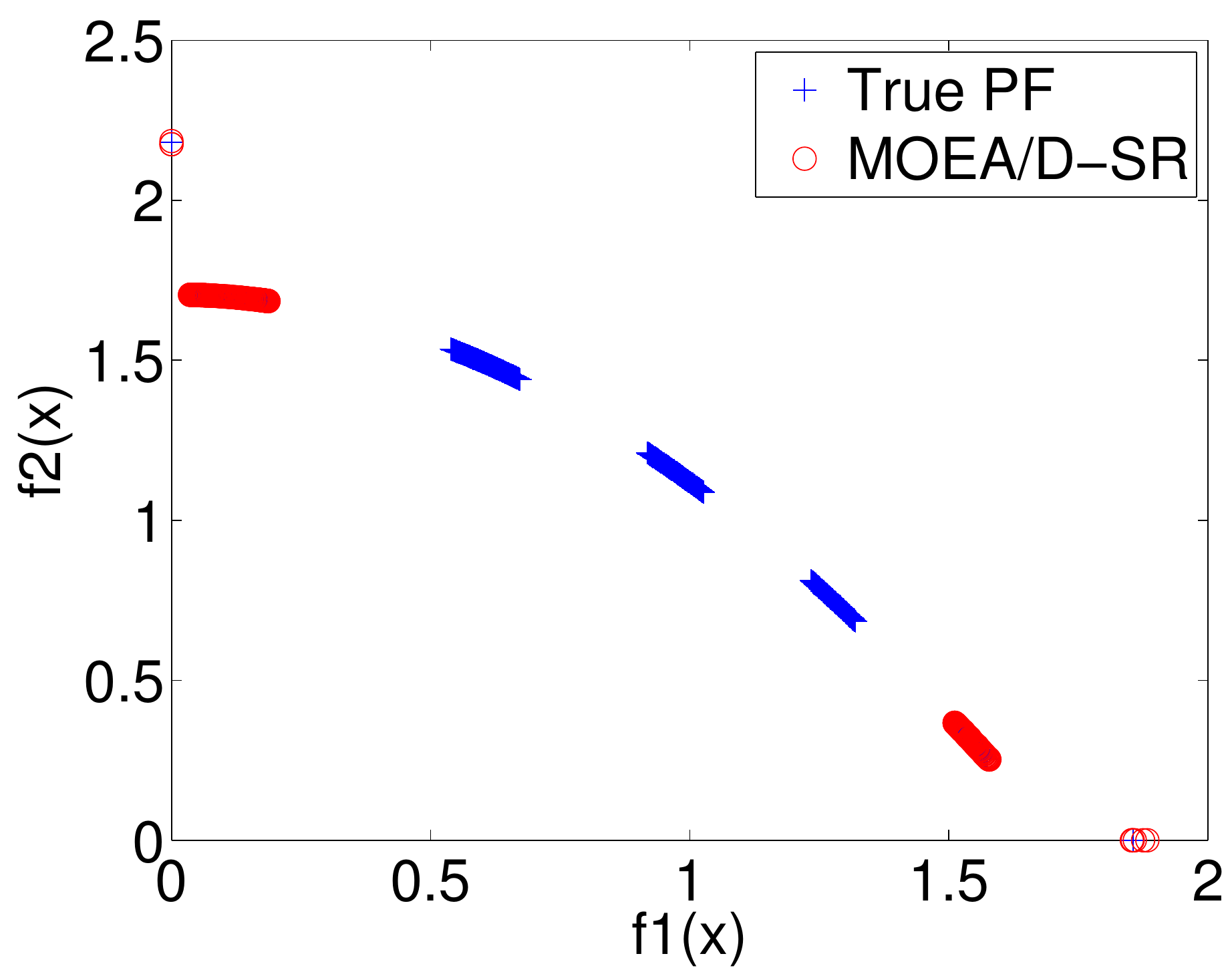}\\
\end{minipage}
\begin{minipage}[t]{0.25\linewidth}
\includegraphics[width = 4.5cm]{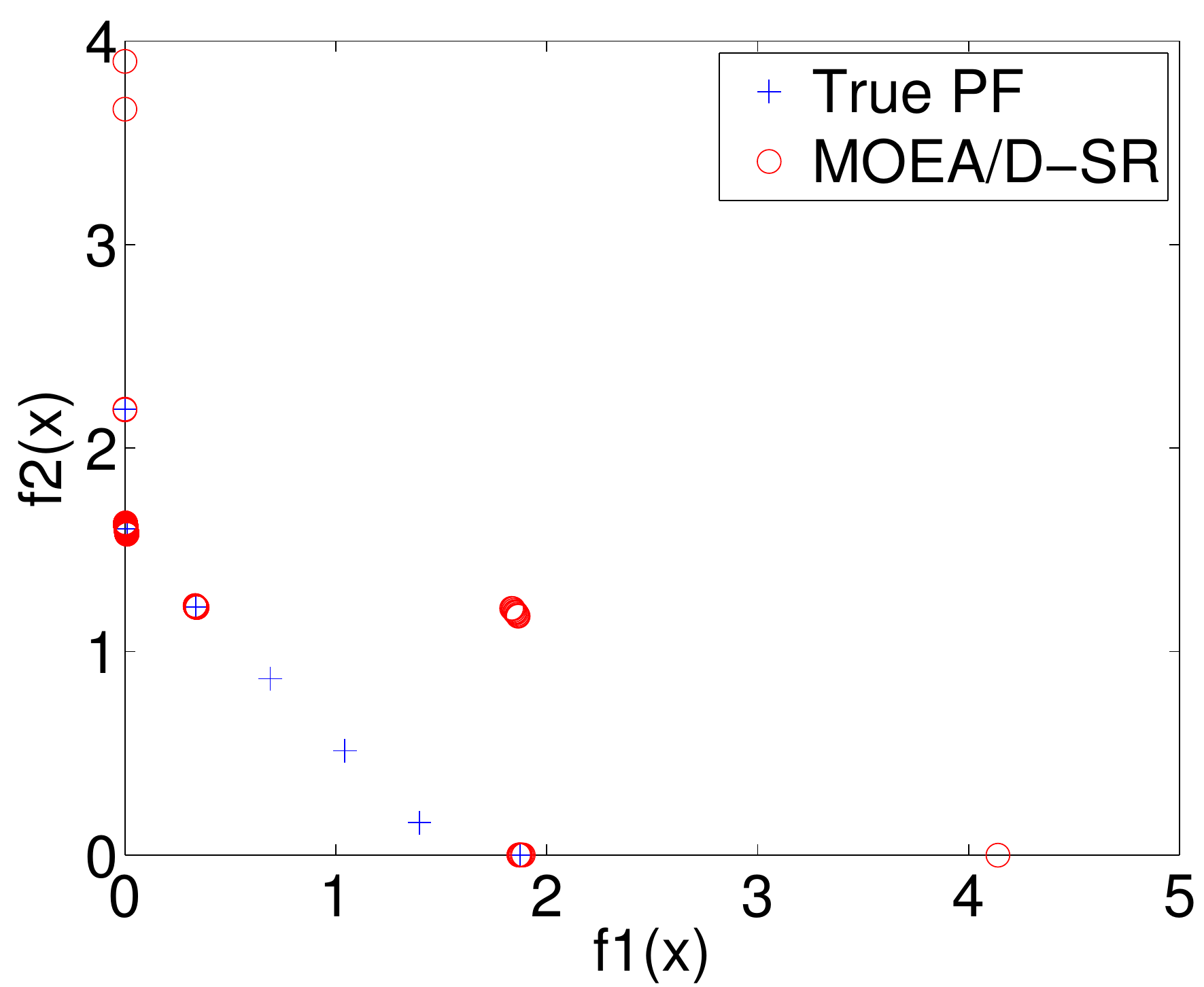}\\
\end{minipage}
\end{tabular}

\begin{tabular}{cc}
\begin{minipage}[t]{0.25\linewidth}
\includegraphics[width = 4.5cm]{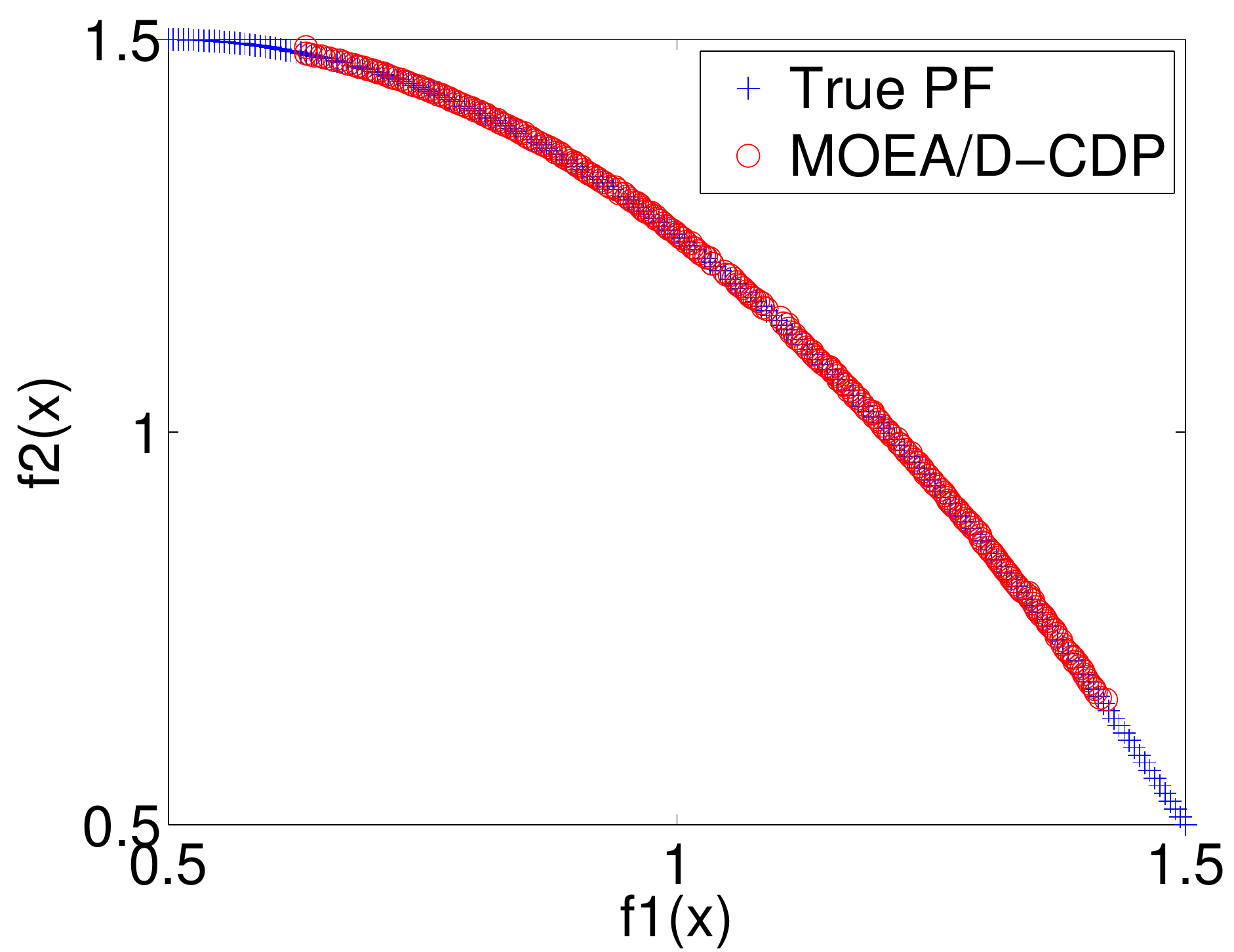}\\
\end{minipage}
\begin{minipage}[t]{0.25\linewidth}
\includegraphics[width = 4.5cm]{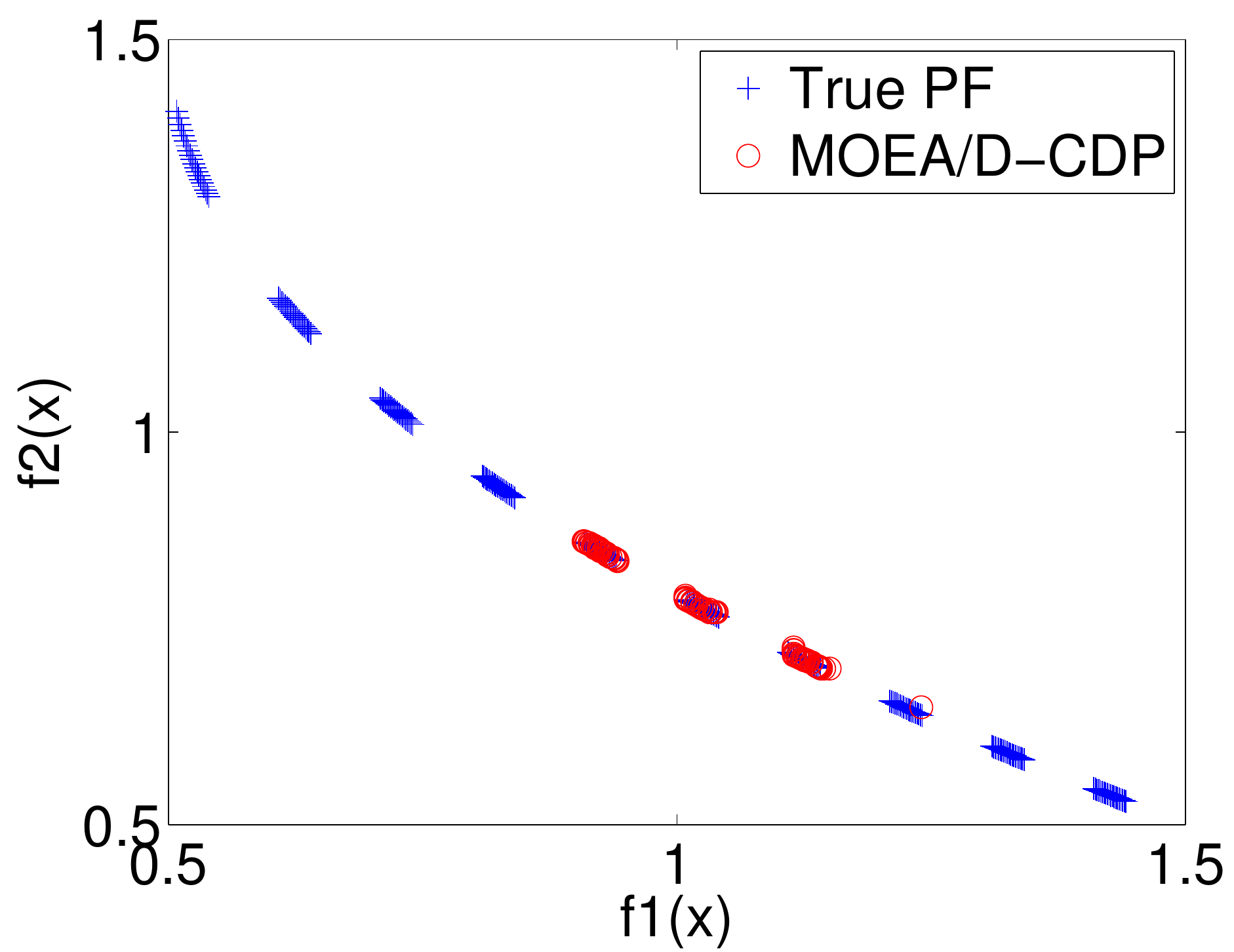}\\
\end{minipage}
\begin{minipage}[t]{0.25\linewidth}
\includegraphics[width = 4.5cm]{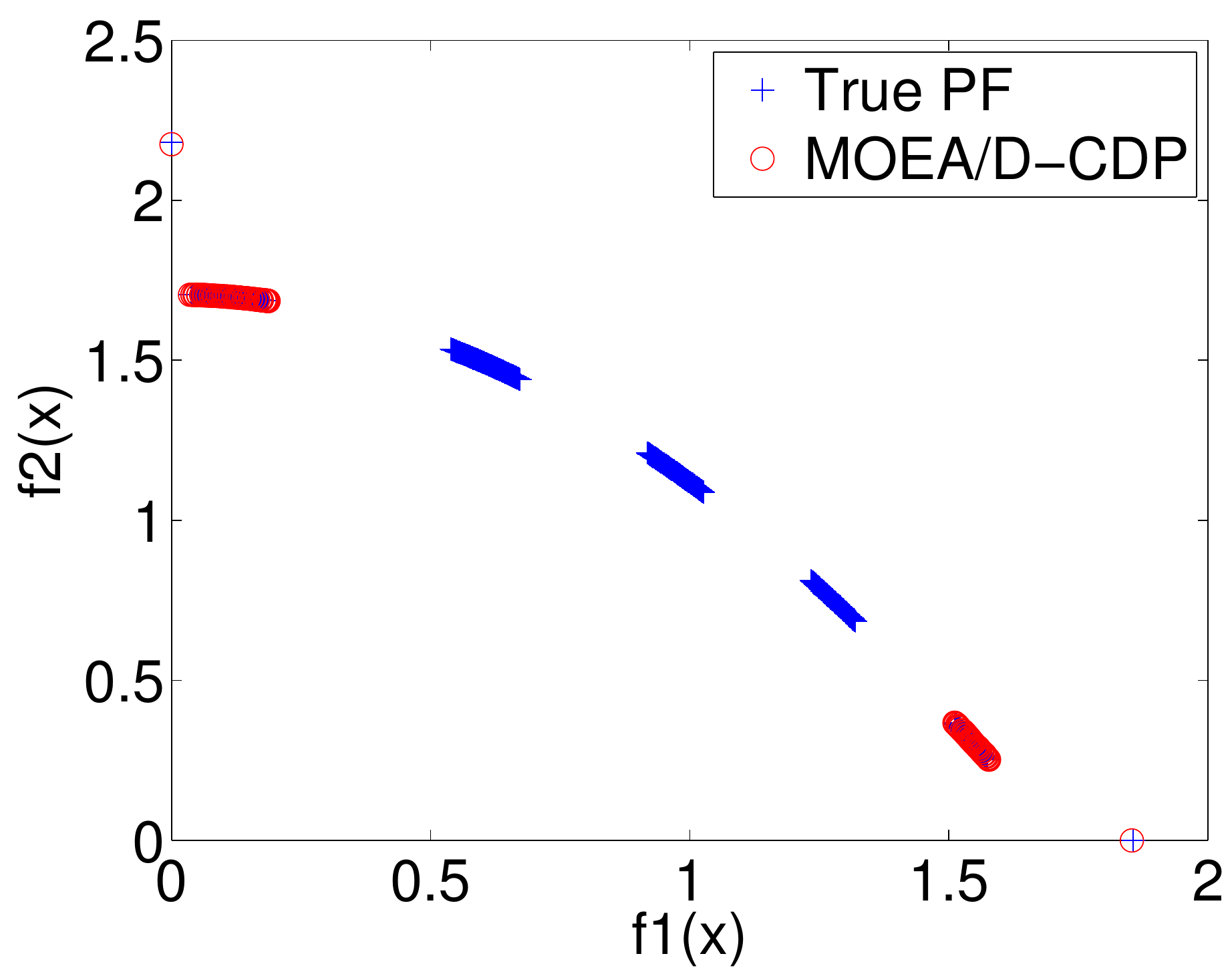}\\
\end{minipage}
\begin{minipage}[t]{0.25\linewidth}
\includegraphics[width = 4.5cm]{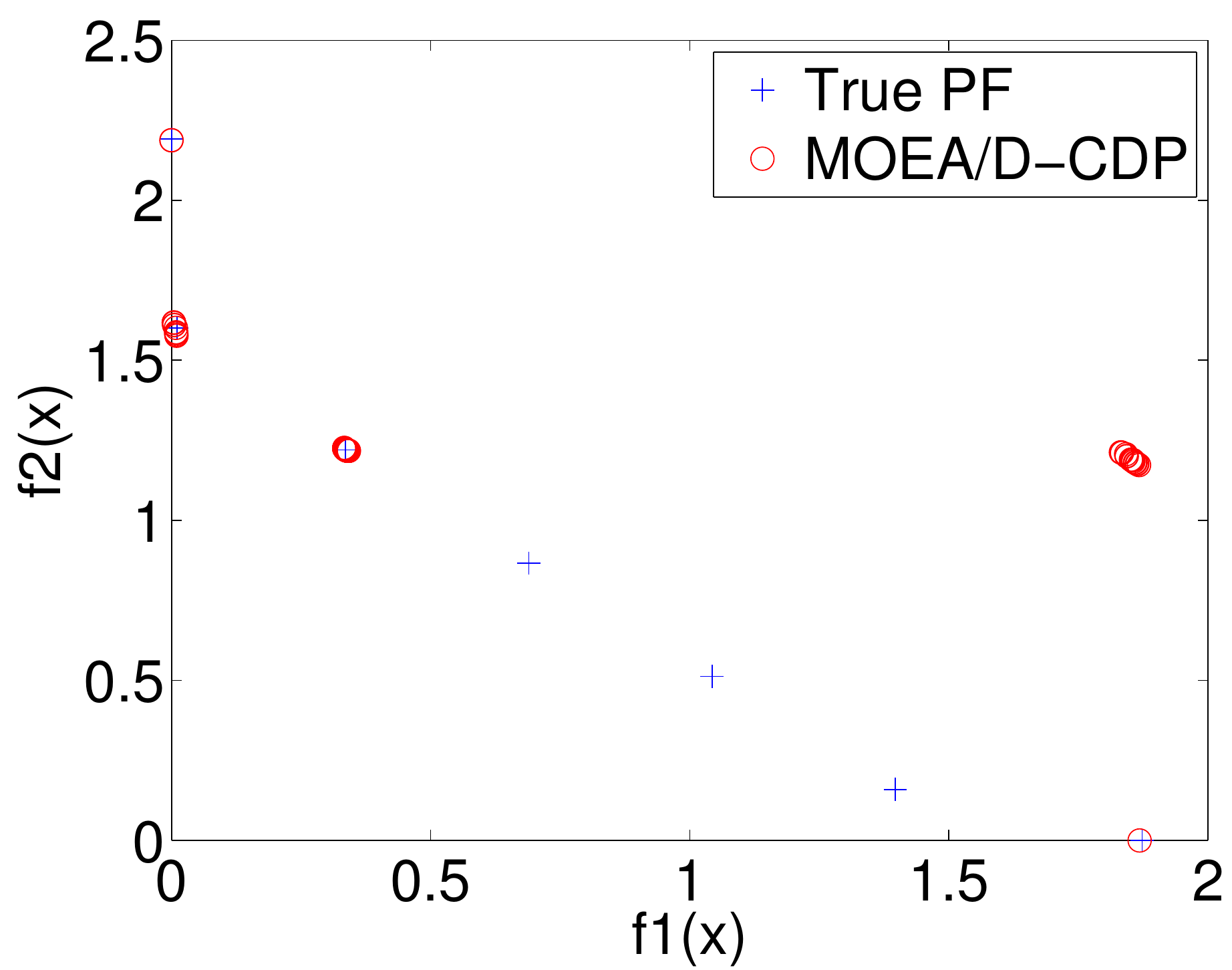}\\
\end{minipage}
\end{tabular}

\begin{tabular}{cc}
\begin{minipage}[t]{0.25\linewidth}
\includegraphics[width = 4.5cm]{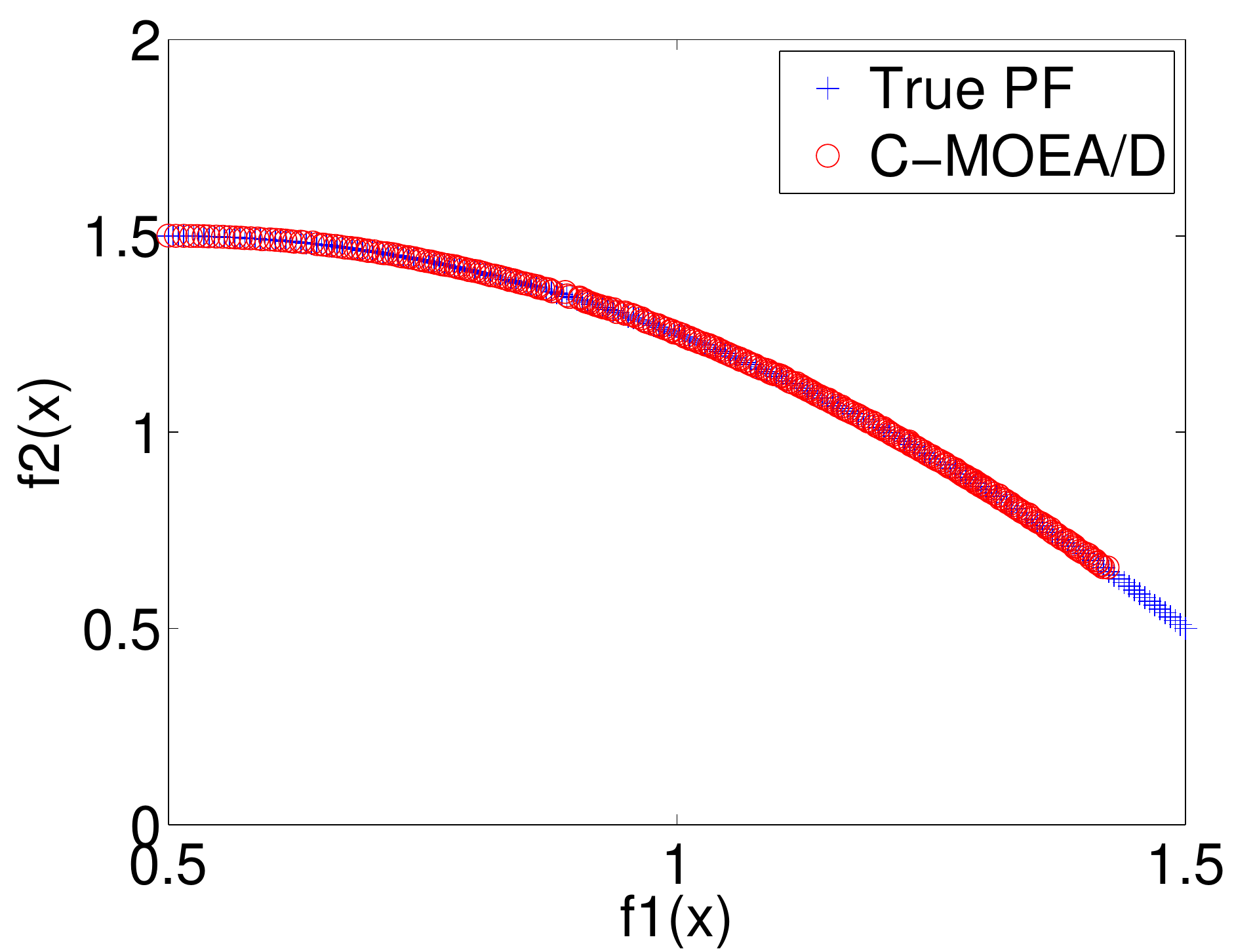}\\
\centering{\scriptsize{(a) LIR-CMOP1}}
\end{minipage}
\begin{minipage}[t]{0.25\linewidth}
\includegraphics[width = 4.5cm]{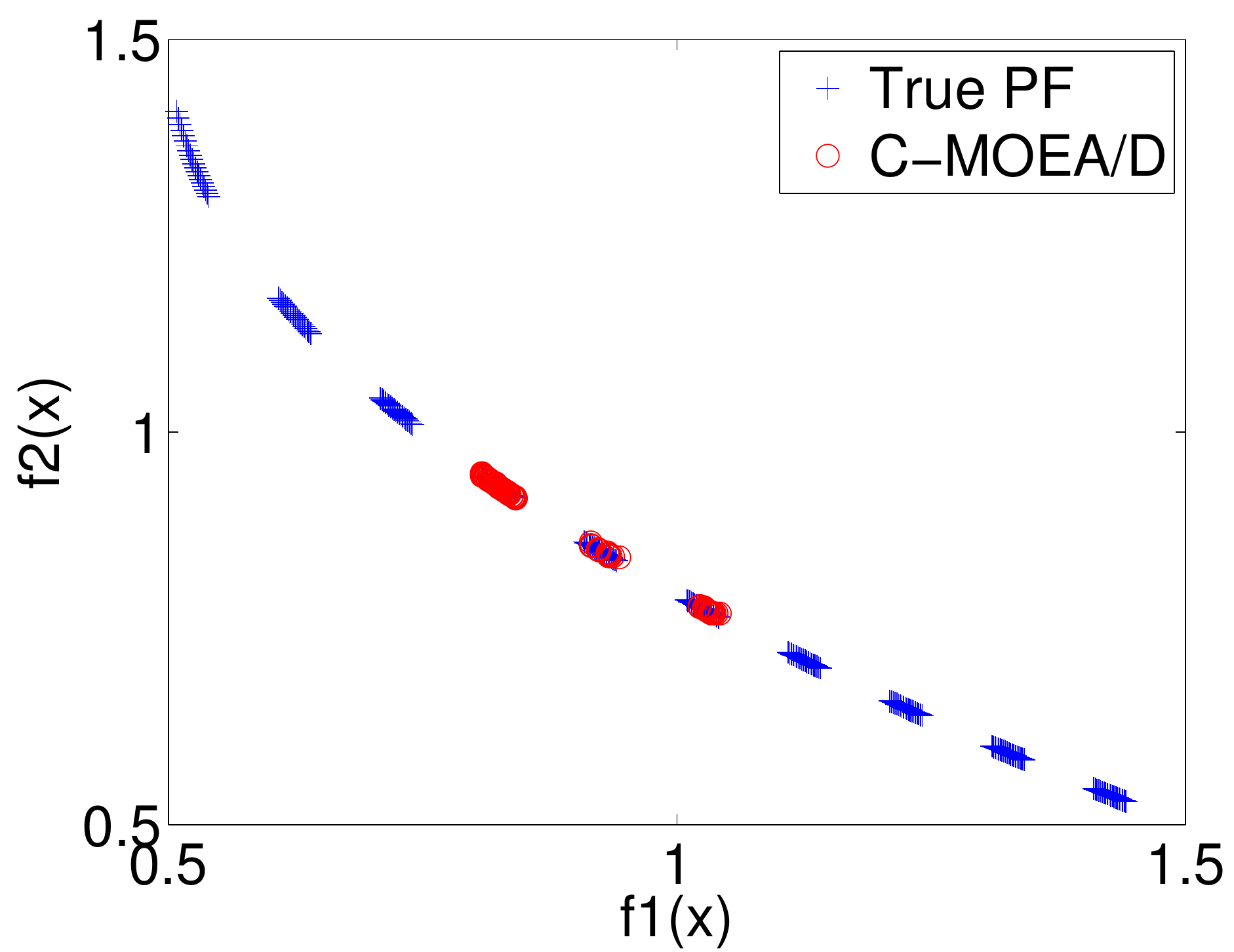}\\
\centering{\scriptsize{(b) LIR-CMOP4}}
\end{minipage}
\begin{minipage}[t]{0.25\linewidth}
\includegraphics[width = 4.5cm]{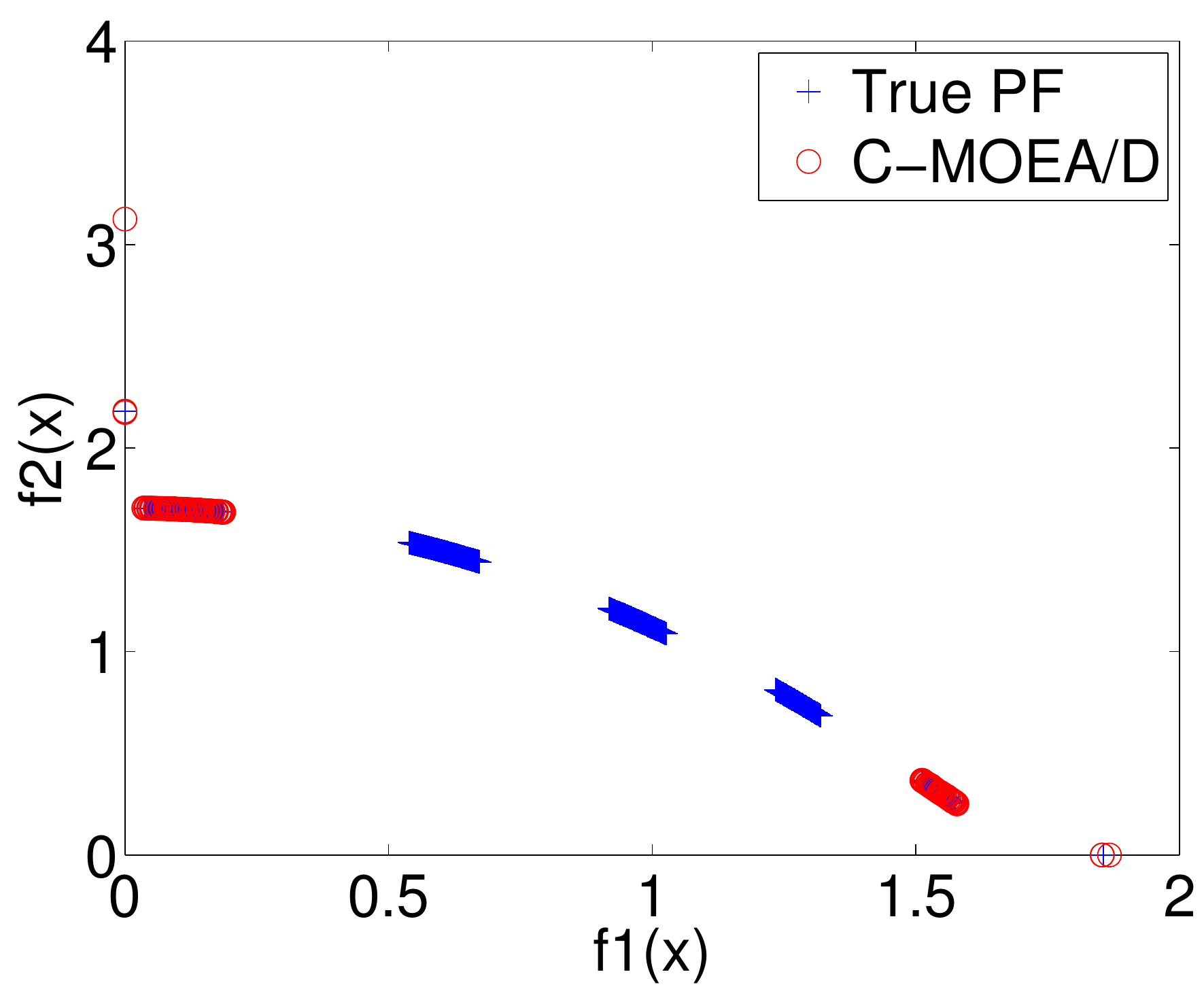}\\
\centering{\scriptsize{(c) LIR-CMOP9}}
\end{minipage}
\begin{minipage}[t]{0.25\linewidth}
\includegraphics[width = 4.5cm]{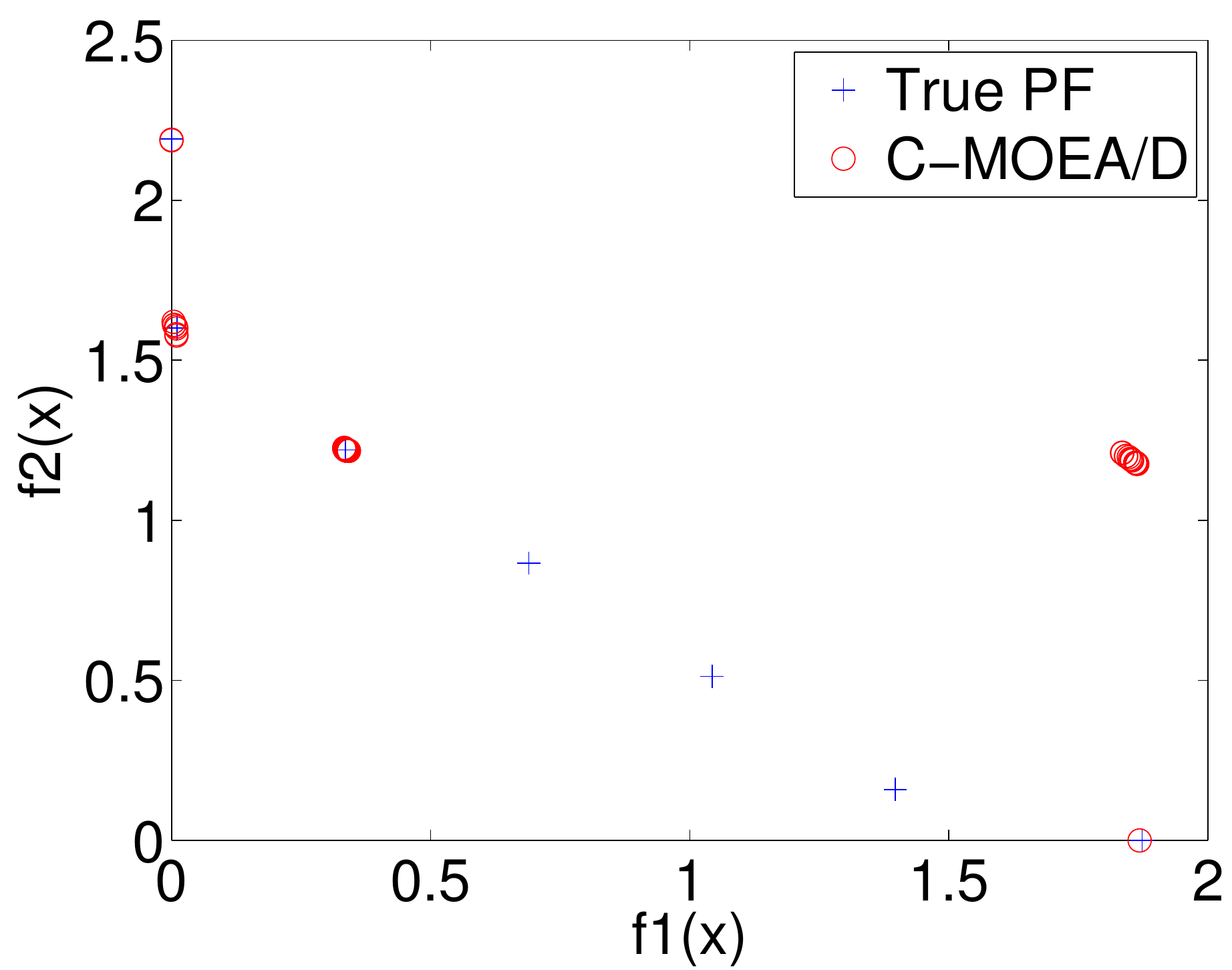}\\
\centering{\scriptsize{(d) LIR-CMOP11}}
\end{minipage}
\end{tabular}

\caption{The non-dominated solutions achieved by each algorithm with the minimized IGD in the 30 independent runs for LIR-CMOP1, LIR-CMOP4, LIR-CMOP9 and LIR-CMOP11.} \label{fig:lir-cmop-selected}
\end{figure*}

The statistical results of the IGD values on LIR-CMOP1-14 achieved by five CMOEAs in 30 independent runs are listed in Table \ref{tab:lir-cmop-igd}. As discussed in Section \ref{sec:4}, LIR-CMOP1-4 have large infeasible regions in the entire search space. For these four test instances, MOEA/D-IEpsilon is significantly better than the other four tested CMOEAs in term of the IGD metric. Fig. \ref{fig:lir-cmop-selected}(a)-(b) shows the final populations achieved by each CMOEA with the best IGD values during the 30 runs on LIR-CMOP1 and LIR-CMOP4. It is clear that MOEA/D-IEpsilon has the best performance regarding diversity among the five CMOEAs under test.

LIR-CMOP5-12 have large infeasible regions, as discussed in Section \ref{sec:4}. It can be observed that MOEA/D-IEpsilon is significantly better than the other four tested CMOEAs on NCMOP5-12. The final populations achieved by each CMOEA on LIR-CMOP9 and LIR-CMOP11 with the best IGD values are plotted in Fig. \ref{fig:lir-cmop-selected}(c)-(d). For LIR-CMOP9, MOEA/D-Epsilon, MOEA/D-SR, MOEA/D-CDP and C-MOEA/D only achieve a part of the real PF. However, MOEA/D-IEpsilon can obtain the whole real PF. Thus, MOEA/D-IEpsilon performs better than the other four CMOEAs in terms of diversity. For LIR-CMOP11, the proposed method MOEA/D-IEpsilon can achieve the whole PF. However, the other four CMOEAs do not converge to the whole PF. Thus, MOEA/D-IEpsilon has better convergence performance than the other four CMOEAs. For three-objective test instances (LIR-CMOP13 and LIR-CMOP14), MOEA/D-IEpsilon is also significantly better than the other four CMOEAs. 

Table \ref{tab:lir-cmop-hv} shows the results of the HV values of LIR-CMOP1-14 achieved by five CMOEAs in 30 independent runs. It is clear that MOEA/D-IEpilon is significantly better than the other four CMOEAs on all of the fourteen test instances in terms of the $HV$ metric. 

\subsubsection{Analysis of Experimental Results} 
From the above performance comparison on the fourteen test instances LIR-CMOP1-14, it is clear that MOEA/D-IEpsilon has better diversity and convergence performance than the other four decomposition-based CMOEAs on these fourteen test instances. A common feature of these test instances is that each of them has a number of large infeasible regions, which demonstrates that the proposed epsilon constraint-handling method can deal with the large infeasible regions very well using its automatic adjustment of the epsilon level.

\begin{table*}[htbp]
  \centering
  \caption{IGD results of MOEA/D-IEpsilon and the other four CMOEAs on LIR-CMOP1-14 test instances}
    \begin{tabular}{c|c|ccccc}
    \toprule
    \multicolumn{2}{c|}{Test Instances} & MOEA/D-IEpsilon & MOEA/D-Epsilon & MOEA/D-SR & MOEA/D-CDP & C-MOEA/D \\
    \hline
    \multirow{2}[0]{*}{LIR-CMOP1} & mean  & \textbf{7.213E-03} & 7.432E-02$^{\dag}$ & 1.719E-02$^{\dag}$ & 1.163E-01$^{\dag}$ & 1.290E-01$^{\dag}$ \\
          & std   & 2.425E-03 & 3.538E-02 & 1.554E-02 & 7.265E-02 & 8.055E-02 \\
    \hline
    \multirow{2}[0]{*}{LIR-CMOP2} & mean  & \textbf{5.461E-03} & 6.407E-02$^{\dag}$ & 9.274E-03$^{\dag}$ & 1.244E-01$^{\dag}$ & 1.627E-01$^{\dag}$ \\
          & std   & 1.520E-03 & 3.869E-02 & 9.723E-03 & 5.492E-02 & 5.819E-02 \\
    \hline
    \multirow{2}[0]{*}{LIR-CMOP3} & mean  & \textbf{1.117E-02} & 9.570E-02$^{\dag}$ & 1.792E-01$^{\dag}$ & 2.460E-01$^{\dag}$ & 2.751E-01$^{\dag}$ \\
          & std   & 6.856E-03 & 4.529E-02 & 7.306E-02 & 4.444E-02 & 3.895E-02 \\
    \hline
    \multirow{2}[0]{*}{LIR-CMOP4} & mean  & \textbf{4.859E-03} & 6.141E-02$^{\dag}$ & 2.034E-01$^{\dag}$ & 2.486E-01$^{\dag}$ & 2.631E-01$^{\dag}$ \\
          & std   & 1.591E-03 & 4.127E-02 & 6.038E-02 & 3.858E-02 & 3.331E-02 \\
    \hline
    \multirow{2}[0]{*}{LIR-CMOP5} & mean  & \textbf{2.107E-03} & 9.455E-01$^{\dag}$ & 1.041E+00$^{\dag}$ & 9.827E-01$^{\dag}$ & 8.637E-01$^{\dag}$ \\
          & std   & 2.616E-04 & 4.705E-01 & 3.833E-01 & 4.140E-01 & 5.071E-01 \\
    \hline
    \multirow{2}[0]{*}{LIR-CMOP6} & mean  & \textbf{2.058E-01} & 1.177E+00$^{\dag}$ & 8.699E-01$^{\dag}$ & 1.224E+00$^{\dag}$ & 1.277E+00$^{\dag}$ \\
          & std   & 4.587E-01 & 4.376E-01 & 5.992E-01 & 3.726E-01 & 2.587E-01 \\
    \hline
    \multirow{2}[0]{*}{LIR-CMOP7} & mean  & \textbf{4.598E-02} & 1.475E+00$^{\dag}$ & 1.074E+00$^{\dag}$ & 1.402E+00$^{\dag}$ & 1.511E+00$^{\dag}$ \\
          & std   & 6.855E-02 & 5.309E-01 & 7.606E-01 & 6.226E-01 & 5.032E-01 \\
    \hline
    \multirow{2}[0]{*}{LIR-CMOP8} & mean  & \textbf{3.445E-02} & 1.522E+00$^{\dag}$ & 1.253E+00$^{\dag}$ & 1.361E+00$^{\dag}$ & 1.575E+00$^{\dag}$ \\
          & std   & 6.002E-02 & 4.716E-01 & 6.597E-01 & 5.888E-01 & 3.849E-01 \\
    \hline
    \multirow{2}[0]{*}{LIR-CMOP9} & mean  & \textbf{1.290E-02} & 4.902E-01$^{\dag}$ & 4.883E-01$^{\dag}$ & 4.994E-01$^{\dag}$ & 4.902E-01$^{\dag}$ \\
          & std   & 3.300E-02 & 4.221E-02 & 4.130E-02 & 2.526E-02 & 4.221E-02 \\
    \hline
    \multirow{2}[0]{*}{LIR-CMOP10} & mean  & \textbf{2.143E-03} & 2.202E-01$^{\dag}$ & 1.898E-01$^{\dag}$ & 2.042E-01$^{\dag}$ & 2.114E-01$^{\dag}$ \\
          & std   & 1.261E-04 & 3.589E-02 & 6.277E-02 & 6.573E-02 & 5.641E-02 \\
    \hline
    \multirow{2}[0]{*}{LIR-CMOP11} & mean  & \textbf{4.713E-02} & 3.809E-01$^{\dag}$ & 2.911E-01$^{\dag}$ & 3.221E-01$^{\dag}$ & 3.321E-01$^{\dag}$ \\
          & std   & 5.410E-02 & 1.131E-01 & 3.525E-02 & 7.723E-02 & 7.109E-02 \\
    \hline
    \multirow{2}[0]{*}{LIR-CMOP12} & mean  & \textbf{4.711E-02} & 2.574E-01$^{\dag}$ & 2.045E-01$^{\dag}$ & 2.289E-01$^{\dag}$ & 2.472E-01$^{\dag}$ \\
          & std   & 5.662E-02 & 8.768E-02 & 6.771E-02 & 7.823E-02 & 8.883E-02 \\
    \hline
    \multirow{2}[0]{*}{LIR-CMOP13} & mean  & \textbf{6.447E-02} & 1.239E+00$^{\dag}$ & 1.059E+00$^{\dag}$ & 1.190E+00$^{\dag}$ & 1.215E+00$^{\dag}$ \\
          & std   & 1.844E-03 & 2.555E-01 & 5.033E-01 & 3.290E-01 & 3.140E-01 \\
    \hline
    \multirow{2}[0]{*}{LIR-CMOP14} & mean  & \textbf{6.502E-02} & 1.172E+00$^{\dag}$ & 9.005E-01$^{\dag}$ & 1.204E+00$^{\dag}$ & 1.054E+00$^{\dag}$ \\
          & std   & 1.635E-03 & 3.043E-01 & 5.455E-01 & 2.244E-01 & 4.515E-01 \\
    \bottomrule
    \end{tabular}%
  \label{tab:lir-cmop-igd} \\
  Wilcoxon’s rank sum test at a 0.05 significance level is performed between MOEA/D-IEpsilon and each of the other four CMOEAs. $\dag$ and $\ddag$ denote that the performance of the corresponding algorithm is significantly worse than or better than that of MOEA/D-IEpsilon, respectively. The best mean is highlighted in boldface.
\end{table*}%

\begin{table*}[htbp]
  \centering
  \caption{HV results of MOEA/D-IEpsilon and the other four CMOEAs on LIR-CMOP1-14 test instances}
    \begin{tabular}{c|c|ccccc}
    \toprule
    \multicolumn{2}{c|}{Test Instances} & MOEA/D-IEpsilon & MOEA/D-Epsilon & MOEA/D-SR & MOEA/D-CDP & C-MOEA/D \\
    \hline
    \multirow{2}[0]{*}{LIR-CMOP1} & mean     & \textbf{1.015E+00} & 9.413E-01$^{\dag}$ & 9.840E-01$^{\dag}$ & 7.499E-01$^{\dag}$ & 7.344E-01$^{\dag}$ \\
          & std     & 1.490E-03 & 3.751E-02 & 4.630E-02 & 1.202E-01 & 1.269E-01 \\
    \hline
    \multirow{2}[0]{*}{LIR-CMOP2} & mean     & \textbf{1.348E+00} & 1.267E+00$^{\dag}$ & 1.337E+00$^{\dag}$ & 1.093E+00$^{\dag}$ & 1.033E+00$^{\dag}$ \\
          & std     & 1.717E-03 & 5.526E-02 & 2.252E-02 & 1.016E-01 & 9.522E-02 \\
    \hline
    \multirow{2}[0]{*}{LIR-CMOP3} & mean     & \textbf{8.686E-01} & 7.964E-01$^{\dag}$ & 5.892E-01$^{\dag}$ & 5.034E-01$^{\dag}$ & 4.715E-01$^{\dag}$ \\
          & std    & 3.373E-03 & 3.618E-02 & 1.105E-01 & 5.141E-02 & 3.786E-02 \\
    \hline
    \multirow{2}[0]{*}{LIR-CMOP4} & mean     & \textbf{1.093E+00} & 1.024E+00$^{\dag}$ & 8.048E-01$^{\dag}$ & 7.397E-01$^{\dag}$ & 7.203E-01$^{\dag}$ \\
          & std     & 1.910E-03 & 5.903E-02 & 8.956E-02 & 5.264E-02 & 4.480E-02 \\
    \hline
    \multirow{2}[0]{*}{LIR-CMOP5} & mean     & \textbf{1.461E+00} & 2.833E-01$^{\dag}$ & 1.773E-01$^{\dag}$ & 2.428E-01$^{\dag}$ & 3.870E-01$^{\dag}$ \\
          & std    & 9.488E-04 & 5.766E-01 & 4.619E-01 & 5.031E-01 & 6.151E-01 \\
    \hline
    \multirow{2}[0]{*}{LIR-CMOP6} & mean    & \textbf{9.412E-01} & 1.255E-01$^{\dag}$ & 3.341E-01$^{\dag}$& 8.582E-02$^{\dag}$ & 3.750E-02$^{\dag}$ \\
          & std    & 3.848E-01 & 3.325E-01 & 4.458E-01 & 2.707E-01 & 1.446E-01 \\
    \hline
    \multirow{2}[0]{*}{LIR-CMOP7} & mean    & \textbf{2.847E+00} & 3.516E-01$^{\dag}$ & 9.943E-01$^{\dag}$ & 4.811E-01$^{\dag}$ & 2.933E-01$^{\dag}$\\
          & std    & 2.205E-01 & 9.304E-01 & 1.268E+00 & 1.083E+00 & 8.776E-01 \\
    \hline
    \multirow{2}[0]{*}{LIR-CMOP8} & mean    & \textbf{2.905E+00} & 2.690E-01$^{\dag}$ & 7.043E-01$^{\dag}$ & 5.223E-01$^{\dag}$ & 1.788E-01$^{\dag}$ \\
          & std    & 2.103E-01 & 8.100E-01 & 1.094E+00 & 9.669E-01 & 6.669E-01 \\
    \hline
    \multirow{2}[0]{*}{LIR-CMOP9} & mean    & \textbf{3.692E+00} & 2.737E+00$^{\dag}$ & 2.733E+00$^{\dag}$ & 2.705E+00$^{\dag}$ & 2.737E+00$^{\dag}$ \\
          & std    & 6.318E-02 & 1.484E-01 & 1.284E-01 & 8.883E-02 & 1.483E-01 \\
    \hline
    \multirow{2}[0]{*}{LIR-CMOP10} & mean    & \textbf{3.241E+00} & 2.874E+00$^{\dag}$ & 2.929E+00$^{\dag}$ & 2.899E+00$^{\dag}$ & 2.886E+00$^{\dag}$ \\
          & std    & 3.537E-04 & 7.851E-02 & 1.064E-01 & 1.207E-01 & 1.126E-01 \\
    \hline
    \multirow{2}[0]{*}{LIR-CMOP11} & mean    & \textbf{4.263E+00} & 3.218E+00$^{\dag}$ & 3.479E+00$^{\dag}$ & 3.406E+00$^{\dag}$ & 3.386E+00$^{\dag}$ \\
          & std    & 1.685E-01 & 3.542E-01 & 1.252E-01 & 2.135E-01 & 1.831E-01 \\
    \hline
    \multirow{2}[0]{*}{LIR-CMOP12} & mean    & \textbf{5.552E+00} & 4.858E+00$^{\dag}$ & 5.059E+00$^{\dag}$ & 4.972E+00$^{\dag}$ & 4.902E+00$^{\dag}$ \\
          & std    & 1.730E-01 & 3.280E-01 & 2.103E-01 & 2.596E-01 & 3.233E-01 \\
    \hline
    \multirow{2}[0]{*}{LIR-CMOP13} & mean    & \textbf{5.710E+00} & 3.097E-01$^{\dag}$ & 1.184E+00$^{\dag}$ & 5.320E-01$^{\dag}$ & 4.642E-01$^{\dag}$ \\
          & std    & 1.084E-02 & 1.048E+00 & 2.250E+00 & 1.442E+00 & 1.426E+00 \\
    \hline
    \multirow{2}[0]{*}{LIR-CMOP14} & mean    & \textbf{6.184E+00} & 5.617E-01$^{\dag}$ & 1.912E+00$^{\dag}$ & 4.032E-01$^{\dag}$ & 1.162E+00$^{\dag}$ \\
          & std    & 1.053E-02 & 1.540E+00 & 2.705E+00 & 1.127E+00 & 2.281E+00 \\
    \bottomrule
    \end{tabular}%
    \label{tab:lir-cmop-hv}\\
    Wilcoxon’s rank sum test at a 0.05 significance level is performed between MOEA/D-IEpsilon and each of the other four CMOEAs. $\dag$ and $\ddag$ denotes that the performance of the corresponding algorithm is significantly worse than or better than that of MOEA/D-IEpsilon, respectively. The best mean is highlighted in boldface.
\end{table*}%

\section{Robot Gripper Optimization}
\label{sec:6}

To verify the capability of MOEA/D-IEpsilon to solve real world optimization problems, a robot gripper optimization problem with two conflicting objectives and eight constraints is explored.

\subsection{Definition of the robot gripper optimization}
\label{sec:6.1}

\begin{figure}
\includegraphics[width = 8cm]{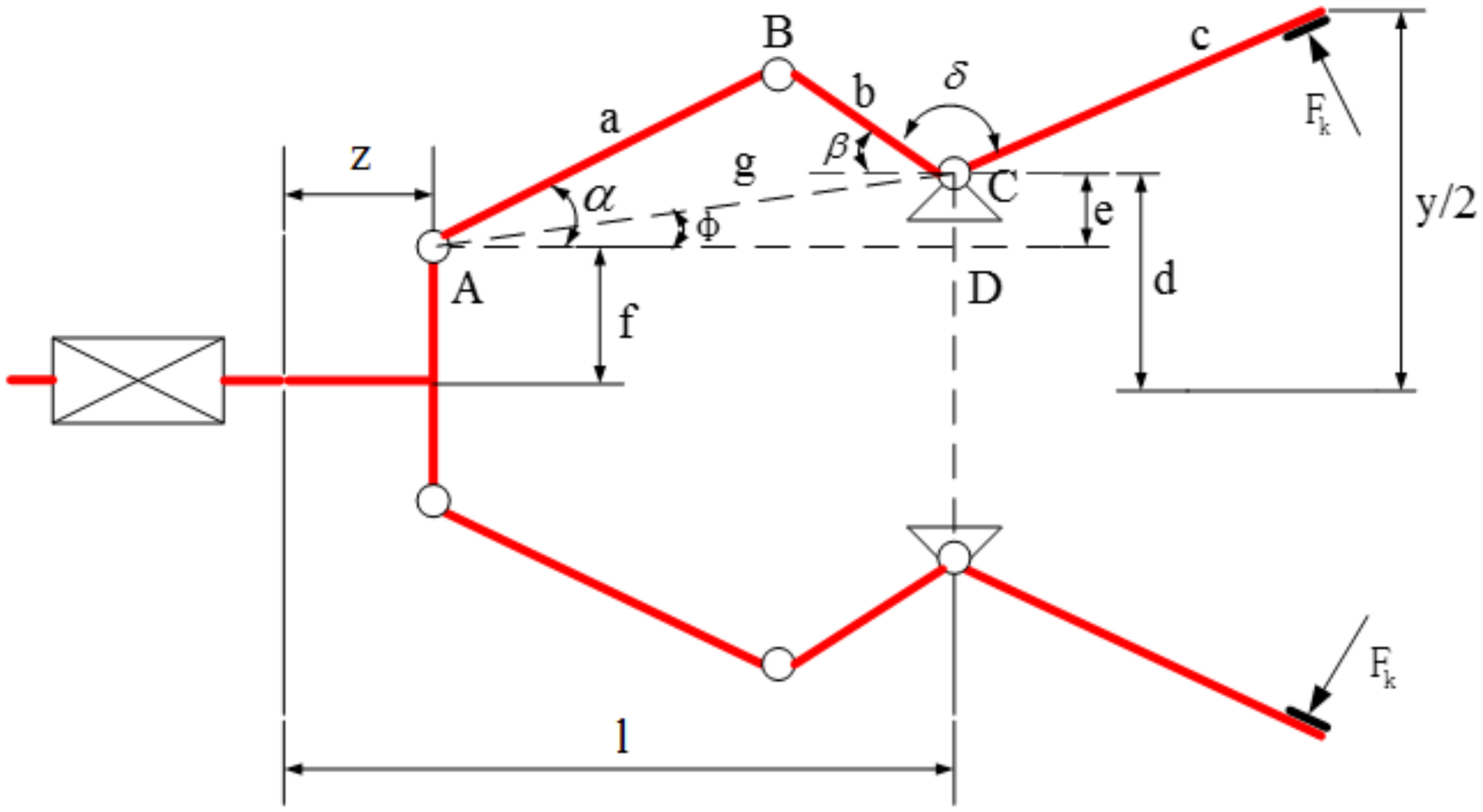}
\caption{The geometrical structure of robot gripper mechanism.}
\label{fig:gripper}
\end{figure}

The robot gripper optimization problem is defined in (\cite{saravanan2009evolutionary, Datta2011Multi}). Five objectives are formulated in these papers. The robot gripper optimization problem considered in this paper has two conflicting objectives and eight constraints. The geometrical structure of the gripper is plotted in Fig. \ref{fig:gripper}.

% \begin{figure}
% \includegraphics[width = 8cm]{figures/Gripper/geometrical_structure.png}
% \caption{The geometrical structure of robot gripper mechanism.}
% \label{fig:gripper_config}
% \end{figure}

The robot gripper optimization problem considered in this paper is defined as follows:
\begin{equation}
\label{equ:gripper_definition}
\begin{cases}
\mbox{minimize} &f_{1}(\mathbf{x}) = \frac{P}{\min_z{F_k(\mathbf{x},z)}} \\
\mbox{minimize} &f_2(x)=a+b+c+e+l \\
\mbox{subject to} & c_1(\mathbf{x}) =Y_{min}-y(\mathbf{x},Z_{max}) \geq 0 \\
& c_2(\mathbf{x}) = y(\mathbf{x},Z_{max}) \geq 0\\
& c_3(\mathbf{x}) = y(\mathbf{x},0)-Y_{max} \geq 0\\
& c_4(\mathbf{x}) = Y_G-y(\mathbf{x},0) \geq 0\\
& c_5(\mathbf{x}) = {(a+b)}^2-l^2-e^2 \geq 0\\
& c_6(\mathbf{x}) = {(l-Z_{max})}^2+{(a-e)}^2 \geq b^2\\
& c_7(\mathbf{x}) = l-Z_{max} \geq 0\\
& c_8(\mathbf{x}) = \min F_k(\mathbf{x},z)-F_G \geq 0
\end{cases}
\end{equation}
where ${\mathbf{x}}=[a, b, c, e, l, f, \delta]^{T}$ has seven decision variables, and each variable is shown in Fig. \ref{fig:gripper}. The range of each decision variable is as follows: $10mm\leq a\leq150mm$, $10mm\leq b\leq150mm$, $100mm\leq c\leq200mm$, $0mm\leq e\leq 50mm$, $10mm\leq f\leq 150mm$, $100mm\leq l\leq 300mm$ and $1.0\leq\delta\leq 3.14$. Two rules are applied to fix the value of $f$, and they are defined as follows:

\[Rule1: if\ (a<4b\ and\ c<a+b)\ then\ f=2e+10\]
\[Rule2: if\ (a<4b\ and\ c>a+b)\ then\ f=e+50\]

According to the geometric analysis, the gripping force $F_k$ in Fig. \ref{fig:gripper} can be defined as follows:
\begin{equation}
F_k=\frac{Pb\sin(\alpha+\beta)}{2c\cos\alpha}.
\end{equation}

The displacement of the gripper end is defined as follows:

\begin{equation}
y(\mathbf{x},z)=2[e+f+c+\sin(\beta+\delta)].
\end{equation}
where $g=\sqrt{{(l-z)}^2+e^2}+\phi$,\ $\alpha=\arccos(\frac{a^2+g^2-b^2}{2ag})$,\ $\beta=\arccos(\frac{b^2+g^2-a^2}{2bg})-\phi$,\
$\phi=\arctan\frac{e}{l-z}$ and $z$ denotes a dynamic displacement of the gripper actuator in the range of 0 to 100 mm.

The first objective $f_1(x)$ represents a force transmission ratio between the actuating force $P$ and the minimum gripping force $\min F_k(\mathbf{x},z)$. We prefer to transform more actuating force into the gripper force. Thus, this objective should be minimized.

The second objective $f_2(x)$ is the sum of all elements of the robot gripper. It is relevant to the weight of the robot gripper, and minimizing $f_2(x)$ can lead to a lightweight design.

\begin{figure}
\includegraphics[width = 9cm]{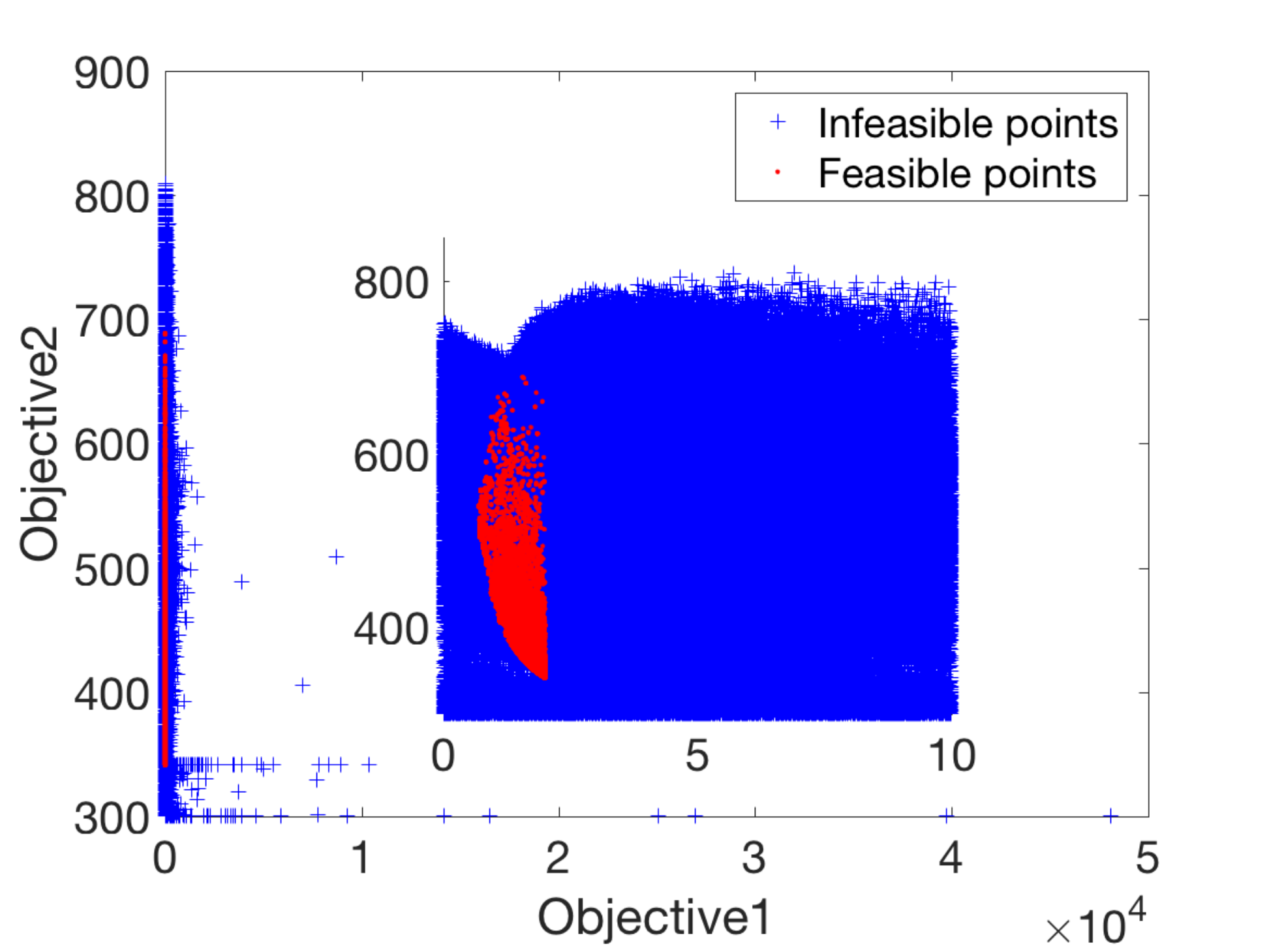}
\caption{The distribution of solutions of the robot gripper optimization problem in the objective space, $RFS=0.1396$.}
\label{fig:gripper_sampling}
\end{figure}

To study the distribution of solutions in the objective space for the robot gripper optimization problem, 3,000,000 solutions are generated, where 1,500,000 solutions are generated randomly, and the other 1,500,000 solutions are generated by MOEA/D-IEpsilon. In Fig. \ref{fig:gripper_sampling}, we can observe that the robot gripper optimization problem has large infeasible regions ($RFS=0.1396$), which can be solved well by the proposed method MOEA/D-IEpsilon according to our previous analysis. To verify this hypothesis, MOEA/D-IEpislon and the other four decomposition-based CMOEAs are tested on the robot gripper optimization problems.

\begin{table}[htbp]
  \centering
  \caption{The parameters of the gripper optimization problem}
    \begin{tabular}{|c|c|c|}
    \hline
    Symbol & Meaning of parameter & Value \\
    \hline
    \multirow{2}[0]{*}{$Y_{min}$}  & Minimal dimension of & \multirow{2}[0]{*}{50mm} \\
                            & object to be gripped                     &   \\
    \hline
   \multirow{2}[0]{*} {$Y_{G}$}   & Maximal range of the &  \multirow{2}[0]{*}{150mm} \\
                            & gripper ends displacement &  \\

    \hline
   \multirow{2}[0]{*} {$Y_{max}$}  & Maximal dimension of  & \multirow{2}[0]{*}{100mm} \\
                            & object to be gripped       &       \\
    \hline
   \multirow{2}[0]{*} {$z_{max}$}  & Maximal displacement of & \multirow{2}[0]{*}{100mm} \\
                            & the gripper actuator     &       \\
    \hline
    \multirow{2}[0]{*} {$P$}     & Actuating force of the  & \multirow{2}[0]{*}{100N} \\
                             & gripper                 &      \\
    \hline
    \multirow{2}[0]{*}{$F_{G}$}    & The lower bound of & \multirow{2}[0]{*}{50N} \\
                            & gripping  force          &     \\
    \hline

    \end{tabular}%
  \label{tab:parameters}%
\end{table}%

\subsection{Experimental study}
\label{sec:6.2}

\subsubsection{Experimental settings}
To solve the robot gripper optimization problem and evaluate the performance of the proposed MOEA/D-IEpsilon, five decomposition-based CMOEAs, including MOEA/D-IEpsilon, MOEA/D-Epsilon, MOEA/D-SR, MOEA/D-CDP and C-MOEA/D with the differential evolution (DE) crossover, are tested on the robot gripper optimization problem. The detailed parameters of these five CMOEAs are the same as listed in Section \ref{sec:5.1} except for the number of function evaluations. In the case of the robot gripper optimization problem, each CMOEA stops when 600,000 function evaluations are reached. As the ideal PF of the gripper optimization problem is not known in advance, we use only the hypervolume metric (\cite {Zitzler1999Multiobjective}) to measure the performance of the five tested CMOEAs. In the robot gripper optimization case, the reference point $z^r = [5,800]^T$.

\subsubsection{Analysis of experiments}
\label{sec:6.2.1}
\begin{figure*}
\begin{tabular}{cc}
\begin{minipage}[t]{0.33\linewidth}
\includegraphics[width = 6cm]{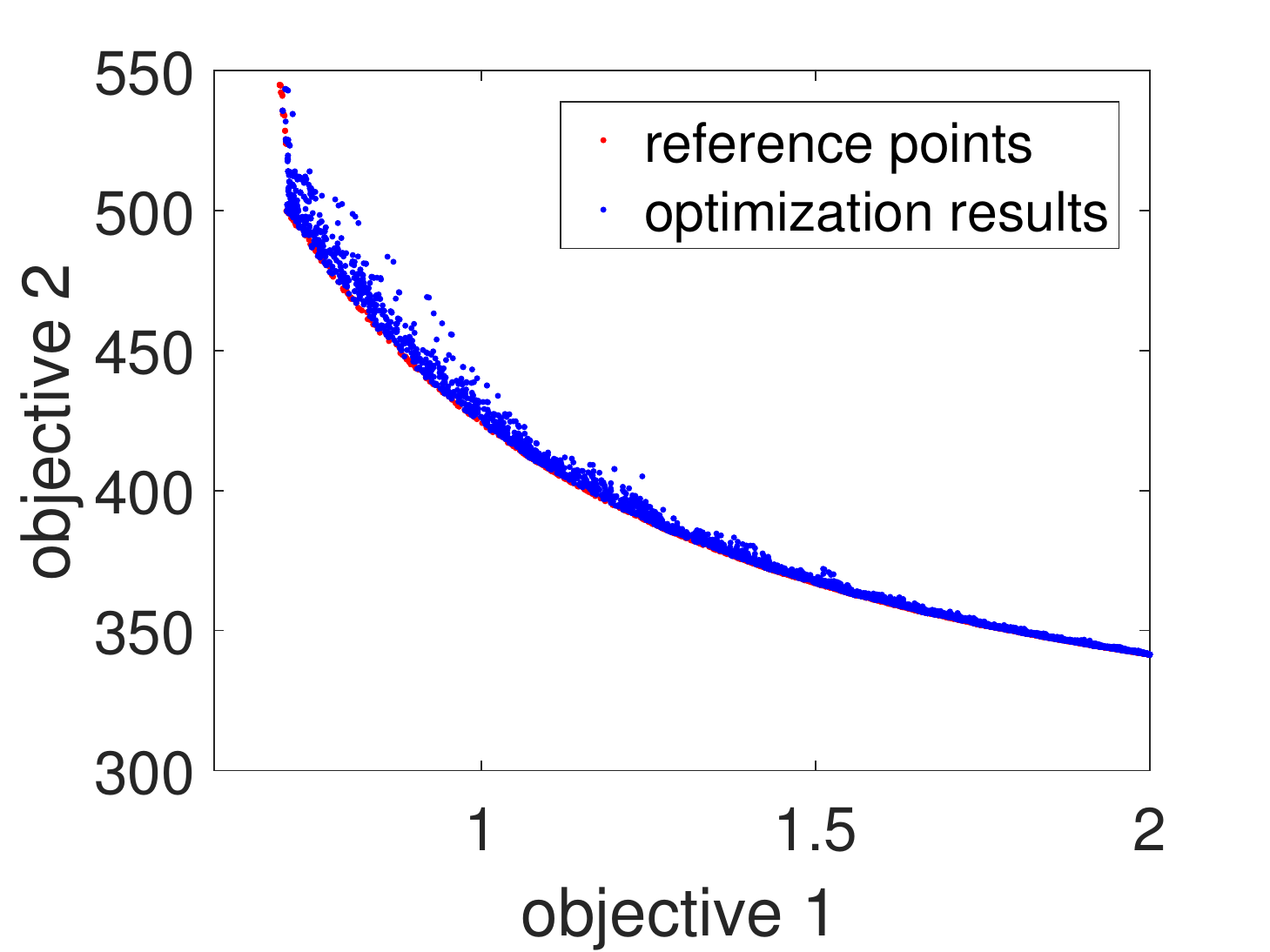}\\
\centering{\scriptsize{(a) MOEA/D-IEpsilon}}
\end{minipage}
\begin{minipage}[t]{0.33\linewidth}
\includegraphics[width = 6cm]{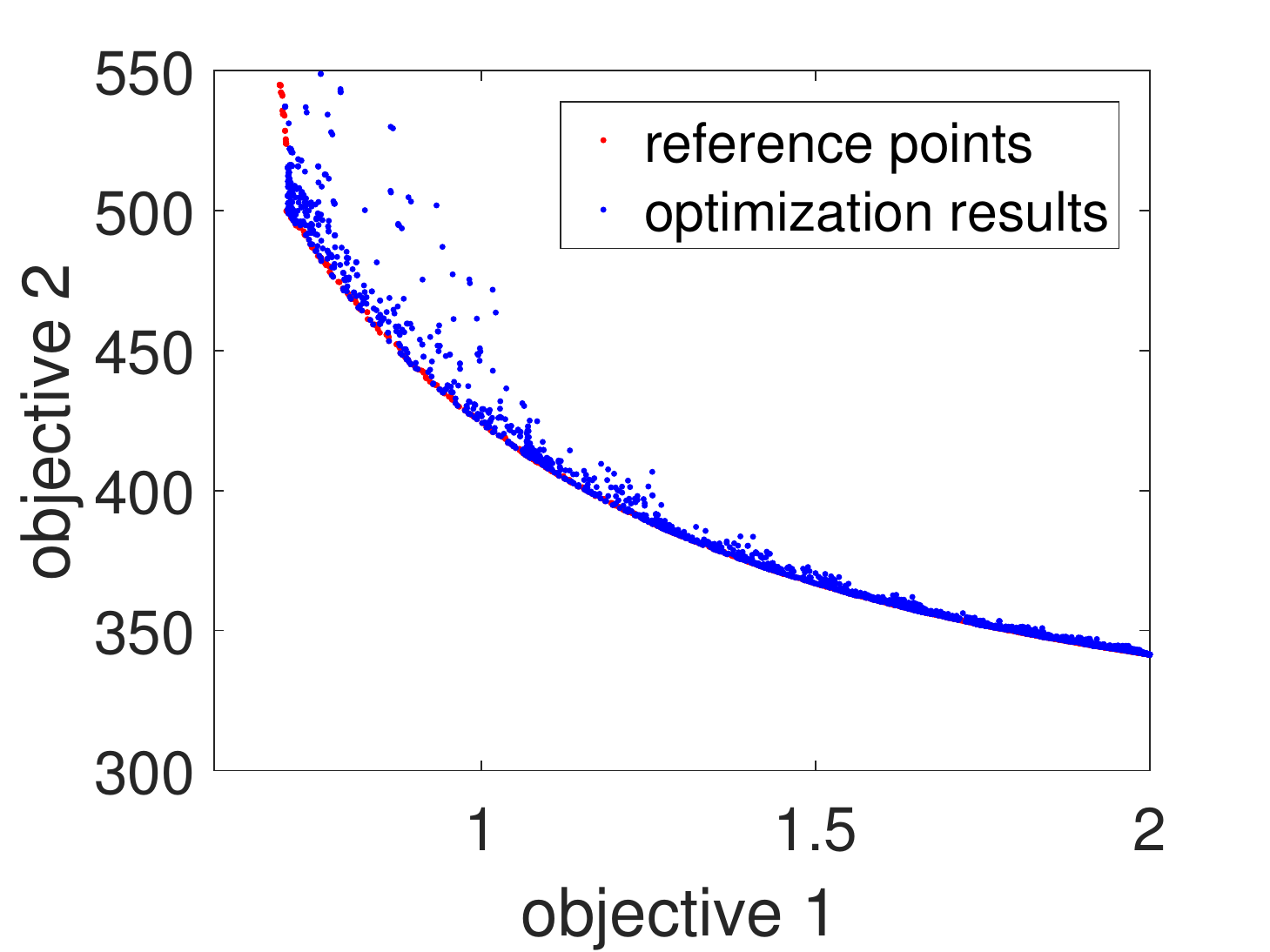}\\
\centering{\scriptsize{(b) MOEA/D-Epsilon}}
\end{minipage}
\begin{minipage}[t]{0.33\linewidth}
\includegraphics[width = 6cm]{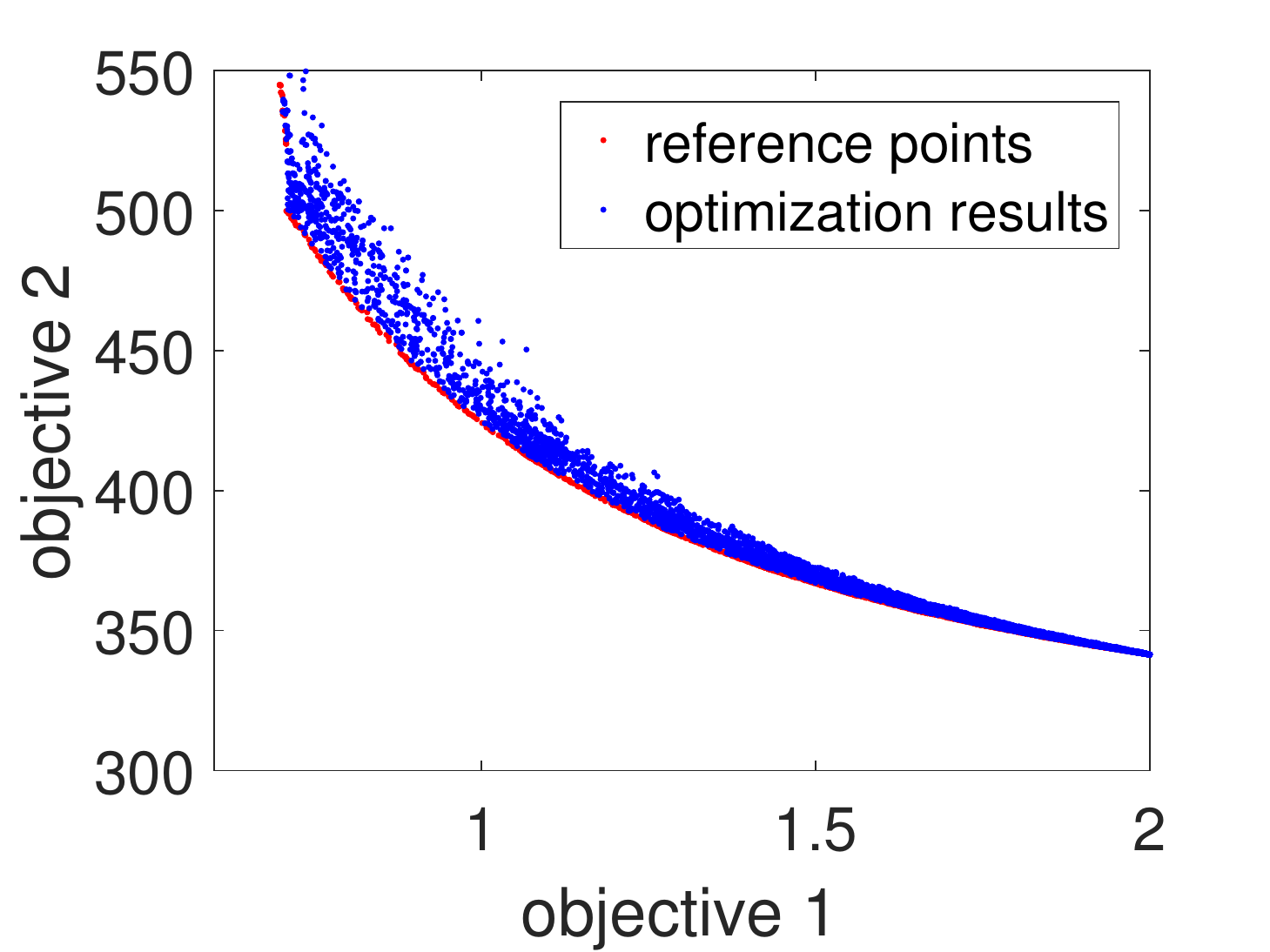}\\
\centering{\scriptsize{(c) MOEA/D-SR}}
\end{minipage}
\end{tabular}

\begin{tabular}{cc}
\begin{minipage}[t]{0.33\linewidth}
\includegraphics[width = 6cm]{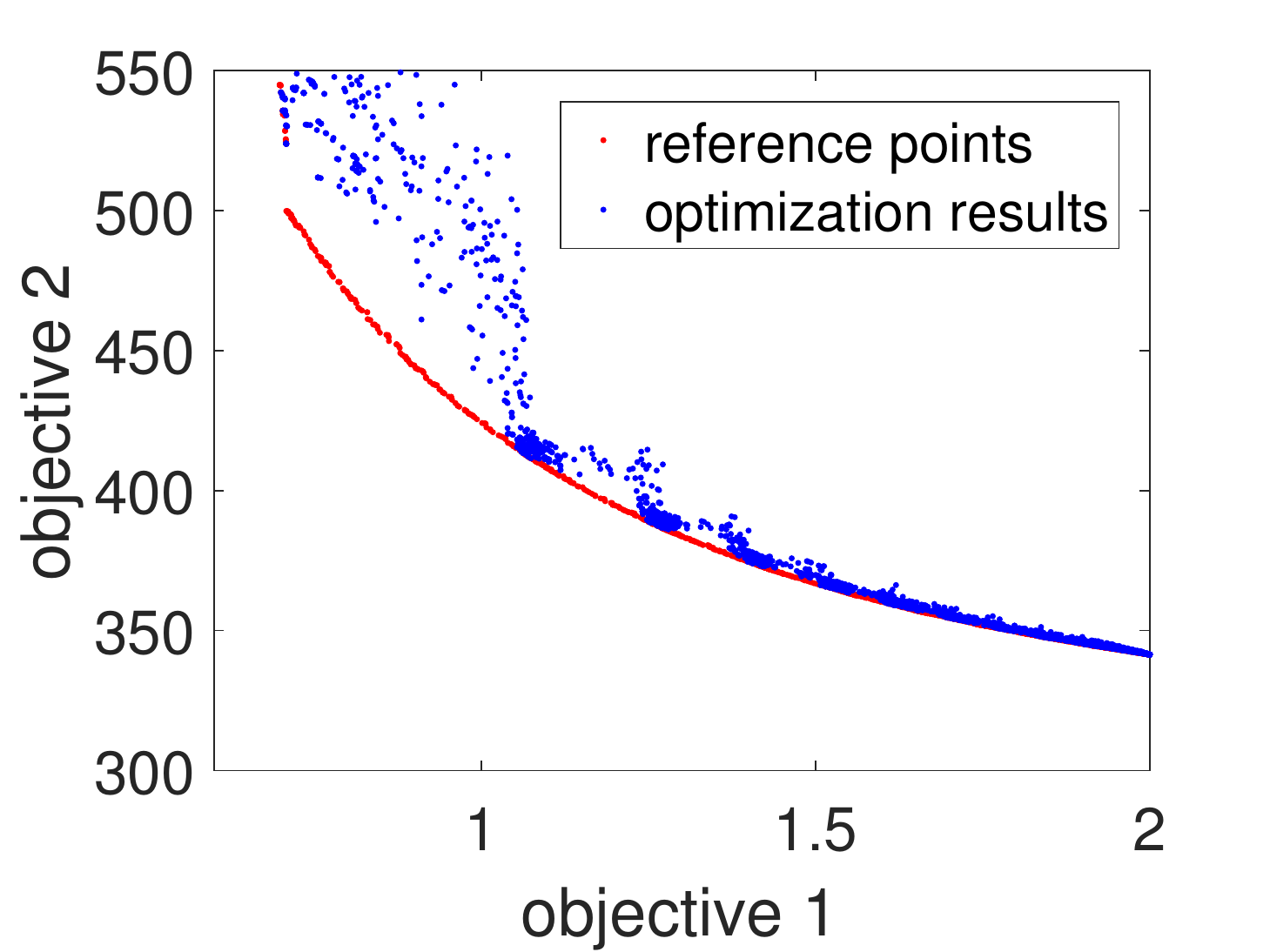}\\
\centering{\scriptsize{(d) MOEA/D-CDP}}
\end{minipage}
\begin{minipage}[t]{0.33\linewidth}
\includegraphics[width = 6cm]{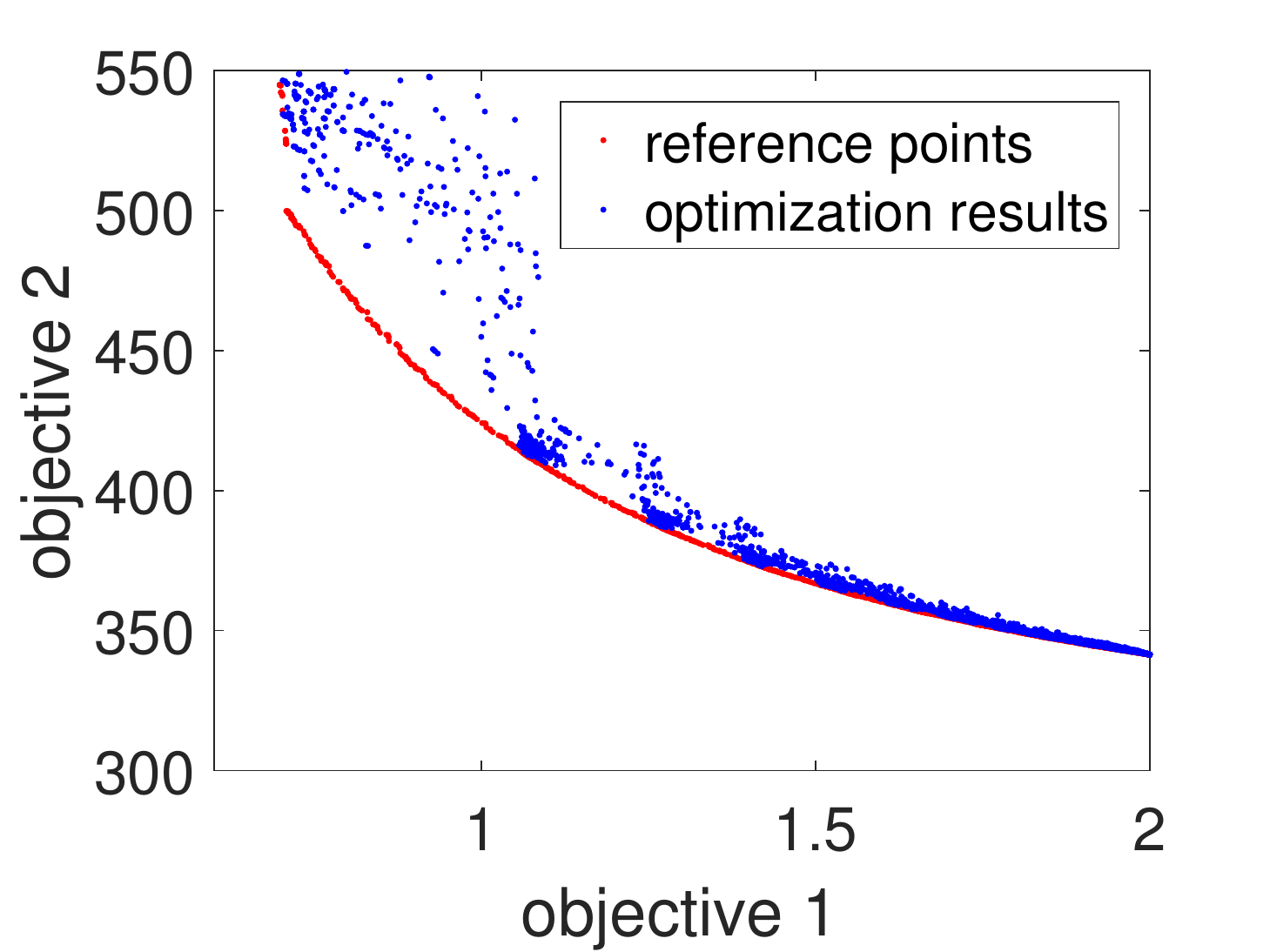}\\
\centering{\scriptsize{(e) C-MOEA/D}}
\end{minipage}

\begin{minipage}[t]{0.33\linewidth}
\includegraphics[width = 6cm,height = 4.5cm]{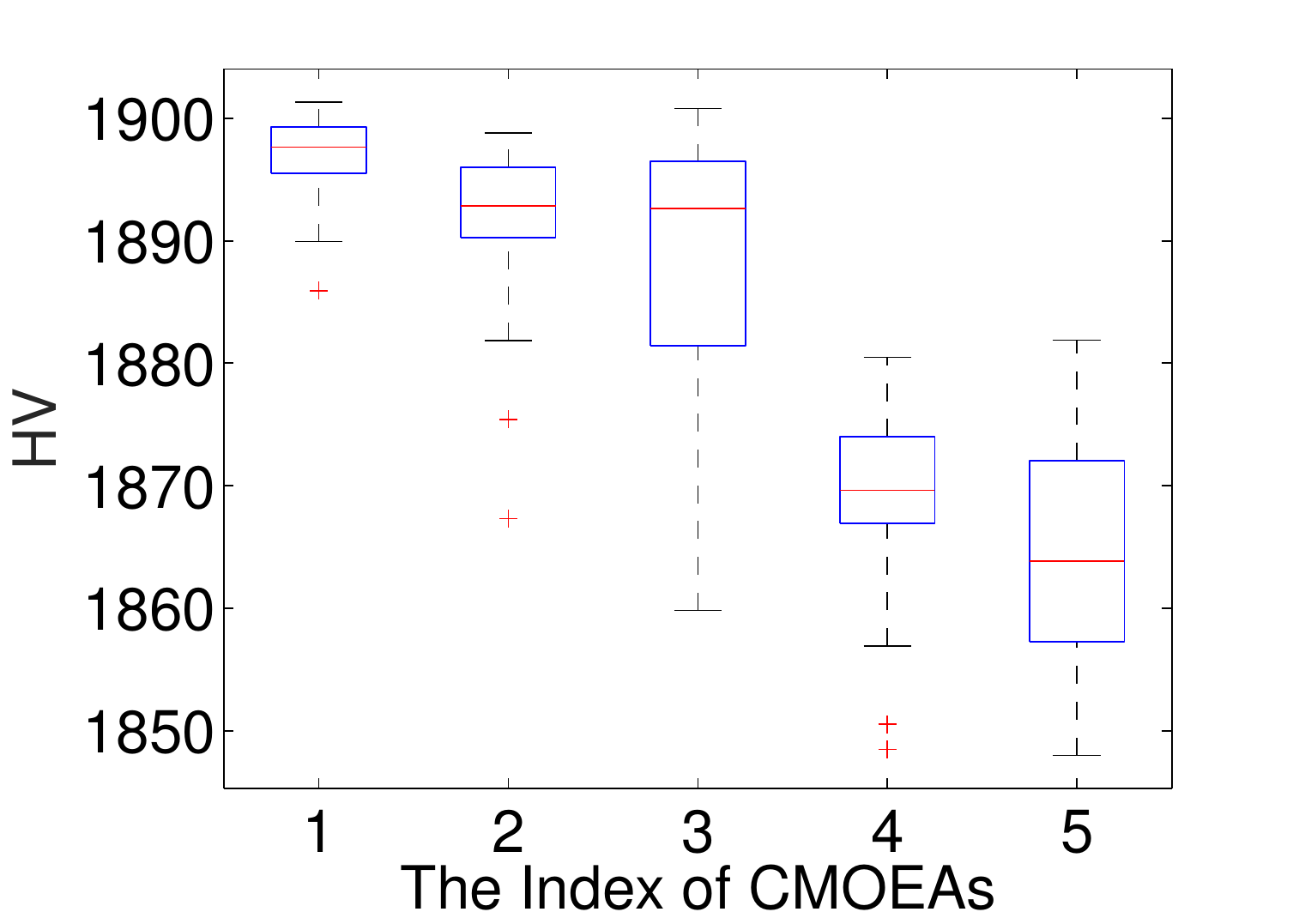}\\
\centering{\scriptsize{(f) The box plots of each CMOEA}}
\end{minipage}

\end{tabular}

\caption{The non-dominated solutions achieved by each algorithm during the 30 independent runs are plotted in (a)-(e). In (f), the box plots of each CMOEA are plotted.} 
\label{Fig:Gripper_result}
\end{figure*}

Table \ref{tab:gripper-HV} shows the statistical results of $HV$ values of MOEA/D-IEpsilon and the other four CMOEAs on the robot gripper optimization problem. It is clear that MOEA/D-IEpsilon is significantly better than the other four CMOEAs. To further demonstrate the superiority of the proposed method MOEA/D-IEpsilon, the non-dominated solutions achieved by each CMOEA during the 30 independent runs are plotted in Fig. \ref{Fig:Gripper_result}(a)-(e). The box plot of $HV$ values of the five CMOEAs is shown in Fig. \ref{Fig:Gripper_result}(f). From Fig. \ref{Fig:Gripper_result}, we see that MOEA/D-IEpsion has better performance than the other four CMOEAs.

\begin{table*}[htbp]
  \centering
  \caption{HV results of MOEA/D-IEpsilon and the other four CMOEAs on the gripper optimization problem}
    \begin{tabular}{c|c|ccccc}
    \toprule
    {Test Instances} & MOEA/D-IEpsilon & MOEA/D-Epsilon & MOEA/D-SR & MOEA/D-CDP & cMOEA/D \\
    \hline
    mean  & \textbf{1.897E+03} & 1.891E+03$^{\dag}$ & 1.889E+03$^{\dag}$
    & 1.869E+03$^{\dag}$ & 1.865E+03$^{\dag}$ \\
    std   & 3.510E+00 & 7.151E+00 & 9.839E+00 & 8.124E+00 & 9.048E+00 \\
    \bottomrule
    \end{tabular}%
  \label{tab:gripper-HV} \\
  Wilcoxon’s rank sum test at a 0.05 significance level is performed between MOEA/D-IEpsilon and each of the other four CMOEAs. $\dag$ and $\ddag$ denote that the performance of the corresponding algorithm is significantly worse than or better than that of MOEA/D-IEpsilon, respectively. The best mean is highlighted in boldface.
\end{table*}%

\begin{figure}
\includegraphics[width = 9cm]{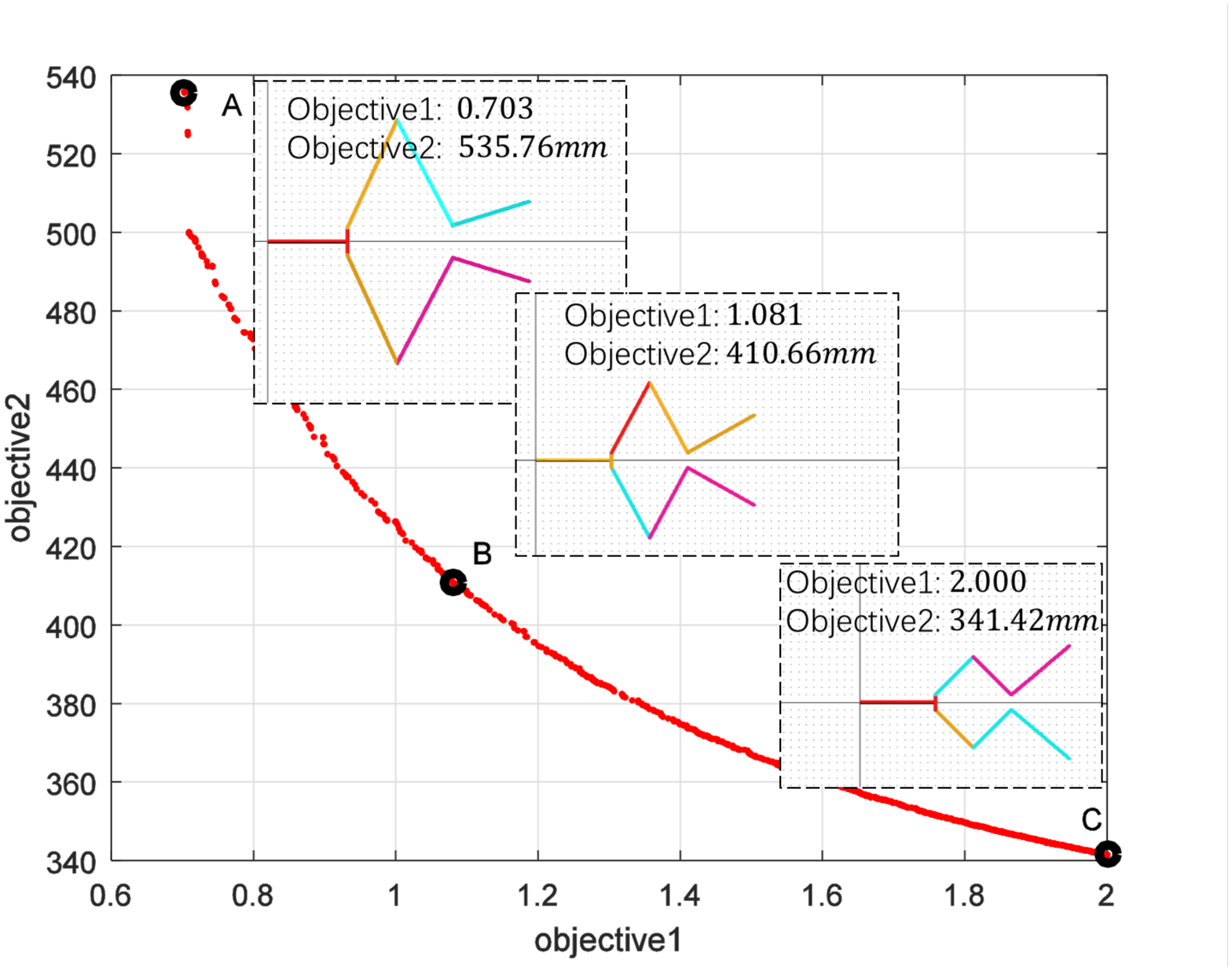}
\caption{The non-dominated solutions achieved by MOEA/D-IEpsilon.} 
\label{Fig:Gripper_opt3}
\end{figure}

In order to verify the correctness of the optimization results of the robot gripper optimization problem, three representative individuals (A, B and C) are selected from the non-dominated solutions achieved by MOEA/D-IEpsilon as shown in Fig. \ref{Fig:Gripper_opt3}. The configurations of the robot gripper mechanism at each point are also plotted in Fig. \ref{Fig:Gripper_opt3}. 

To measure the minimum gripping force $\min_z F_k(\mathbf{x},z)$, a spring with a large stiffness coefficient is set vertically at the end of the robot gripper during the simulation process. The spring force is regarded as the gripping force when the robot gripper is balanced by the spring. The simulation tool is ADAMS 2013, and the stiffness coefficient of the spring is $10^{13}$ N/m.

Table \ref{tab:gripper-sim} shows the simulation results of the minimum gripping force $\min_z F_k(\mathbf{x},z)$ with three different configurations of the robot gripper. The relative errors between the theoretical gripping forces and the simulated gripping forces are less than $0.1 \%$. Thus, we can conclude that the optimization results of the robot gripper optimization problem achieved by MOEA/D-IEpsilon are correct. 

\begin{table*}[htbp]
  \centering
  \caption{The simulated results of the minimum gripping force $\min_z F_k(\mathbf{x},z)$ with three different robot gripper configurations.}
    \begin{tabular}{|c|c|c|c|}
    \hline
    Sampled point & The theoretical gripping force (N)& The simulated result (N)& Relative error\\
    \hline
     A    & 50.0000  & 50.0002  & 0.0004\% \\
     \hline
     B    & 142.3168  & 142.4582  & 0.0994\% \\
     \hline
    C     & 92.5285  & 92.5877  & 0.0639\% \\
    \hline
    \end{tabular}%
  \label{tab:gripper-sim}%
\end{table*}%

\section{Conclusion}
\label{sec:7}
This paper proposes an improved epsilon constraint-handling method embedded in the framework of MOEA/D. A new CMOEA named MOEA/D-IEpsilon has been proposed. The comprehensive experimental results indicate that MOEA/D-IEpsilon has the ability to cross the large infeasible regions. Compared with the other four decomposition-based CMOEAs including MOEA/D-Epsilon, MOEA/D-SR, MOEA/D-CDP and C-MOEA/D, MOEA/D-IEpsilon has following advantages:

\begin{itemize}
  \item The performance of MOEA/D-IEpsilon is not sensitive to the initial epsilon value.
  \item MOEA/D-IEpsilon has the ability to explore the feasible and infeasible regions simultaneously during the evolutionary process.
  \item MOEA/D-IEpsilon utilizes the feasible ratio of the current population to dynamically balance the exploration between the feasible regions and infeasible regions. It keeps a good balance of the searching between infeasible and feasible regions.
  \item MOEA/D-IEpsilon is suitable for solving CMOPs with large infeasible regions.
\end{itemize}

In terms of CMOPs, a new set of CMOPs named LIR-CMOP1-14 was designed and presented in this paper. A common feature of these test instances is that they have large infeasible regions. The experimental results show that MOEA/D-IEpsion is significantly better than the other four CMOEAs on this test suite. Thus, we hypothesize that MOEA/D-IEpsilon is better than the other four CMOEAs in solving CMOPs with large infeasible regions, in general. To demonstrate the capacity of MOEA/D-IEpsilon to solve real engineering problems, a robot gripper optimization problem with two conflicting objectives and eight constraints was used as a test problem. The experimental results also demonstrated that MOEA/D-IEpsilon outperformed the other four CMOEAs.

Proposed further work includes studying new constraint-handling mechanisms to solve CMOPs with different types of difficulty. One possible way is to collect more information about the working population, and utilize such information to guide a CMOEA to select appropriate constraint-handling methods in different evolutionary stages.

\begin{acknowledgements}
This work was supported in part by the National Natural Science Foundation of China (NSFC) under grant 61300159, 61473241 and 61332002, by the Natural Science Foundation of Jiangsu Province of China under grant BK20130808, by the Project of Internation as well as Hongkong，Macao\&Taiwan Science and Technology Cooperation Innovation Platform in Universities in Guangdong Province under grant 2015KGJH2014, by China Postdoctoral Science Foundation under grant 2015M571751, by the Science and Technology Planning Project of Guangdong Province of China under grant 2013B011304002, by Educational Commission of Guangdong Province of China under grant 2015KGJHZ014, by the Fundamental Research Funds for the Central Universities of China under grant NZ2013306, and by the Guangdong High-Level University Project ``Green Technologies'' for Marine Industries.
\end{acknowledgements}

\section {Appendix}
\label{sec:8}   
In this section, the detailed definitions of LIR-CMOP1-14 are listed in Table \ref{tab:ncmops}.

\begin {table*} [tbp]
\tiny
\centering
\tabcolsep 0.01 in
\caption{The objectives and constraints of LIR-CMOP1-14.}
\label{tab:ncmops}
\begin{tabular}{|l|l|l|}
\hline
Problem & Objectives & Constraints \\
\hline
LIR-CMOP1
& $\begin{cases}
f_1(x) = x_1 + g_1(x) \\
f_2(x) = 1 - x_1^2 + g_2(x) \\
g_1(x) = \sum_{j\in J_1}  {(x_j  - sin(0.5 \pi x_1))} ^{2} \\
g_2(x) = \sum_{j\in J_2} {(x_j  - cos(0.5 \pi x_1))} ^{2} \\
J_1 = \{3,5,\ldots,29\}, J_2 = \{2,4,\ldots,30\}
\end{cases}$
& $\begin{cases}
c_1(x) = (a - g_1(x)) * (g_1(x) - b) \ge 0\\
c_2(x) = (a - g_2(x)) * (g_2(x) - b) \ge 0\\
a = 0.51, b = 0.5 \\
x \in [0,1] ^ {30}
\end{cases}$\\
 \hline
LIR-d2
& $\begin{cases}
f_1(x) = x_1 + g_1(x) \\
f_2(x) = 1 -\sqrt{x_1} + g_2(x) \\
g_1(x) = \sum_{j\in J_1}  {(x_j  - sin(0.5 \pi x_1))} ^{2} \\
g_2(x) = \sum_{j\in J_2} {(x_j  - cos(0.5 \pi x_1))} ^{2} \\
J_1 = \{3,5,\ldots,29\}, J_2 = \{2,4,\ldots,30\}
 \end{cases}$
 & $\begin{cases}
 c_1(x) = (a - g_1(x)) * (g_1(x) - b) \ge 0\\
 c_2(x) = (a - g_2(x)) * (g_2(x) - b) \ge 0\\
 a = 0.51, b = 0.5\\
 x \in [0,1]^{30}
 \end{cases}$\\
 \hline
LIR-CMOP3
& $\begin{cases}
f_1(x) = x_1 + g_1(x) \\
f_2(x) = 1 - x^2_1 + g_2(x) \\
g_1(x) = \sum_{j\in J_1}  {(x_j  - sin(0.5 \pi x_1))} ^{2} \\
g_2(x) = \sum_{j\in J_2} {(x_j  - cos(0.5 \pi x_1))} ^{2} \\
J_1 = \{3,5,\ldots,29\}, J_2 = \{2,4,\ldots,30\}
\end{cases}$
 & $\begin{cases}
 c_1(x) = (a - g_1(x)) * (g_1(x) - b) \ge 0\\
 c_2(x) = (a - g_2(x)) * (g_2(x) - b) \ge 0\\
 c_3(x) = sin(c\pi x_1) - 0.5 \ge 0 \\
 a = 0.51, b = 0.5, c = 20 \\
 x \in [0,1]^{30}\\
 \end{cases}$\\
 \hline
LIR-CMOP4
& $\begin{cases}
f_1(x) = x_1 + g_1(x) \\
f_2(x) = 1 - \sqrt{x_1} + g_2(x) \\
g_1(x) = \sum_{j\in J_1}  {(x_j  - sin(0.5 \pi x_1))} ^{2} \\
g_2(x) = \sum_{j\in J_2} {(x_j  - cos(0.5 \pi x_1))} ^{2} \\
J_1 = \{3,5,\ldots,29\}, J_2 = \{2,4,\ldots,30\}
\end{cases}$
 & $\begin{cases}
 c_1(x) = (a - g_1(x)) * (g_1(x) - b) \ge 0\\
 c_2(x) = (a - g_2(x)) * (g_2(x) - b) \ge 0\\
 c_3(x) = sin(c\pi x_1) - 0.5 \ge 0 \\
 a = 0.51, b = 0.5, c = 20\\
 x \in [0,1]^{30}\\
 \end{cases}$\\
 \hline

LIR-CMOP5 & $\begin{cases} & f_1(x) = x_1 + 10 * g_1(x) + 0.7057 \\ & f_2(x) = 1 - \sqrt{x_1} + 10 * g_2(x) + 0.7057 \\ & g_1(x) = \sum_{i \in J_1} {(x_i  - \sin(\frac{0.5i}{30} \pi x_1))} ^{2}\\& g_2(x) = \sum_{j \in J_2} {(x_j  - \cos(\frac{0.5j}{30} \pi x_1))} ^{2} \\ & J_1 = \{3,5,\ldots,29\}, J_2 = \{2,4,\ldots,30\}\end{cases}$ &
$\begin{cases}
&c_k(x) = ((f_1 - p_k) \cos\theta_k - ( f_2 - q_k) \sin\theta_k)^2 / a_k^2 \\
& + ((f_1 - p_k) \sin\theta_k + (f_2 - q_k) \cos\theta_k)^2 / b_k ^{2} \ge r \\
& p_{k}=[1.6,2.5], q_{k}=[1.6,2.5]\\
& a_k = [2,2], b_k = [4,8]\\
&r = 0.1, \theta_k = -0.25\pi \\
&x \in [0,1] ^ {30}, k= 1,2\\
\end{cases}$
 \\
\hline
LIR-CMOP6 & $\begin{cases} & f_1(x) = x_1 + 10 * g_1(x) + 0.7057 \\ & f_2(x) = 1 - x_1^2 + 10 * g_2(x) + 0.7057 \\ & g_1(x) = \sum_{i \in J_1} {(x_i  - \sin(\frac{0.5i}{30} \pi x_1))} ^{2}\\& g_2(x) = \sum_{j \in J_2} {(x_j  - \cos(\frac{0.5j}{30} \pi x_1))} ^{2} \\ & J_1 = \{3,5,\ldots,29\}, J_2 = \{2,4,\ldots,30\}\end{cases}$ &
$\begin{cases}
&c_k(x) = ((f_1 - p_k) \cos\theta_k - ( f_2 - q_k) \sin\theta_k)^2 / a_k^2 \\
& + ((f_1 - p_k) \sin\theta_k + (f_2 - q_k) \cos\theta_k)^2 / b_k ^{2} \ge r \\
& p_{k}=[1.8,2.8], q_{k}=[1.8,2.8]\\
& a_k = [2,2], b_k = [8,8]\\
&r = 0.1, \theta_k = -0.25\pi \\
&x \in [0,1] ^ {30}, k= 1,2\\
\end{cases}$
 \\
\hline

\hline
LIR-CMOP7 & $\begin{cases} & f_1(x) = x_1 + 10 * g_1(x) + 0.7057 \\ & f_2(x) = 1 - \sqrt{x_1} + 10 * g_2(x) + 0.7057 \\ & g_1(x) = \sum_{i \in J_1} {(x_i  - \sin(\frac{0.5i}{30} \pi x_1))} ^{2}\\& g_2(x) = \sum_{j \in J_2} {(x_j  - \cos(\frac{0.5j}{30} \pi x_1))} ^{2} \\ & J_1 = \{3,5,\ldots,29\}, J_2 = \{2,4,\ldots,30\}\end{cases}$ &
$\begin{cases}
&c_k(x) = ((f_1 - p_k) \cos\theta_k - ( f_2 - q_k) \sin\theta_k)^2 / a_k^2 \\
& + ((f_1 - p_k) \sin\theta_k + (f_2 - q_k) \cos\theta_k)^2 / b_k ^{2} \ge r \\
& p_{k}=[1.2,2.25,3.5], q_{k}=[1.2,2.25,3.5]\\
& a_k = [2,2.5,2.5], b_k = [6,12,10]\\
& r = 0.1, \theta_k = -0.25\pi \\
& x \in [0,1]^{30}, k= 1,2,3\\
\end{cases}$
 \\
\hline
LIR-CMOP8 & $\begin{cases} & f_1(x) = x_1 + 10 * g_1(x) + 0.7057 \\ & f_2(x) = 1 - x_1^2 + 10 * g_2(x) + 0.7057 \\ & g_1(x) = \sum_{i \in J_1} {(x_i  - \sin(\frac{0.5i}{30} \pi x_1))} ^{2}\\& g_2(x) = \sum_{j \in J_2} {(x_j  - \cos(\frac{0.5j}{30} \pi x_1))} ^{2} \\ & J_1 = \{3,5,\ldots,29\}, J_2 = \{2,4,\ldots,30\}\end{cases}$ &
$\begin{cases}
&c_k(x) = ((f_1 - p_k) \cos\theta_k - ( f_2 - q_k) \sin\theta_k)^2 / a_k^2 \\
& + ((f_1 - p_k) \sin\theta_k + (f_2 - q_k) \cos\theta_k)^2 / b_k ^{2} \ge r \\
& p_{k}=[1.2,2.25,3.5], q_{k}=[1.2,2.25,3.5]\\
& a_k = [2,2.5,2.5], b_k = [6,12,10]\\
&r = 0.1, \theta_k = -0.25\pi \\
& x \in [0,1] ^{30}, k= 1,2,3\\
\end{cases}$
 \\
\hline
LIR-CMOP9 & $\begin{cases} & f_1(x) = 1.7057x_1(10 * g_1(x)+1) \\ & f_2(x) = 1.7057(1 - x_1^2)(10 * g_2(x)+1) \\ & g_1(x) = \sum_{i \in J_1} {(x_i  - \sin(\frac{0.5i}{30} \pi x_1))} ^{2}\\& g_2(x) = \sum_{j \in J_2} {(x_j  - \cos(\frac{0.5j}{30} \pi x_1))} ^{2} \\ & J_1 = \{3,5,\ldots,29\}, J_2 = \{2,4,\ldots,30\}\end{cases}$ &
$\begin{cases}
&c_1(x) = ((f_1 - p_1) \cos\theta_1 - ( f_2 - q_1) \sin\theta_1)^2 / a_1^2 \\
& + ((f_1 - p_1) \sin\theta_1 + (f_2 - q_1) \cos\theta_1)^2 / b_1 ^{2} \ge r \\
&c_2(x) = f_1 \sin\alpha + f_2 \cos\alpha\\
& - \sin(4 \pi (f_1 \cos\alpha - f_2 \sin\alpha)) - 2 \ge 0\\
& p_{1}=1.4, q_{1}=1.4, a_1 = 1.5, b_1 = 6.0\\
&r = 0.1,\alpha = 0.25\pi, \theta_1 = -0.25\pi \\
& x \in [0,1]^{30}\\
\end{cases}$
 \\
\hline
LIR-CMOP10 & $\begin{cases} & f_1(x) = 1.7057x_1(10 * g_1(x)+1) \\ & f_2(x) = 1.7057(1 - \sqrt{x_1})(10 * g_2(x)+1) \\ & g_1(x) = \sum_{i \in J_1} {(x_i  - \sin(\frac{0.5i}{30} \pi x_1))} ^{2}\\& g_2(x) = \sum_{j \in J_2} {(x_j  - \cos(\frac{0.5j}{30} \pi x_1))} ^{2} \\ & J_1 = \{3,5,\ldots,29\}, J_2 = \{2,4,\ldots,30\}\end{cases}$ &
$\begin{cases}
&c_1(x) = ((f_1 - p_1) \cos\theta_1 - ( f_2 - q_1) \sin\theta_1)^2 / a_1^2 \\
& + ((f_1 - p_1) \sin\theta_1 + (f_2 - q_1) \cos\theta_1)^2 / b_1 ^{2} \ge r \\
&c_2(x) = f_1 \sin\alpha + f_2 \cos\alpha\\
& - \sin(4 \pi (f_1 \cos\alpha - f_2 \sin\alpha)) - 1 \ge 0\\
& p_{1}=1.1, q_{1}=1.2, a_1 = 2.0, b_1 = 4.0\\
&r = 0.1,\alpha = 0.25\pi, \theta_1 = -0.25\pi \\
& x \in [0,1]^{30}\\
\end{cases}$
 \\
\hline
LIR-CMOP11 & $\begin{cases} & f_1(x) = 1.7057x_1(10 * g_1(x)+1) \\ & f_2(x) = 1.7057(1 - \sqrt{x_1})(10 * g_2(x)+1) \\ & g_1(x) = \sum_{i \in J_1} {(x_i  - \sin(\frac{0.5i}{30} \pi x_1))} ^{2}\\& g_2(x) = \sum_{j \in J_2} {(x_j  - \cos(\frac{0.5j}{30} \pi x_1))} ^{2} \\ & J_1 = \{3,5,\ldots,29\}, J_2 = \{2,4,\ldots,30\}\end{cases}$ &
$\begin{cases}
&c_1(x) = ((f_1 - p_1) \cos\theta_1 - ( f_2 - q_1) \sin\theta_1)^2 / a_1^2 \\
& + ((f_1 - p_1) \sin\theta_1 + (f_2 - q_1) \cos\theta_1)^2 / b_1 ^{2} \ge r \\
&c_2(x) = f_1 \sin\alpha + f_2 \cos\alpha\\
& - \sin(4 \pi (f_1 \cos\alpha - f_2 \sin\alpha)) - 2.1 \ge 0\\
& p_{1}=1.2, q_{1}=1.2, a_1 = 1.5, b_1 = 5.0\\
&r = 0.1,\alpha = 0.25\pi, \theta_1 = -0.25\pi \\
& x \in [0,1]^{30}\\
\end{cases}$
 \\
\hline
LIR-CMOP12 & $\begin{cases} & f_1(x) = 1.7057x_1(10 * g_1(x)+1) \\ & f_2(x) = 1.7057(1 - x_1^2)(10 * g_2(x)+1) \\ & g_1(x) = \sum_{i \in J_1} {(x_i  - \sin(\frac{0.5i}{30} \pi x_1))} ^{2}\\& g_2(x) = \sum_{j \in J_2} {(x_j  - \cos(\frac{0.5j}{30} \pi x_1))} ^{2} \\ & J_1 = \{3,5,\ldots,29\}, J_2 = \{2,4,\ldots,30\}\end{cases}$ &
$\begin{cases}
&c_1(x) = ((f_1 - p_1) \cos\theta_1 - ( f_2 - q_1) \sin\theta_1)^2 / a_1^2 \\
& + ((f_1 - p_1) \sin\theta_1 + (f_2 - q_1) \cos\theta_1)^2 / b_1 ^{2} \ge r \\
&c_2(x) = f_1 \sin\alpha + f_2 \cos\alpha\\
& - \sin(4 \pi (f_1 \cos\alpha - f_2 \sin\alpha)) - 2.5 \ge 0\\
& p_{1}=1.6, q_{1}=1.6, a_1 = 1.5, b_1 = 6.0\\
&r = 0.1,\alpha = 0.25\pi, \theta_1 = -0.25\pi \\
& x \in [0,1]^{30}\\
\end{cases}$
 \\
\hline
LIR-CMOP13 & $\begin{cases} & f_1(x) = (1.7057 + g_1) \cos (0.5 \pi x_1) \cos (0.5 \pi x_2) \\& f_2(x) = (1.7057 + g_1) \cos (0.5 \pi x_1) \sin (0.5 \pi x_2) \\ & f_3(x) = (1.7057 + g_1) \sin(0.5 \pi x_1) \\ & g_1 = \sum_{i \in J} {10 (x_i - 0.5)^2}\\ & J = \{3,4,\ldots,30\}\end{cases}$ &
$\begin{cases}
&c_1(x) = (g(x) - 9)(g(x) - 4)\\
&c_2(x) = (g(x) - 3.61)(g(x) - 3.24)\\
& g(x) = f_1^2 + f_2^2 + f_3^2\\
& x \in [0,1]^{30}\\
\end{cases}$
 \\
\hline
LIR-CMOP14 & $\begin{cases} & f_1(x) = (1.7057 + g_1) \cos (0.5 \pi x_1) \cos (0.5 \pi x_2) \\& f_2(x) = (1.7057 + g_1) \cos (0.5 \pi x_1) \sin (0.5 \pi x_2) \\ & f_3(x) = (1.7057 + g_1) \sin(0.5 \pi x_1) \\ & g_1 = \sum_{i \in J} {10 (x_i - 0.5)^2}\\ & J = \{3,4,\ldots,30\}\end{cases}$ &
$\begin{cases}
&c_1(x) = (g(x) - 9)(g(x) - 4)\\
&c_2(x) = (g(x) - 3.61)(g(x) - 3.24)\\
&c_3(x) = (g(x) - 3.0625)(g(x) - 2.56)\\
& g(x) = f_1^2 + f_2^2 + f_3^2\\
& x \in [0,1]^{30}\\
\end{cases}$
 \\
\hline
\end{tabular}
\end{table*}

\section{Compliance with Ethical Standards}

\noindent \textbf{Conflict of Interest} The authors declare that they have no conflict of interest.

\noindent \textbf{Ethical approval} This article does not contain any studies with human participants or animals performed by any of the authors.

%\begin{acknowledgements}
%\end{acknowledgements}


% \begin{thebibliography}{}
% %
% % and use \bibitem to create references. Consult the Instructions
% % for authors for reference list style.
% % Format for Journal Reference
% Author, Article title, Journal, Volume, page numbers (year)
% Author, Book title, page numbers. Publisher, place (year)
% \end{thebibliography}
\bibliographystyle{spbasic}
\bibliography{improved_epsilon}

\end{document}